\documentclass[journal]{IEEEtran}

\usepackage{epsfig}
\usepackage{amssymb}
\usepackage{amsthm}
\usepackage{graphicx}
\usepackage{xcolor}
\usepackage{boldline,multirow}
\usepackage{hyperref}
\usepackage{amsmath}

\usepackage{pifont}

\begin{document}

\title{Lane-Keeping Control of Autonomous Vehicles Through a Soft-Constrained Iterative LQR}

\author{Der-Hau Lee
\thanks{Copyright (c) 20xx IEEE. Personal use of this material is permitted. However, permission to use this material for any other purposes must be obtained from the IEEE by sending a request to pubs-permissions@ieee.org.}%
\thanks{The author was with the Department of Electrophysics,  National Yang Ming Chiao Tung University, Hsinchu 300, Taiwan. e-mail: derhaulee@gmail.com.}
}

\maketitle

\begin{abstract}
The accurate prediction of smooth steering inputs is crucial for automotive applications because control actions with jitter might cause the vehicle  system  to become unstable. To address this problem in automobile lane-keeping control without the use of additional smoothing algorithms, we  developed a novel soft-constrained iterative linear\textendash quadratic regulator (soft-CILQR) algorithm by integrating CILQR algorithm and a model predictive control (MPC) constraint relaxation method. We incorporated  slack variables into the state and control barrier functions of the soft-CILQR solver to soften the constraints in the optimization process such that control input stabilization can be achieved in a computationally simple manner. Two types of automotive lane-keeping experiments (numerical simulations and experiments involving challenging vision-based  maneuvers) were conducted with a linear system dynamics model to test the performance of the proposed soft-CILQR algorithm, and its performance was compared with that of the CILQR algorithm. In the numerical simulations, the soft-CILQR and CILQR solvers managed to drive the system  toward the reference state asymptotically; however, the soft-CILQR solver obtained smooth steering input trajectories  more easily than did the CILQR solver under conditions involving additive disturbances. The results of the vision-based experiments in which an ego vehicle drove in perturbed TORCS environments with various road friction settings were consistent with those of the numerical tests.  The proposed soft-CILQR algorithm achieved an average runtime of 2.55 ms and is thus applicable for real-time autonomous driving scenarios.

\end{abstract}

\section{Introduction}
\IEEEPARstart{G}eNERATING smooth control actions is a fundamental requirement in controller design because input jitter might increase the computational burden  of actuators or even  damage them. This problem has become crucial in practical applications. For example, the presence of various types of disturbances in vision-based vehicle control systems \cite{Che15, Li19, Wu23, Wa23} causes poor perception performance, which might result in the generation of jittery control predictions by the controller. For optimization-based controllers, a straightforward solution for the aforementioned problem involves employing smoothing algorithms, such as the Savitzky\textendash Golay filter \cite{Sav64}, to smooth control signals. However, such operations are not included in the optimization process; thus, control inputs might not be optimal. Studies have proposed rigorous approaches for the smoothing of control actions without the use of external smoothing algorithms. Examples of such approaches include those involving the tuning of the weighting matrices of control variables terms in model predictive control (MPC) cost functions \cite{Mat19},  the regularization of policy output in a reinforcement-learning-based solution \cite{Chi21}, and  the addition of derivative action terms in  the cost functions of a model predictive path integral control algorithm \cite{Kim22}. Specifically, stability can also be achieved using MPC schemes with relaxed constraints.

The authors of \cite{Zhe95} formulated a soft-constrained MPC approach to handle infeasible optimization problems. They relaxed state constraints by introducing slack variables into the optimization process. They added a standard quadratic penalty term to the cost function to penalize  excessively high values of  slack variables. Consequently, their controller was determined to be less conservative and allowed temporal violations of  constraints. Subsequently, the authors of \cite{Raw00} introduced an additional linear penalty term into the cost function \cite{Zhe95} to remove counterintuitive solutions of the aforementioned controller. In \cite{Ask17}, a procedure was proposed for computing the upper bound of the prediction horizon, and the corresponding finite-horizon soft-constrained MPC problem was solved under stability guarantees. For linear systems with polytopic constraints, the authors of \cite{Zei22} proposed a soft-constrained MPC approach that uses polytopic terminal sets and can guarantee asymptotic system stability. In \cite{Svr23}, a soft-constrained MPC method was developed for linear systems; the model considers the general forms of the relaxed stage and terminal constraints. The novel feature of this method is that it involves the consideration of the terminal dynamics  of the augmented state\textendash slack variables, which guarantees a decrease in the terminal cost of the considered MPC problem.

The replacement of constraints with barrier functions can increase the stability of solutions for MPC problems. In the MPC formulation developed in \cite{Wil04}, weighted barrier functions that ensure that inequality constraints are strictly satisfied are included in the cost function. These barriers enable smooth control actions to be achieved when the system approaches constraint bounds. The authors of \cite{ Fel17a} leveraged relaxed barrier functions in linear MPC. This approach enables the enlargement of the feasible solution set for a hard-constrained MPC problem \cite{Wil04}. In a  numerical example, the authors of \cite{ Fel17a} showed  that state trajectories can converge to the reference point in MPC schemes even if the initial conditions are infeasible. The  authors of \cite{Fel17a} also investigated solutions to closed-loop systems with unknown additive disturbances by using relaxed barrier-function-based linear MPC schemes and demonstrated robust input-to-state stability. The state trajectory can thus be guaranteed to converge to a neighborhood of the desired set point \cite{Fel15}. By contrast, tube-based MPC approaches \cite{Mat19} use predefined bounded disturbances and are prone to produce conservative (restrictive) solution trajectories.

However, general MPC solvers, such as sequential quadratic programming (SQP) has poor computational efficiency in real-time applications. Researchers \cite{Che17, Che19} have therefore investigated the use of barrier functions to shape system constraints in the context of the iterative linear\textendash quadratic regulator (ILQR) algorithm \cite{Li04, Li05}, resulting in the creation of the constrained ILQR (CILQR) algorithm. Instead of solving computationally expensive nonlinear  optimization problems online, the system dynamic model is linearized at each time step in the  CILQR framework. Thus, CILQR can be considered to be a suboptimal linear time-varying MPC approach. The CILQR algorithm has been used to solve motion planning problems for  automotive systems and has exhibited higher computation efficiency than the standard SQP algorithm in these problems \cite{Che19}. Moreover, the authors in \cite{Liu24} proposed a spatiotemporal trajectory planning  method based on reachable set and ILQR technique to control a real vehicle in high-density complex traffic flow.

In \cite{Lee22}, we proposed an algorithm comprising a deep neural network (DNN) for extracting image features and CILQR modules for efficient  automotive motion planning and control in the open racing car simulator (TORCS)  \cite{Wym13, Car15}. Although vision-based autonomous driving methods that use cameras for environment perception are cheaper than expensive LiDAR-based methods \cite{Che15, Wa23}, visual perception systems have reduced performance in adverse driving conditions that can result in control instability for self-driving vehicles \cite{Li19}. An analysis of the robustness  of the CILQR algorithm in these conditions was lacking in our previous work \cite{Lee22}. Disturbances must be considered in autonomous driving algorithms because they may compromise driving comfort and safety. Motivated by well-established MPC techniques \cite{Svr23, Wil04, Fel15}, this study presents an  algorithm that integrates the CILQR solver \cite{Che17, Che19} and the soft-constrained MPC approach developed in \cite{Svr23} to achieve increased robustness while maintaining high computational performance. This algorithm is referred to as soft-CILQR  throughout this paper. Our goal was to develop a fast lane-keeping controller that  can generate smooth control actions for vision-based automated vehicles and is robust to additive disturbance.

In this paper, a control action refers to the selection of an appropriate vehicle steering angle. Because the steering angle is physically limited, the bounds of these angles are not relaxed in our algorithm. Slack variables are incorporated into the  barrier functions for the steering angle and lateral offset constraints in the soft-CILQR solver, thereby enabling the solver to compute stabilizing control inputs when nonsmooth control actions are caused by disturbances. Specifically, the contributions of this study are as follows:
\begin{enumerate}
  \item  A novel soft-CILQR controller is designed by integrating CILQR and a soft-constrained MPC technique to preserve steering angle stability guarantees for the lane-keeping control of  automotive systems under additive disturbances.
  \item The proposed soft-CILQR algorithm was analyzed  and validated by solving corresponding optimization problems and conducting numerical simulations and vision-based autonomous driving experiments in TORCS environments with some challenging scenarios.
\end{enumerate}

The remainder of the paper is organized as follows. The adopted path-tracking dynamic model is introduced in Section II. The proposed soft-CILQR algorithm is described in Section III. The experimental results and relevant discussion are provided in Section IV. Finally, the conclusions are presented in Section V.

\section{Dynamic Lane-Keeping Model}
The objective of this study was to develop a rapid lane-keeping (path-tracking) controller  that can ensure the control stability of autonomous vehicles. The path-tracking model \cite{Li19, Mat19, Lee19, Raj11} adopted in this study describes ego vehicle states by using the lateral offset  (the distance of the ego vehicle from the lane centerline, denoted as $\Delta $), the heading angle (the orientation error between the car's yaw angle and lane direction, denoted as $\theta $), and the derivatives of $\Delta$ and $\theta$. This model is introduced in this section.

Consider an ego car with speed along the longitudinal (heading) direction of $v_x$ and whose motion in the lateral direction has two degrees of freedom: the lateral position $y$ and yaw angle $\psi $. In this case, the lateral tire forces acting on the front ($f$) and real ($r$) wheels of the car can be expressed as follows:
\begin{subequations}
\begin{equation}
 F_{yf}  = 2C_{\alpha f} \left( {\delta  - \theta _{vf} } \right),
\end{equation}
\begin{equation}
F_{yr}  = 2C_{\alpha r} \left( { - \theta _{vr} } \right),
\end{equation}
\end{subequations}
where $C_{\alpha f}$ and $C_{\alpha r}$ are the cornering stiffness values of the front and real tires, respectively; $\delta$ is the steering angle; and ${\theta _{vf} }$ and ${\theta _{vr} }$ are the velocity angles of the front and rear tires, respectively. The steering angle $\delta$ is the average of the two front-wheel steering angles, and the distances traveled by these wheels are different. The small-angle approximation method can be used to calculate ${\theta _{vf} }$ and ${\theta _{vr} }$ as follows:
\begin{subequations}
\begin{equation}
\theta _{vf}  = \frac{{\dot y + l_f \dot \psi }}{{v_x }},
\end{equation}
\begin{equation}
\theta _{vr}  = \frac{{\dot y - l_r \dot \psi }}{{v_x }},
\end{equation}
\end{subequations}
where $l_f$ and $l_r$ are the distances from the center of gravity of the vehicle to the front and rear tires, respectively. Newton's second law of motion and a dynamic yaw equation can be used to represent $F_{yf}$ and $F_{yr} $ as follows:
\begin{equation}
F_{yf}  + F_{yr}  = m\left( {\ddot y + v_x \dot \psi  } \right)
\end{equation}
and 
\begin{equation}
l_f F_{yf}  - l_r F_{yr}  = I_z \ddot \psi, 
\end{equation}
where $I_z$ is the moment of inertia along the $z$-axis. The road bank angle is ignored in the preceding equations.

To solve the vehicle path-tracking problem, the ego car's $\Delta$ and $\theta$ are defined as follows:
\begin{subequations}
\begin{equation}
\ddot \Delta  = \ddot y + v_x \left( {\dot \psi  - \dot \psi _{des} } \right),
\end{equation}
\begin{equation}
\theta  = \psi  - \psi _{des}, 
\end{equation}
\end{subequations}
where $\dot \psi _{des}  = \kappa v_x $ is the desired yaw rate and $\kappa $ is the road curvature. The dynamic path-tracking model can be expressed as follows after combining Eqs. (3)\textendash(5):
\begin{subequations}
\begin{align}
\begin{split}
\ddot \Delta  = & - \frac{{2C_{\alpha f}  + 2C_{\alpha r} }}{{mv_x }}\dot \Delta  + \frac{{2C_{\alpha f}  + 2C_{\alpha r} }}{m}\theta  \\
&- \frac{{2l_f C_{\alpha f}  - 2l_r C_{\alpha r} }}{{mv_x }}\dot \theta  + \frac{{2C_{\alpha f} }}{m}\delta ,
\end{split}
\end{align} 
\begin{align}
\begin{split}
\ddot \theta  = & - \frac{{2l_f C_{\alpha f}  - 2l_r C_{\alpha r} }}{{I_z v_x }}\dot \Delta  + \frac{{2l_f C_{\alpha f}  - 2l_r C_{\alpha r} }}{{I_z }}\theta  \\
&- \frac{{2l_f^2 C_{\alpha f}  + 2l_r^2 C_{\alpha r} }}{{I_z v_x }}\dot \theta  + \frac{{2l_f C_{\alpha f} }}{{I_z }}\delta ,
\end{split}
\end{align} 
\end{subequations}
where $v_x$ is assumed to be constant and $\kappa $ is assumed to be small. To solve the aforementioned  continuous-time  model in an MPC framework, a tractable Euler integration scheme using zero-order hold (ZOH) sampling can be employed to discretize the model.  The resulting discretized states are expressed as follows: 
\begin{subequations}
\begin{equation}
\Delta _{i + 1}  = \Delta _i  + \dot \Delta _i dt, 
\end{equation}
\begin{equation}
 \dot \Delta _{i + 1}  = \dot \Delta _i  +\ddot \Delta _i dt ,
\end{equation}
\begin{equation}
\theta _{i + 1}  = \theta _i  + \dot \theta _i dt,
\end{equation}
\begin{equation}
\dot \theta _{i + 1}  = \dot \theta _i  + \ddot \theta _i dt,
\end{equation}
\end{subequations}
where $dt$ is the discretization size and $i$ is its associated index. ZOH is commonly used in commercial microprocessors with limited memory because it does not require previous sample data (i.e. discrete states with time indexes smaller than $i$) \cite{Lew92}. Thus, the following linear discrete-time path-tracking  model is obtained \cite{Lee19}:
\begin{equation}
{\bf x}_{i + 1}  \equiv {\bf f}\left( {{\bf x}_i ,{\bf u}_i } \right) = {\bf Ax}_i  + {\bf Bu}_i,
\end{equation}
where the state vector and control input are defined as $
{\bf x} \equiv  \left[ {\begin{array}{*{20}c}
   \Delta  & {\dot \Delta } & \theta  & {\dot \theta }  \\
\end{array}} \right]^T
$ and ${\bf u} \equiv  \left[  \delta  \right]$, respectively. Moreover, the model matrices are expressed as follows: 
\[
{\bf A} = \left[ {\begin{array}{*{20}c}
{\alpha _{11} } & {\alpha _{12} } & 0 & 0 \\
0 & {\alpha _{22} } & {\alpha _{23} } & {\alpha _{24} } \\
0 & 0 & {\alpha _{33} } & {\alpha _{34} } \\
0 & {\alpha _{42} } & {\alpha _{43} } & {\alpha _{44} } \\
\end{array}} \right],\quad{\bf B} = \left[ {\begin{array}{*{20}c}
0 \\
{\beta _1 } \\
0 \\
{\beta _2 } \\
\end{array}} \right].
\]
The matrix coefficients are expressed as follows:
\[
\begin{array}{l}
\alpha _{11} = \alpha _{33} = 1, \quad\alpha _{12} = \alpha _{34} = dt, \\
\alpha _{22}  = 1 - \frac{{\left( {2C_{\alpha f}  + 2C_{\alpha r} } \right)dt}}{{mv_x }},\quad\alpha _{23}  = \frac{{\left( {2C_{\alpha f}  + 2C_{\alpha r} } \right)dt}}{m}, \\
\alpha _{24}  =  - \frac{{\left( {2l_f C_{\alpha f}  - 2l_r C_{\alpha r} } \right)dt}}{{mv_x }},\quad\alpha _{42}  =  - \frac{{\left( {2l_f C_{\alpha f}  - 2l_r C_{\alpha r} } \right)dt}}{{I_z v_x }} , \\
\alpha _{43}  = \frac{{\left( {2l_f C_{\alpha f}  - 2l_r C_{\alpha r} } \right)dt}}{{I_z }},\quad\alpha _{44}  = 1 - \frac{{\left( {2l_f^2 C_{\alpha f}  + 2l_r^2 C_{\alpha r} } \right)dt}}{{I_z v_x }}, \\
\beta _1 = {\frac{{2C_{\alpha f} dt}}{m}} ,\quad\beta _2  = \frac{{2l_f C_{\alpha f} dt}}{{I_z }}. \\
\end{array}
\]
The model and vehicle parameters used in this study are listed in Table I.

\begin{table}[!t]
\caption{Values of the model and vehicle parameters used in this study}
\begin{center}
\begin{tabular}{llll}
\hline
$dt$ &  0.01 s & $l_f$  & 1.27 m  \\
$v_x$ & \{16.6, 20.0, 22.2\} m/s  &  $l_r$ & 1.37 m  \\
$m$ &  1150 kg & ${C_{\alpha f} }$  & 80000 N/rad  \\
$I_z$ &   2000 kgm$^{2}$ &  ${C_{\alpha r} }$ &  80000 N/rad \\\hline
\end{tabular}
\end{center}
\end{table}

\section{Soft-CILQR Algorithm}
In this section, the soft-constrained MPC problem considered in this study is formulated on the basis of the model introduced in the previous section. The relevant constraints are then transformed into  barrier functions, and the resulting unconstrained problem is solved using the proposed  soft-CILQR algorithm. Subsequently, the upper bound of the prediction horizon of the examined MPC problem is calculated. Finally,  a classical soft-MPC scheme with the classical terminal cost and without considering the terminal dynamics of state and slack variables is formulated for comparison.

\subsection{Problem Formulation}
In principle, the infinite-horizon MPC problem should be solved by minimizing the infinite-dimensional cost function; however, this is computationally intractable in practice \cite{Ask17}. Nevertheless, the dual-mode prediction paradigm control strategy enables the infinite-dimensional optimal MPC to be specified as a finite-horizon optimization problem. In this approach, the prediction horizon is divided into stage and terminal modes. The stage mode involves the prediction of variables over the first $N$ time steps, whereas the terminal mode involves the prediction of variables in the subsequent steps. In the terminal mode, predicted control inputs are usually specified by the unconstrained optimal feedback law, and the integer $\bar N$ denotes the upper bound of the prediction horizon. The rationale behind the use of an unconstrained feedback law in the terminal mode is that the constrained optimal sequence coincide with the unconstrained optimal feedback law at long times at which zero state-control pairs are simultaneously feasible for constrained and unconstrained schemes if the initial conditions of the unconstrained scheme are such that it satisfies the constraints of the constrained scheme \cite{Kou16}.

In particular, the slack vector and associated terminal dynamics proposed in \cite{Svr23} for linear systems is incorporated into the dual-mode MPC in this study, forming the considered soft-constrained MPC problem. Thus, the optimization variables in the considered control problem are a sequence of states ${\bf X} \equiv \left\{ {{\bf x}_0 ,{\bf x}_1 ,...,{\bf x}_N } \right\}$,  a control sequence ${\bf U} \equiv \left\{ {{\bf u}_0 ,{\bf u}_1 ,...,{\bf u}_{N - 1} } \right\}$,  and a slack sequence ${\bf E} \equiv \left\{ {{\bf e}_0 ,{\bf e}_1 ,...,{\bf e}_N } \right\}$. The problem is then formulated as follows (Problem 1):

\textit{Problem 1}:
\begin{subequations}
\begin{equation}
\mathop {{\rm min}}\limits_{{\bf X}, {\bf U}, {\bf E}} J = J_s  + J_t ,
\end{equation}
\begin{equation}
J_s  = \sum\limits_{i = 0}^{N - 1} {{\bf x}_i^T {\bf Q} {{\bf x}_i   }  + {\bf u}_i ^T {\bf Ru}_i }  + {\bf e}_i ^T {\bf Se}_i ,
\end{equation}
\begin{equation}
J_t  = \sum\limits_{i = N}^{\bar N } { {\bf x}_i  ^T {\bf P}{{\bf x}_i   } }  + {\bf e}_i ^T {\bf Te}_i 
\end{equation}
\end{subequations}
subject to
\begin{subequations}
\begin{equation}
{\bf x}_{i + 1}  = {\bf Ax}_i  + {\bf Bu}_i ,\quad 0 \le i < N,
\end{equation}
\begin{equation}
{\bf x}_{i + 1}  = \left( {{\bf A} + {\bf B} {\bf K}} \right){\bf x}_i , \quad N \le i < \bar N,
\end{equation}
\begin{equation}
{\bf e}_{i + 1}  = {\bf Me}_i , \quad N \le i < \bar N,
\end{equation}
\begin{equation}
\left| \Delta  \right| \le \bar \Delta \left( {1 + \varepsilon _l } \right) ,
\end{equation}
\begin{equation}
\left| \delta  \right| \le \bar \delta \left( {1 + \varepsilon _s } \right) ,
\end{equation}
\begin{equation}
0 \le \varepsilon _l  \le \bar{\varepsilon_{l}} ,
\end{equation}
\begin{equation}
0 \le \varepsilon _s  \le \bar{\varepsilon_{s}} ,
\end{equation}
\begin{equation}
\left| {\dot \Delta } \right| \le \dot \Delta _{\max } ,
\end{equation}
\begin{equation}
\left| \theta  \right| \le \theta _{\max } ,
\end{equation}
\begin{equation}
\left| {\dot \theta } \right| \le \dot \theta _{\max },
\end{equation}
\end{subequations}
where $J_s$ and $J_t$ denote the stage and terminal cost functions, respectively. The weighting matrixes (${\bf Q}$, ${\bf R}$, ${\bf S}$, ${\bf P}$, and ${\bf T}$) are assumed to be symmetric positive-definite matrices. The terminal dynamics of the state vector ${\bf x}$ and slack vector ${\bf e} \equiv  \left[ {\begin{array}{*{20}c}
   {\varepsilon _l } & {\varepsilon _s }  \\
\end{array}} \right]^T$ are expressed in Eqs. (10b) and (10c), respectively. The matrix ${\bf P}$ is the solution of the discrete-time algebraic Riccati equation, which is expressed as follows:
\begin{equation}
{\bf P} = {\bf A}^T {\bf PA} + {\bf Q} - {\bf A}^T {\bf PB}({\bf B}^T {\bf PB} + {\bf R})^{ - 1} {\bf B}^T {\bf PA}.
\end{equation}
The corresponding feedback gain ${\bf K}$ has the following form:
\begin{equation}
{\bf K} =  - \left( {{\bf B}^T {\bf PB} + {\bf R}} \right)^{ - 1} {\bf B}^T {\bf PA}, 
\end{equation}
where ${\bf P}$ also satisfies the Lyapunov equation   
\begin{equation}
{\bf P} = \left( {{\bf A} + {\bf BK}} \right)^T {\bf P}\left( {{\bf A} + {\bf BK}} \right) + {\bf Q} + {\bf K}^T {\bf RK},
\end{equation}
demonstrating that the system (10b) is stable.

The following assumptions are made: ${\bf S} = S{\bf I}_2$, ${\bf T} = T{\bf I}_2$, and ${\bf M} = M{\bf I}_2$. The scalars $S$, $M\in \left [ 0,1 \right )$, and $T=S  /(1-M^2 )$ are  designed to induce desirable asymptotic behaviors of the relaxed constraints \cite{Svr23}. For the inequality constraints expressed in Eqs. (10d) and (10e), $\bar \Delta   \equiv  {{\Delta _{\max } } \mathord{\left/
 {\vphantom {{\Delta _{\max } } {\left( {1 + \bar{\varepsilon_{l}} } \right)}}} \right. \kern-\nulldelimiterspace} {\left( {1 + \bar{\varepsilon_{l}} } \right)}}$, and
${{\bar \delta   \equiv  \delta _{\max } } \mathord{\left/
 {\vphantom {{\bar \delta  = \delta _{\max } } {\left( {1 + \bar{\varepsilon_{s}} } \right)}}} \right.
 \kern-\nulldelimiterspace} {\left( {1 + \bar{\varepsilon_{s}} } \right)}}$; moreover, $\Delta _{\max}$  and $\delta _{\max}$ are the physical limitations of  $\Delta $  and $\delta$, respectively. The nonnegative slack variables $\varepsilon _{l}$ and $\varepsilon _{s}$ have the upper bounds $\bar{\varepsilon_{l}}$   and $\bar{\varepsilon_{s}}$, respectively, in Eqs. (10f) and (10g). Thus, the values of  $\Delta $ and $\delta $  are restricted to their bounds $\pm \Delta _{\max } $ and $ \pm \delta _{\max } $, respectively, in the optimization process, even if  slack variables are incorporated into this process. Finally, Eqs. (10h), (10i), and (10j) represent the constraints for the state variables ${\dot \Delta }$, $\theta $, and ${\dot \theta }$, respectively, with corresponding upper bounds denoted as ${\dot \Delta }_{\max}$, $\theta _{\max}$, and ${\dot \theta }_{\max}$. The control-related parameters and constraints in this study are presented in Table II.

\begin{table}[!t]
\caption{Control parameter values and constraints}
\begin{center}
\begin{tabular}{llll}
\hline
${\bf Q}$ & diag(20, 1, 20, 1) & $\Delta _{\max } $ & 2.0 m \\
${\bf R}$ & [60] & $\dot \Delta _{\max } $ & 5.0 m/s \\
 $(q_{l1} ,q_{l2})$ & (5, 1) & $\theta _{\max } $ & $\pi  /2$ rad \\
$(q_{s1} ,q_{s2})$ & (80, 1) & $\dot \theta _{\max } $ & 0.5 rad/s \\
$S$ & \{0.01, 0.5\}  & $\delta _{\max } $ & $\pi  /6$ rad\\
$M$ & 0.9  & $\varepsilon _{\max}$ &  \{19,..., 99\}  \\\hline
\end{tabular}
\end{center}
\end{table}

\subsection{Trajectory Optimizer}
Barrier functions that ensure constraint satisfaction can be incorporated into Problem 1 to convert this problem into an effectively unconstrained optimization problem (i.e., Problem 2). Herein, constraints are modeled using exponential barrier functions, which are strictly convex and smooth  functions \cite{Che17, Liu24, Fel14}. Accordingly, the explicit form of Problem 2 can be expressed as follows:

\textit{Problem 2}:
\begin{subequations}
\begin{equation}
\mathop {{\rm min}}\limits_{{\bf X}, {\bf U}, {\bf E}}{\tilde J} = J  + J_{e}  + J_{la}  + J_{st} + J_{th},
\end{equation}
\begin{equation}
J_e  = \sum\limits_{k = l,s} {\sum\limits_{i = 0}^N {\exp \left( { - \varepsilon _{k,i} } \right) + \exp \left( {\varepsilon _{k,i}  - \bar{\varepsilon_{k}} } \right)} } ,
\end{equation}
\begin{align}
\begin{split}
J_{la}  = \sum\limits_{i = 0}^N &{q_{l1} \exp \left\{ {q_{l2} \left[ { - \bar \Delta \left( {1 + \varepsilon _{l,i} } \right) - \Delta _i } \right]} \right\}} \\
& + q_{l1} \exp \left\{ {q_{l2} \left[ {\Delta _i  - \bar \Delta \left( {1 + \varepsilon _{l,i} } \right)} \right]} \right\},
\end{split}
\end{align}
\begin{align}
\begin{split}
J_{st}  = \sum\limits_{i = 0}^{N - 1}& {q_{s1} \exp \left\{ {q_{s2} \left[ { - \bar \delta \left( {1 + \varepsilon _{s,i} } \right) - \delta _i } \right]} \right\}} \\
& + q_{s1} \exp \left\{ {q_{s2} \left[ {\delta _i  - \bar \delta \left( {1 + \varepsilon _{s,i} } \right)} \right]} \right\},
\end{split}
\end{align}  
\begin{equation}
J_{th}  = \sum\limits_{k = 1}^3 {\sum\limits_{i = 0}^N {\exp \left(  { - x_{k,\max }  - x_{k,i} } \right) + \exp \left( {x_{k,i}  - x_{k,\max } } \right)} }
\end{equation}
\end{subequations}
subject to 
\begin{subequations}
\begin{equation}
{\bf x}_{i + 1}  = {\bf Ax}_i  + {\bf Bu}_i ,\quad 0 \le i < N,
\end{equation}
\begin{equation}
{\bf x}_{i + 1}  = \left( {{\bf A} + {\bf B} {\bf K}} \right){\bf x}_i , \quad N \le i < \bar N,
\end{equation}
\begin{equation}
{\bf e}_{i + 1}  = {\bf Me}_i , \quad N \le i < \bar N,
\end{equation}
\end{subequations}
where $q_{s1,2}$ and $q_{l1,2}$ are barrier function parameters. Moreover, $x_1$, $x_2$, and $x_3$ represent ${\dot \Delta }$,  $\theta $, and ${\dot \theta }$, respectively. Herein, $\bar{\varepsilon_{l}} = \bar{\varepsilon_{s}} \equiv \varepsilon _{\max}$ is assumed for simplicity.

Achieving smooth steering inputs of perturbed systems is the main goal of this study. To highlight the uniqueness of the proposed approach for accelerating general methods, a continuous cost \cite{Liu24} ensuring the continuity of control input trajectories, denoted as $J_{cc}$, is considered for comparison. This cost term functions as a penalty for changes of the iterative steering angle solutions and can be expressed as follows:
\begin{align}
\begin{split}
J_{cc}  = \sum\limits_{i = 0}^{N - 1}& {q_{c1} \exp \left\{ {q_{c2} \left[ { \hat\delta _i - \delta _i } \right]} \right\}} \\
& + q_{c1} \exp \left\{ {q_{c2} \left[ {\delta _i  - \hat \delta _i} \right]} \right\},
\end{split}
\end{align}  
where $\hat \delta $ is the steering angle obtained in the previous iteration. The parameters $q_{c1}$ and $q_{c2}$ are the penalty parameters that affect the vehicle agility in the yaw direction. Increasing the magnitudes of $q_{c1}$ and $q_{c2}$ would cause the solver to penalize steering input changes more severely, decreasing the vehicle agility.

To solve  Problem 2, the ILQR algorithm is  employed to compute  updated state and control variables. The ILQR solver performs alternating backward and forward propagation steps to iteratively optimize the control strategy. In the backward propagation procedure, the control gains ${\bf \tilde K}_i $ and ${\bf \tilde k}_i $ are estimated using the analytical expressions of the expansion coefficients of the perturbed value function presented in \cite{ Tas12, Tas14}. The improved  state and control sequences can then be obtained in a forward propagation process as follows:
\begin{subequations}
\begin{equation}
{\bf \hat u}_i  = {\bf u}_i  + {\bf \tilde k}_i  + {\bf \tilde K}_i \left( {{\bf \hat x}_i  - {\bf x}_i } \right),
\end{equation}
\begin{equation}
{\bf \hat x}_{i + 1}  = {\bf f}\left( {{\bf \hat x}_i ,{\bf \hat u}_i } \right),
\end{equation}
\end{subequations}
where  ${\bf \hat x}_0  = {\bf x}_0 $. Moreover, herein, elements of the slack sequence are designed to be updated using the Newton descent method. Studies have indicated that a Newton-based optimizer can rapidly decay the cost function of barrier-function-based MPC problems \cite{Hau06,Fel17b}. The relevant update law is expressed as follows:
\begin{equation}
{\bf \hat e}_i  = {\bf e}_i  - \left[ {{\bf H}\tilde J\left( {{\bf e}_i } \right)} \right]^{ - 1} \nabla \tilde J\left( {{\bf e}_i } \right),
\end{equation}
where the Hessian matrix ${\bf H}\tilde J\left( {{\bf e}_i } \right) \succ 0$. The aforementioned procedures for the computation of the optimal state, control, and slack variables constitute the basis of the proposed soft-CILQR optimizer.

In summary, the main innovative aspect of the soft-CILQR algorithm is that it incorporates slack variables into the framework of the CILQR algorithm. This constraint-softening approach facilitates the achievement of control stability while preserving the high computational efficiency of the CILQR algorithm in real-time applications. The method used in the soft-CILQR algorithm to calculate the upper bound of the prediction horizon ${\bar N}$ is presented in the following section. Note that, the continuous cost $J_{cc}$ (16) is not included in Problem 2 in this study. The CILQR algorithm incorporating $J_{cc}$ is denoted as CILQR-cc throughout the paper.

\subsection{Computation of the Upper Bound of the Prediction Horizon}
As described in the previous  section, the prediction horizon of Problem 1/2 is  divided into two operating modes: the stage and terminal modes. To ensure that the predictions made in the terminal mode satisfy the state, control, and  slack constraints, the end state  of the stage mode should lie in a positively invariant set $\chi _T $ under the dynamics  defined by the state feedback control law ${\bf u} = {\bf Kx}$, Eq. (10b), and Eq. (10c) with the following mixed constraint for the augmented state ${\bf \tilde x} \equiv  \left[ {\begin{array}{*{20}c}
   {\bf x} & {\bf e}  \\
\end{array}} \right]^T$ \cite{Svr23}: 
\begin{equation}
{\bf \tilde F \tilde x } + {\bf \tilde Gu}  \le {\bf \tilde h}.
\end{equation}
The corresponding constraint matrices are expressed as follows:
\[
{\bf \tilde F}  = \left( {\begin{array}{*{20}c}
   { - 1} & 0 & 0 & 0 & 0 & 0  \\
   1 & 0 & 0 & 0 & 0 & 0  \\
   0 & { - 1} & 0 & 0 & 0 & 0  \\
   0 & 1 & 0 & 0 & 0 & 0  \\
   0 & 0 & { - 1} & 0 & 0 & 0  \\
   0 & 0 & 1 & 0 & 0 & 0  \\
   0 & 0 & 0 & { - 1} & 0 & 0  \\
   0 & 0 & 0 & 1 & 0 & 0  \\
   0 & 0 & 0 & 0 & 0 & 0  \\
   0 & 0 & 0 & 0 & 0 & 0  \\
   0 & 0 & 0 & 0 & { - 1} & 0  \\
   0 & 0 & 0 & 0 & 1 & 0  \\
   0 & 0 & 0 & 0 & 0 & { - 1}  \\
   0 & 0 & 0 & 0 & 0 & 1  \\
   { - 1} & 0 & 0 & 0 & { - \bar \Delta } & 0  \\
   1 & 0 & 0 & 0 & { - \bar \Delta } & 0  \\
   {-K_0} & {-K_1} & {-K_2} & {-K_3} & 0 & { - \bar \delta }  \\
   {K_0} & {K_1} & {K_2} & {K_3} & 0 & { - \bar \delta }  \\
\end{array}} \right),
\]
\[
{\bf \tilde G} = \left( {\begin{array}{*{20}c}
   0  \\
   0  \\
   0  \\
   0  \\
   0  \\
   0  \\
   0  \\
   0  \\
   { - 1}  \\
   { 1}  \\
   0  \\
   0  \\
   0  \\
   0  \\
   0  \\
   0  \\
   0  \\
   0  \\
\end{array}} \right), \quad {\bf \tilde h} = \left( {\begin{array}{*{20}c}
   {\Delta _{\max } }  \\
   {\Delta _{\max } }  \\
   {\dot \Delta _{\max } }  \\
   {\dot \Delta _{\max } }  \\
   {\theta _{\max } }  \\
   {\theta _{\max } }  \\
   {\dot \theta _{\max } }  \\
   {\dot \theta _{\max } }  \\
   {\delta _{\max } }  \\
   {\delta _{\max } }  \\
   0  \\
   {\varepsilon _{\max} }  \\
   0  \\
   {\varepsilon _{\max} }  \\
   {\bar \Delta }  \\
   {\bar \Delta }  \\
   {\bar \delta }  \\
   {\bar \delta }  \\
\end{array}} \right),
\]
where $K_{0,1,2,3} $ are  components  of $\bf  K$. To increase the practical utility of the examined control problem, $\chi _T$ should be selected as the maximal positively invariant (MPI) set. The MPI set, denoted as $\chi _{MPI} $, is a polytope that can be expressed  as follows \cite{Kou16}:
\begin{equation}
\chi _{MPI}  \equiv \left\{ {{\bf \tilde x}:({\bf \tilde F} + {\bf \tilde G \tilde K})({\bf \tilde A} + {\bf \tilde B \tilde K})^i{\bf \tilde x} \le {\bf \tilde h}, \quad i = 0,...,N_\nu  } \right\} .
\end{equation}
The corresponding augmented state and control matrices are expressed as follows:
\[
{\bf \tilde A}  = \left( {\begin{array}{*{20}c}
   {\alpha _{11} } & {\alpha _{12} } & 0 & 0 & 0 & 0  \\
   0 & {\alpha _{22} } & {\alpha _{23} } & {\alpha _{24} } & 0 & 0  \\
   0 & 0 & {\alpha _{33} } & {\alpha _{34} } & 0 & 0  \\
   0 & {\alpha _{42} } & {\alpha _{43} } & {\alpha _{44} } & 0 & 0  \\
   0 & 0 & 0 & 0 & M & 0  \\
   0 & 0 & 0 & 0 & 0 & M  \\
\end{array}} \right),
\]
\[
{\bf \tilde B} = \left( {\begin{array}{*{20}c}
   0  \\
   {\beta _1 }  \\
   0  \\
   {\beta _2 }  \\
   0  \\
   0  \\
\end{array}} \right), \quad {\bf \tilde K} = \left( {\begin{array}{*{20}c}
   {K_0 }  \\
   {K_1 }  \\
   {K_2 }  \\
   {K_3 }  \\
   0  \\
   0  \\
\end{array}} \right)^T.
\]
The relationship among the smallest positive integer $N_\nu$, $N$, and $\bar N$ in Problem 1/2 can be expressed as follows: $\bar N =  N + N_\nu +1$. The smallest integer can be obtained by solving the following linear programming problem:
\begin{equation}
\mathop {{\rm max}}\limits_{\bf \tilde x} \left( {{\bf \tilde F} + {\bf \tilde G \tilde K}} \right)_j \left( {{\bf \tilde A} + {\bf \tilde B \tilde K}} \right)^{n + 1} {\bf \tilde x}, \quad j = 1,...,n_c 
\end{equation}
subject to
\begin{equation}
\left( {{\bf \tilde F} + {\bf \tilde G \tilde K}} \right)\left( {{\bf \tilde A} + {\bf \tilde B \tilde K}} \right)^i {\bf \tilde x} \le {\bf \tilde h}, \quad i = 0,...,n,
\end{equation}
where  $n_c$ is the row number in the matrix (${{\bf \tilde F} + {\bf \tilde G \tilde K}}$) and $n$ = 1,..., $N_\nu $. On the basis of the parameters and constraints defined in Tables I and II, the values of $N_\nu$ can be calculated for a system with $\varepsilon _{\max}$ = 19, 29, 39, 49, 59, 69, 79, 89, and 99 (Fig. 1). The calculation results indicate a nearly linear relationship between $N_\nu$ and $\varepsilon _{\max}$. Fig. 2 illustrates a visualization of $\chi _{MPI} $ for $\varepsilon _{\max}$ = 19, 29, and 49. The set $\chi _{MPI} $ becomes smaller as $\varepsilon _{\max}$ increases, and all the obtained $\chi _{MPI} $ sets fully adhere to the constraints defined in Table II. This method, which involves using the unconstrained LQR gain for calculating the upper bound of the prediction horizon, can be performed offline and thus does not increase the online computational burden.

\begin{figure}[!t]
\centerline{\includegraphics[scale=0.158]{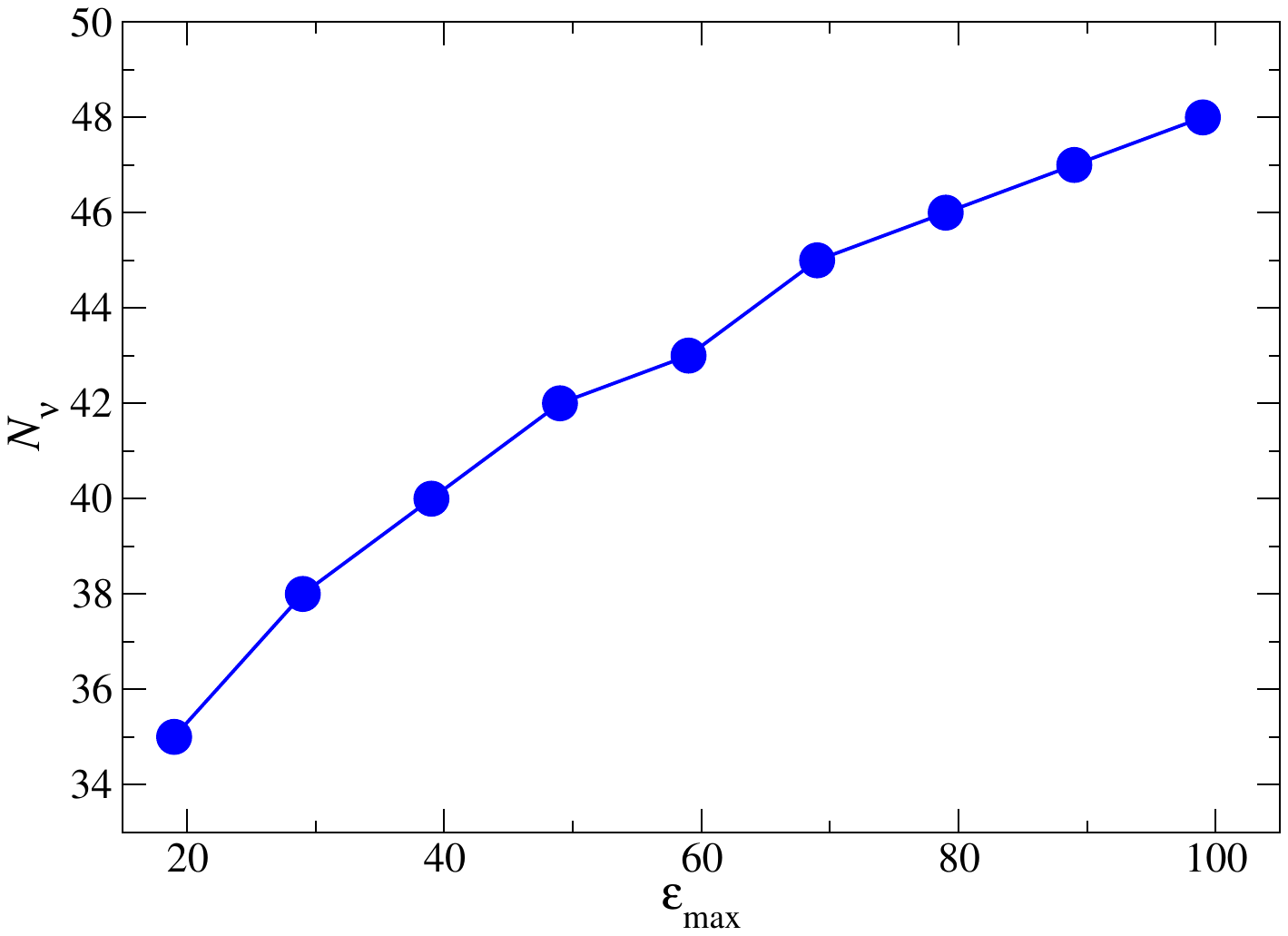}}
\caption{Relationship between $\varepsilon _{\max}$ and the computed $N_\nu$ value when $v_x$ = 20.0 m/s.} 
\end{figure}

\begin{figure}[!t]
{\includegraphics*[width=0.5\linewidth]{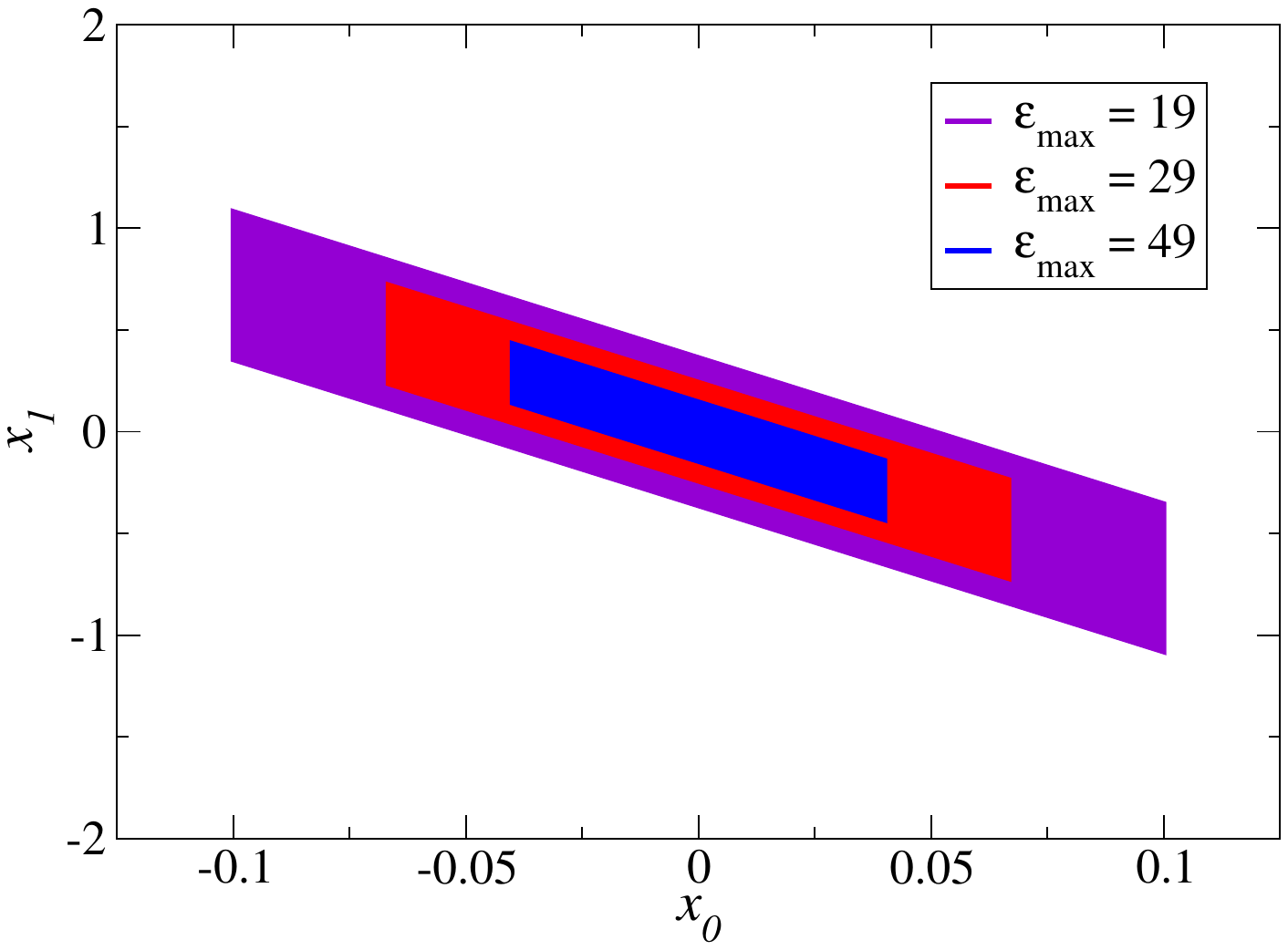}}%
{\includegraphics*[width=0.5\linewidth]{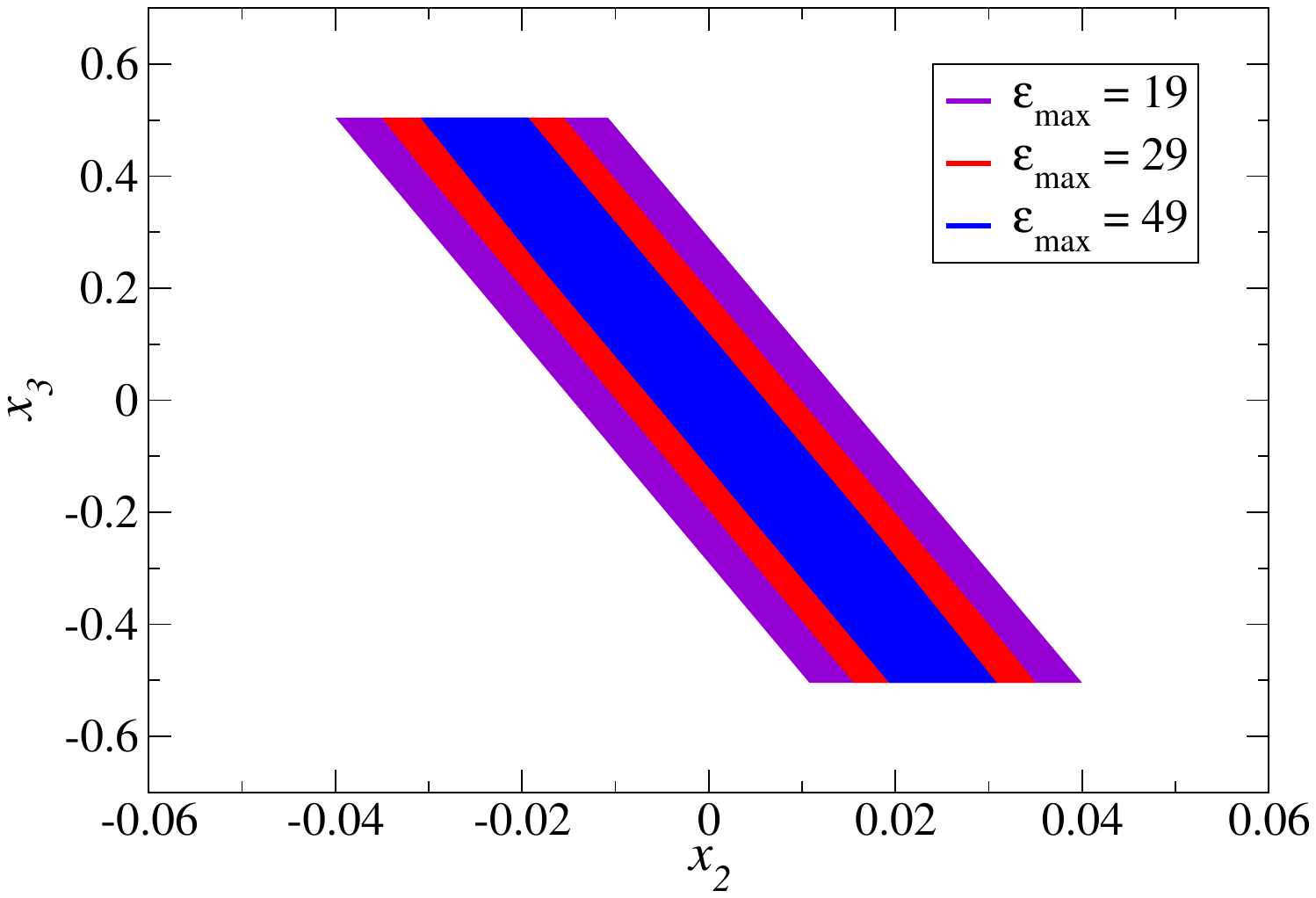}}%
\caption{Visualization of $\chi _{MPI} $ for subsystems ($x_0$, $x_1$) = (${\Delta }$, ${\dot \Delta }$) and ($x_2$, $x_3$) = ($\theta $, ${\dot \theta }$) under $\varepsilon _{\max}$ values of 19, 29, and 49.}
\end{figure}

\subsection{Classical soft-MPC  Scheme}
This section details the classical soft-MPC algorithm. Here, the term “classical soft-MPC” refers to the traditional MPC approach with the same soft constraints as those of soft-CILQR  and classical terminal penalty terms. The associated problem is formulated as follows (Problem 3):

\textit{Problem 3}:
\begin{subequations}
\begin{equation}
\mathop {\min }\limits_{{\bf X},{\bf U},{\bf E}} J' = J'_s  + J'_t ,
\end{equation}
\begin{equation}
J'_s  = \sum\limits_{i = 0}^{N - 1} {{\bf x}_i^T {\bf Q} {{\bf x}_i   }  + {\bf u}_i ^T {\bf Ru}_i }  + {\bf e}_i ^T {\bf Se}_i ,
\end{equation}
\begin{equation}
J'_t  = {\bf x}_N^T {\bf Px}_N  + {\bf e}_N^T {\bf Te}_N  
\end{equation}
\end{subequations}
subject to
\begin{subequations}
\begin{equation}
{\bf x}_{i + 1}  = {\bf Ax}_i  + {\bf Bu}_i ,\quad 0 \le i < N,
\end{equation}
\begin{equation}
- \bar \Delta  \le \frac{\Delta }{{1 + \varepsilon _l }} \le \bar \Delta ,
\end{equation}
\begin{equation}
 - \bar \delta  \le \frac{\delta }{{1 + \varepsilon _s }} \le \bar \delta ,
\end{equation}
\begin{equation}
{\bf \tilde x}_{\min }  \le {\bf \tilde x} \le {\bf \tilde x}_{\max }, 
\end{equation}
\begin{equation}
 - \delta _{\max }  \le \delta  \le \delta _{\max }. 
\end{equation}
\end{subequations}
This problem is solved in this study by using the open-source code Interior Point Optimizer (IPOPT) \cite{Wac06}, and the solution is compared with the solutions of the soft-CILQR and CILQR solvers. The IPOPT solver was interfaced with CasADi \cite{JAE19}, a general numerical optimization framework. In the following sections (Sec. IV-A and Sec. V), the soft-MPC and MPC solvers refer to the optimizers used for solving Problem 3 for lane-keeping control of automated vehicles with and without including slack variables, respectively.

The following section describes validation experiments in which the proposed soft-CILQR algorithm was compared with other control algorithms, including the CILQR, CILQR-cc, soft-MPC, and MPC algorithms, for the ego vehicle lane-following tasks.

\begin{figure*}[!t]
{\includegraphics*[width=0.33\linewidth]{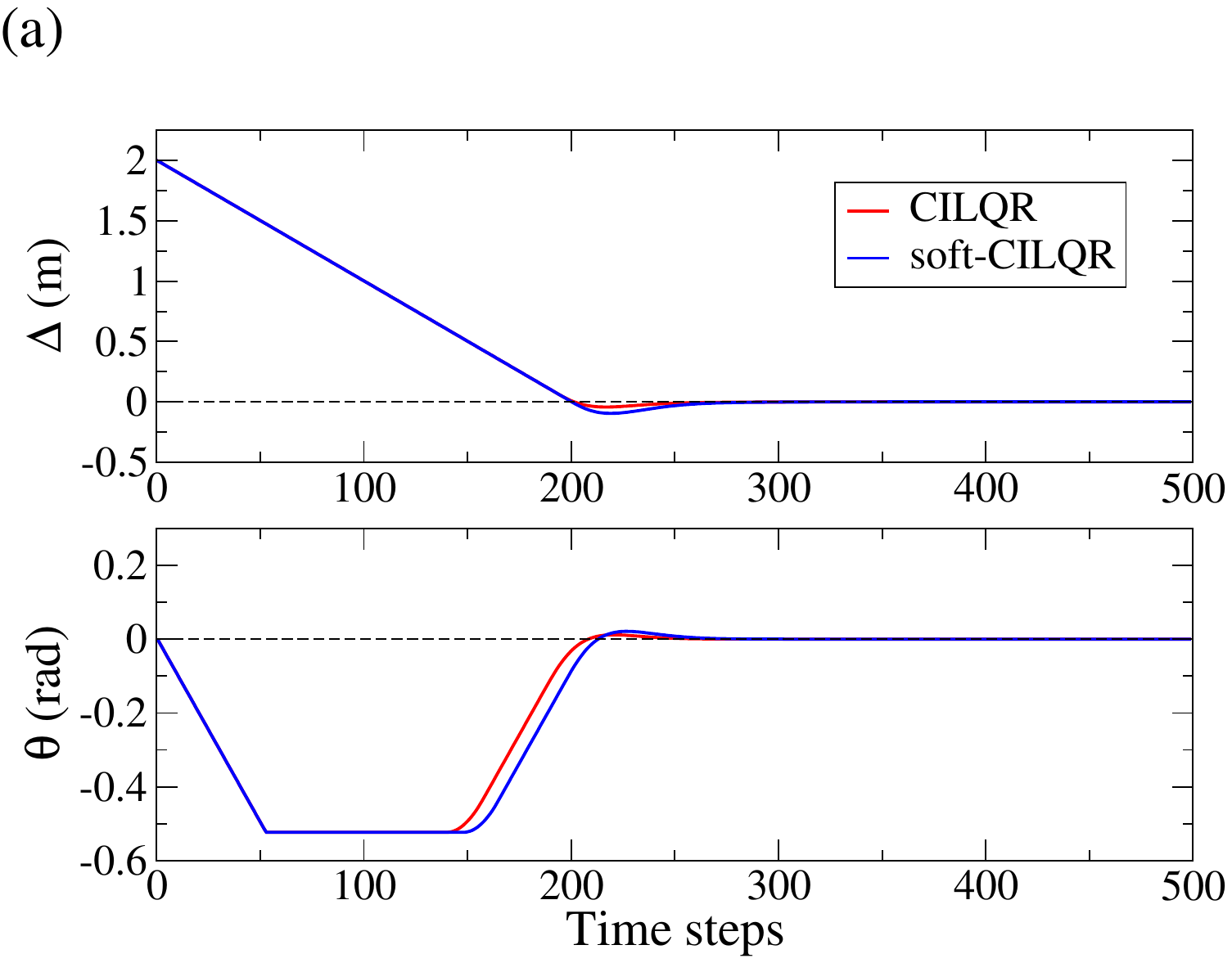}}%
{\includegraphics*[width=0.33\linewidth]{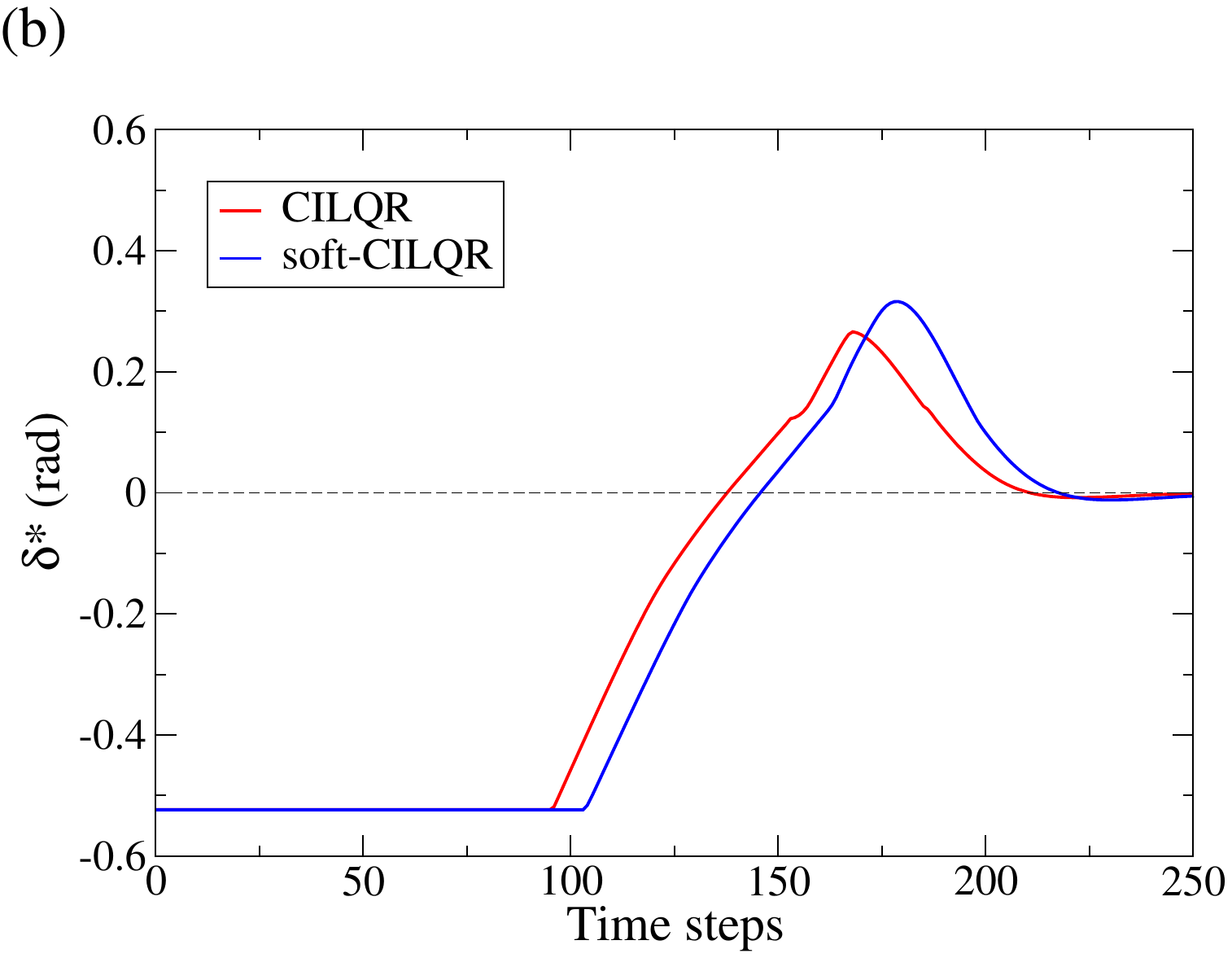}}%
{\includegraphics*[width=0.33\linewidth]{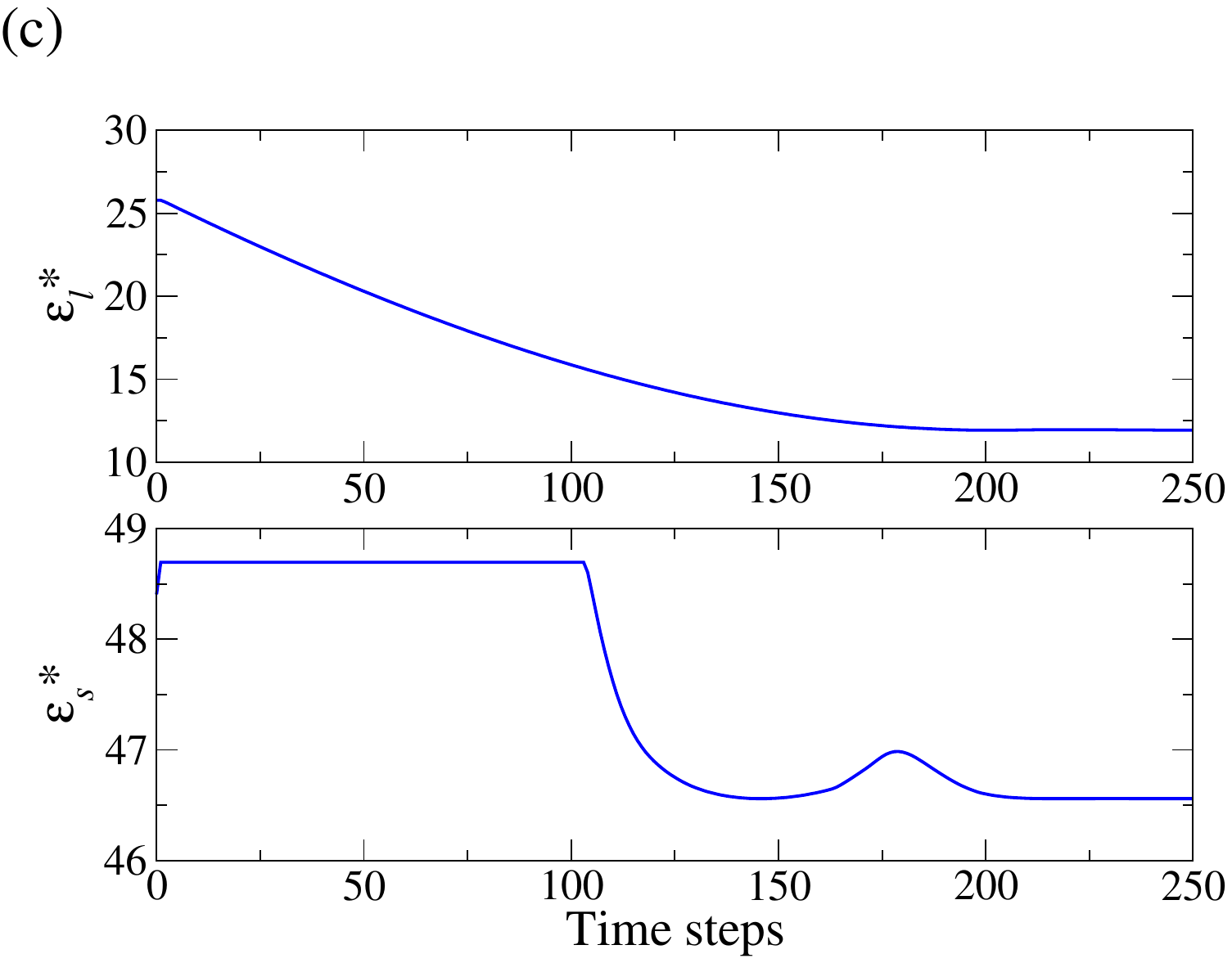}}%
\caption{Numerical simulation results for the CILQR and soft-CILQR solvers obtained when $v_x$ = 20.0 m/s, $S$ = 0.01, $N$ = 40,  $\varepsilon _{\max}$ = 49, and $\sigma $ = 0.0. (a) Trajectories of $\Delta$ and $\theta$. Trajectories of the first component of the optimal (b) steering angle  and (c) slack sequences. Dashed lines represent reference data points (zeros).} 
\end{figure*}

\begin{figure*}[!t]
{\includegraphics*[width=0.33\linewidth]{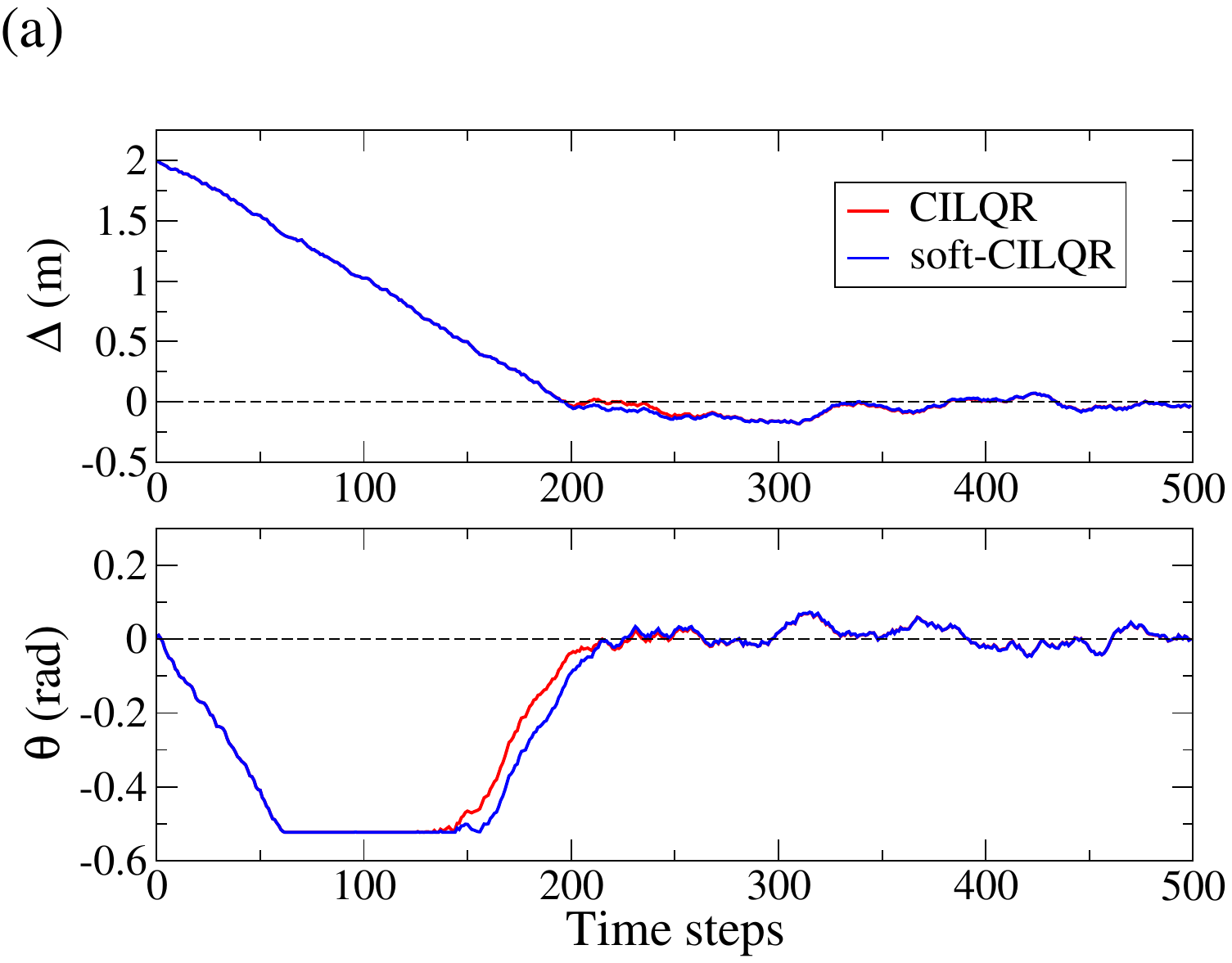}}%
{\includegraphics*[width=0.33\linewidth]{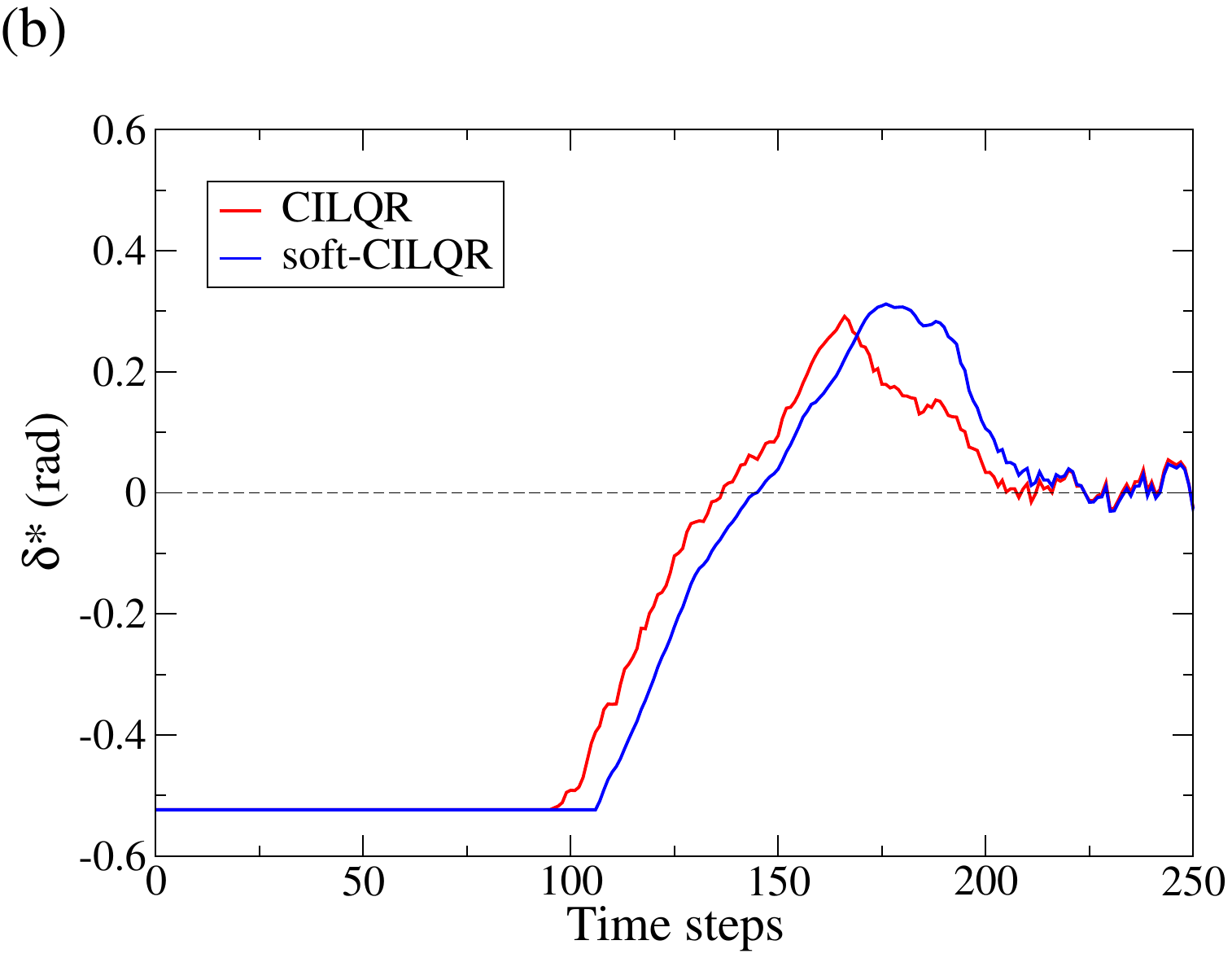}}%
{\includegraphics*[width=0.33\linewidth]{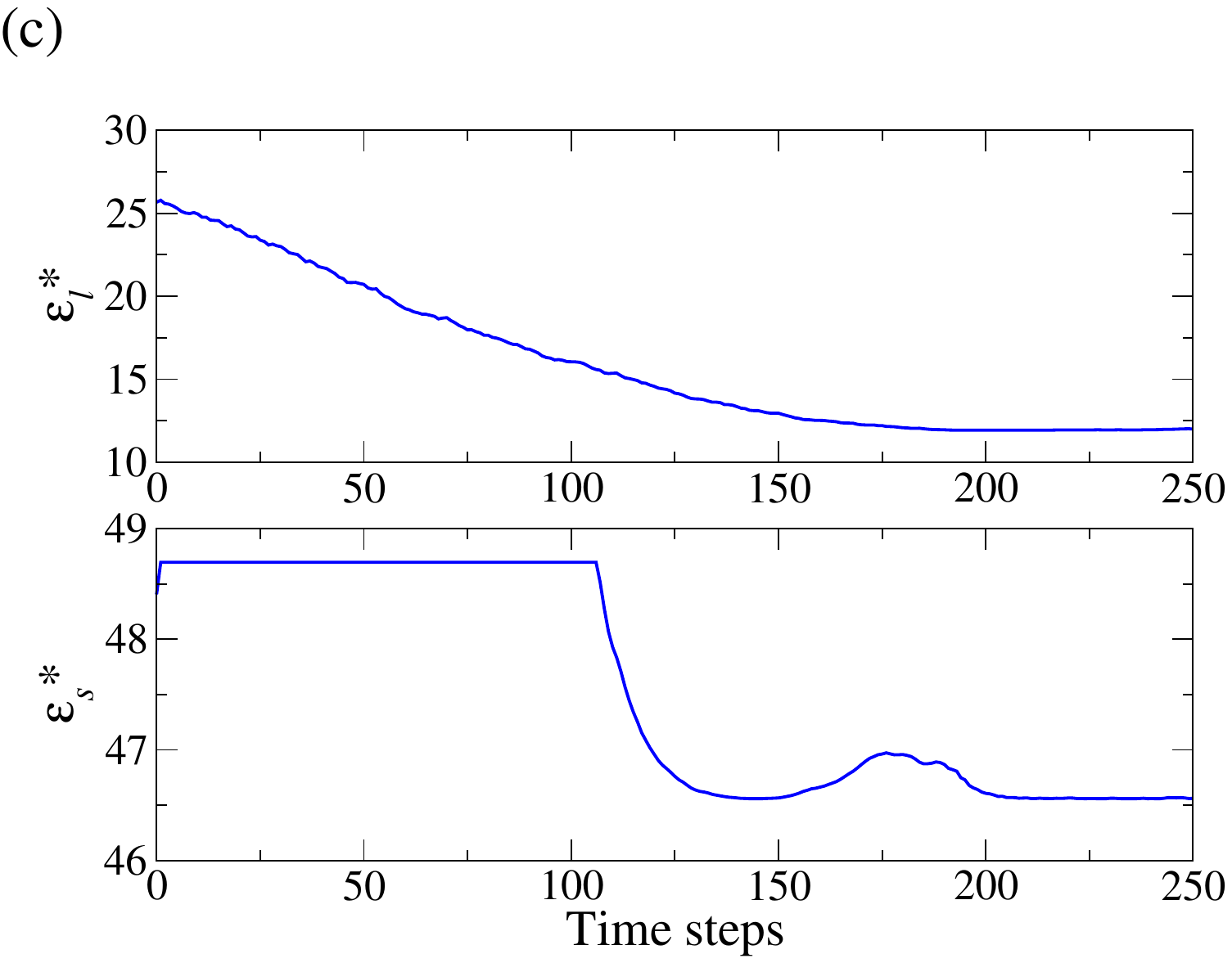}}%
\caption{Numerical simulations results of the CILQR and soft-CILQR solvers obtained when $v_x$ = 20.0 m/s, $S$ = 0.01, $N$ = 40,  $\varepsilon _{\max}$ = 49, and $\sigma $ = 1.0. (a) Trajectories of $\Delta$ and $\theta $. Trajectories of the first component of the optimal (b) steering angle and (c) slack sequences. }
\end{figure*}

\section{Experimental Results and Discussion}
To test the proposed soft-CILQR algorithm, numerical simulations were first conducted to analyze its performance in the presence of  bounded additive disturbances. Second, vision-based vehicle lane-keeping simulations were performed in TORCS to validate the performance of the proposed algorithm in self-driving scenarios. The vision-based experiments were more challenging than the numerical experiments because the vision-based experiments mirrored real-world scenarios and thus had greater uncertainty in the model inputs. The proposed algorithm was evaluated against the CILQR, CILQR-cc, soft-MPC, and MPC algorithms under the same conditions.  All  experiments were conducted on a desktop PC with a 3.6-GHz Intel i9-9900K CPU, an Nvidia RTX 2080 Ti GPU, and 64 GB of RAM.

\subsection{Numerical Simulations}
In the numerical simulations, a vehicle [Eq. (8)] was controlled by the algorithms with the additive disturbance ${\bf w} = \left[ {\begin{array}{*{20}c}
   {w_0 } & {w_1 } & {w_2 } & {w_3 }  \\
\end{array}} \right]^T 
$. The constraints of ${\bf w}$ are expressed as follows: 
\begin{equation}
\begin{array}{l}
 \left| {w_0 } \right| \le 0.013 \quad {\rm m}, \\ 
 \left| {w_1 } \right| \le 0.325 \quad {\rm m/s}, \\ 
 \left| {w_2 } \right| \le 0.010 \quad {\rm rad}, \\ 
 \left| {w_3 } \right| \le 0.170 \quad {\rm rad/s}.\\ 
 \end{array}
\end{equation}
These constraints are similar to those used in \cite{Mat19}, where  ${\bf w}$ represents deviations in the vehicle model [Eq. (8)]. In \cite{Mat19}, ${\bf w}$ was originally related to velocity ($v_x$) differences between real and nominal systems; however, the term is related to the uncertainties of the cornering stiffness values ($C_{\alpha f}$ and $C_{\alpha r}$) because $v_x$, $C_{\alpha f}$, and $C_{\alpha r}$ are relevant parameters of the coefficients ($\alpha _{22}$, $\alpha _{24}$, $\alpha _{42}$, and $\alpha _{44}$) in Eq. (8). 

At each simulation time step ($t$), the solvers were initialized by setting the slack sequence to zeros. The current state vector ${\bf x}_t $ and vector  ${\bf y}_t$ for computing the optimal control sequence are described as follows:
\begin{subequations}
\begin{equation}
{\bf x}_{t}  = {\bf f}\left( {{\bf x}_{t - 1} ,{\bf u}_{t - 1}^* } \right)   + \sigma {\bf w}_t,
\end{equation}
\begin{equation}
{\bf y}_t  = {\bf Cx}_t , 
\end{equation}
\end{subequations}
where
\[
{\bf x} = \left[ {\begin{array}{*{20}c}
   \Delta   \\
   {\dot \Delta }  \\
   \theta   \\
   {\dot \theta }  \\
\end{array}} \right],\quad{\bf C} = \left[ {\begin{array}{*{20}c}
   1 & 0 & 0 & 0  \\
   0 & 1 & 0 & 0  \\
   0 & 0 & 1 & 0  \\
   0 & 0 & 0 & 1  \\
\end{array}} \right].
\]
Here ${\bf u}_{t - 1}^*  = \left[ {\delta _{t - 1}^* } \right]$ is the first element of the optimal control sequence of the previous iteration, ${\bf w}_t $ is the disturbance caused by uniformly distributed noise in the current sampling step, and $\sigma$ is the corresponding noise level. The selected initial state was 
${\bf x}_0  = \left[ {\begin{array}{*{20}c}
   2 & 0 & 0 & 0  \\
\end{array}} \right]^T$, and the parameter and constraint values listed in Tables I and II were used. Three noise levels ($\sigma $ = 0.0, 1.0, and 2.0) were used to evaluate the algorithms' robustness to disturbance.

\begin{figure*}[!t]
{\includegraphics*[width=0.33\linewidth]{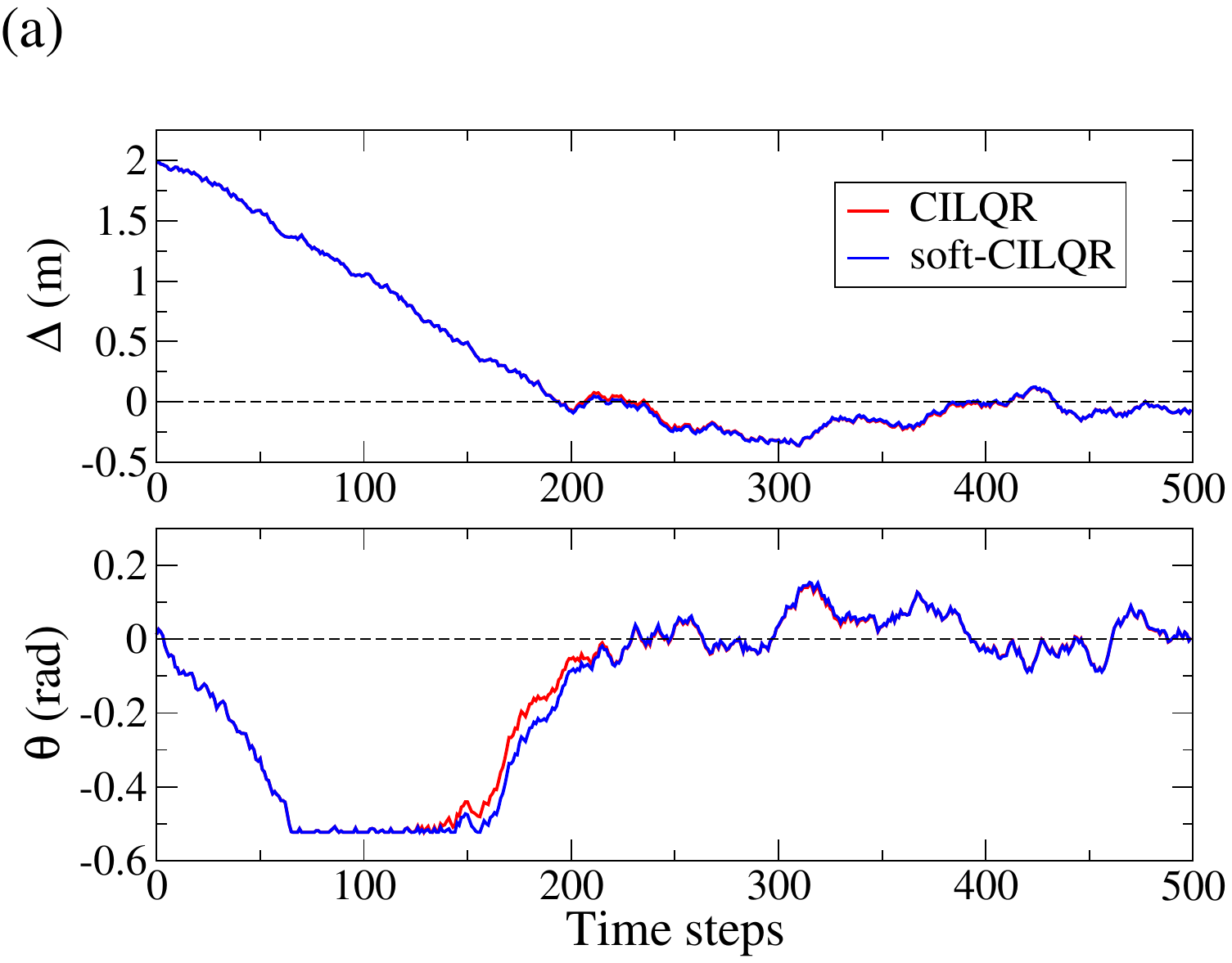}}%
{\includegraphics*[width=0.33\linewidth]{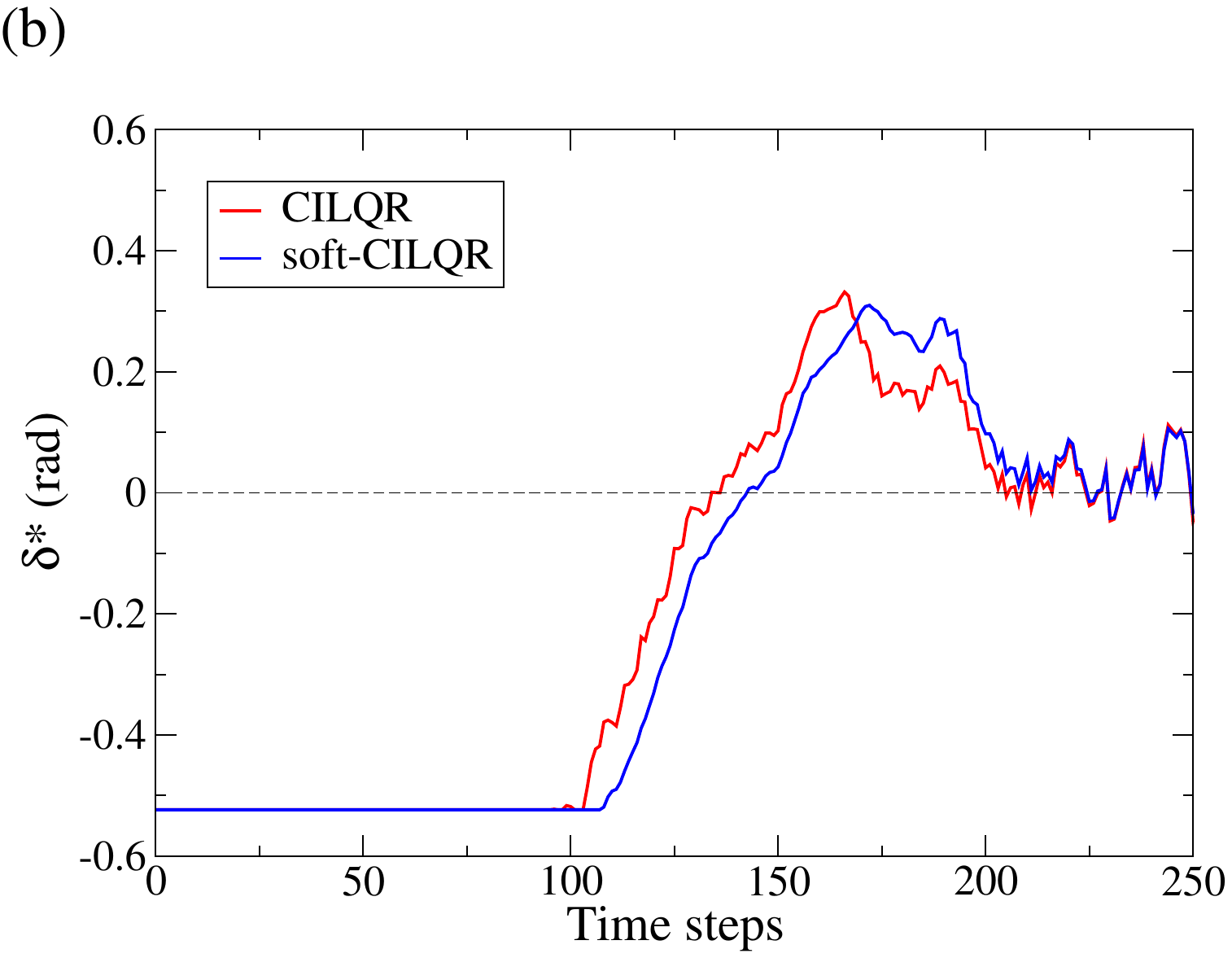}}%
{\includegraphics*[width=0.33\linewidth]{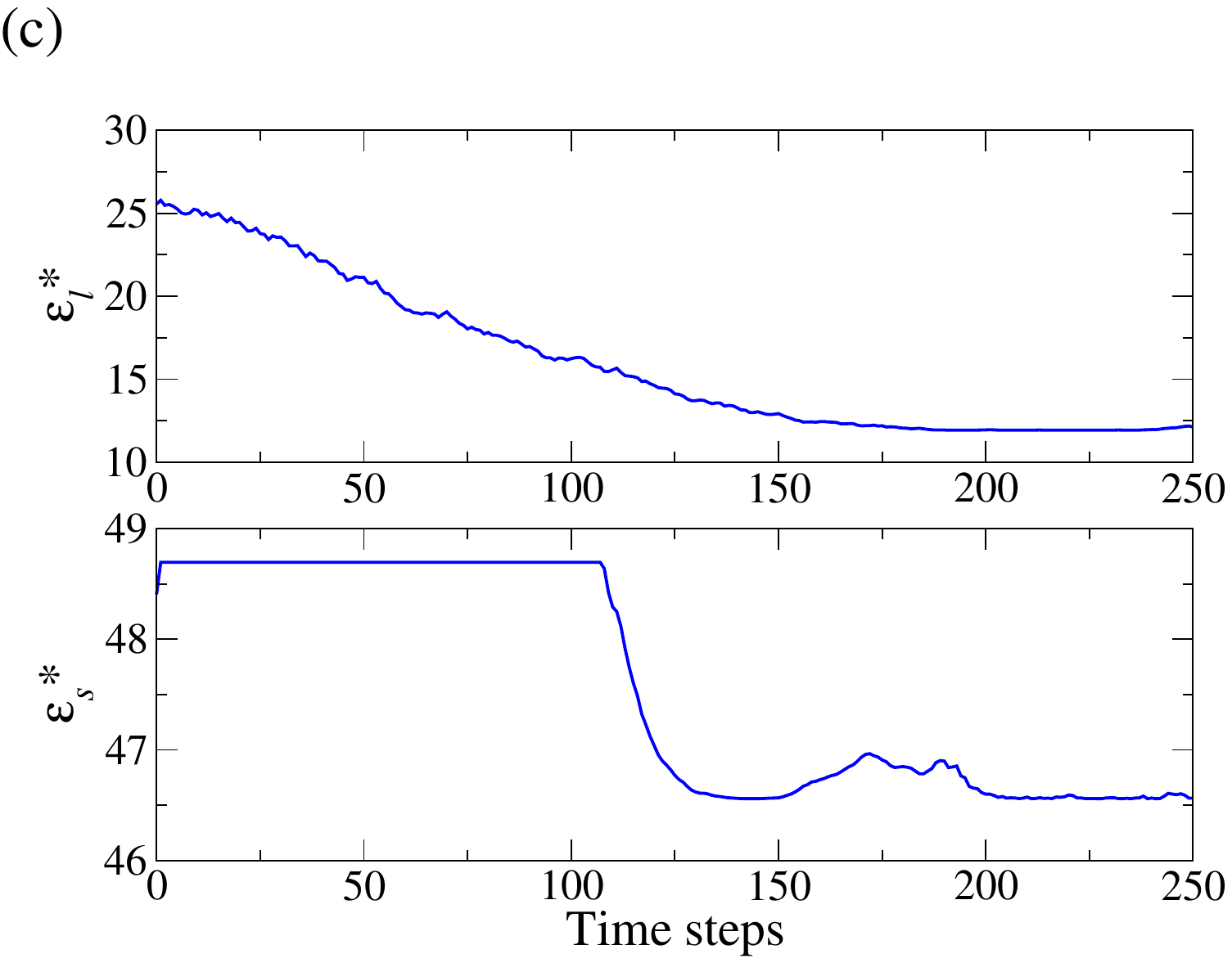}}%
\caption{Numerical simulations results obtained for the CILQR and soft-CILQR solvers when $v_x$ = 20.0 m/s, $S$ = 0.01, $N$ = 40,  $\varepsilon _{\max}$ = 49, and $\sigma $ = 2.0. (a) Trajectories of $\Delta$ and $\theta $. Trajectories of the first component of the optimal (b) steering angle  and (c) slack sequences. }
\end{figure*}

\begin{figure*}[!t]
{\includegraphics*[width=0.33\linewidth]{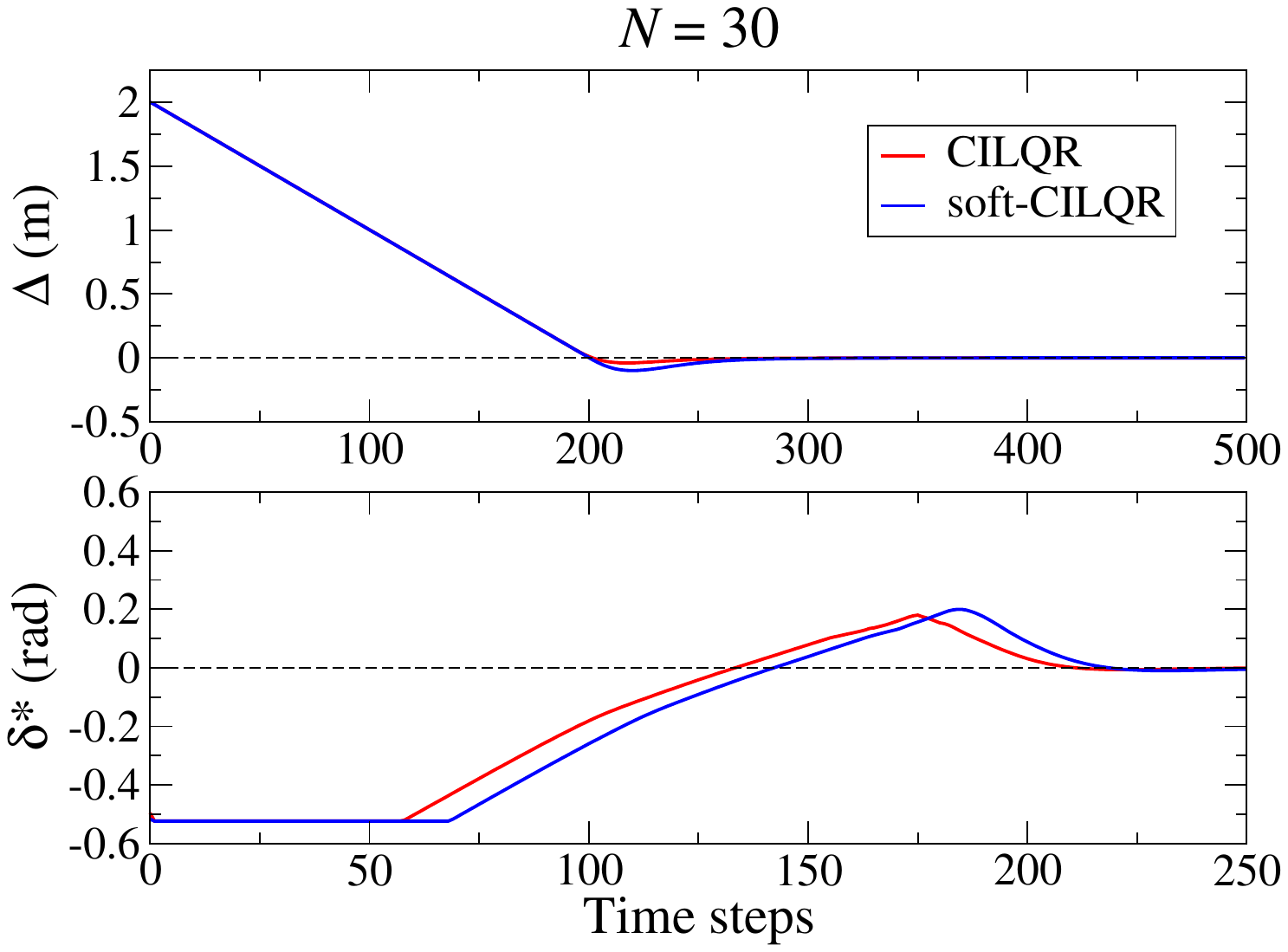}}%
{\includegraphics*[width=0.33\linewidth]{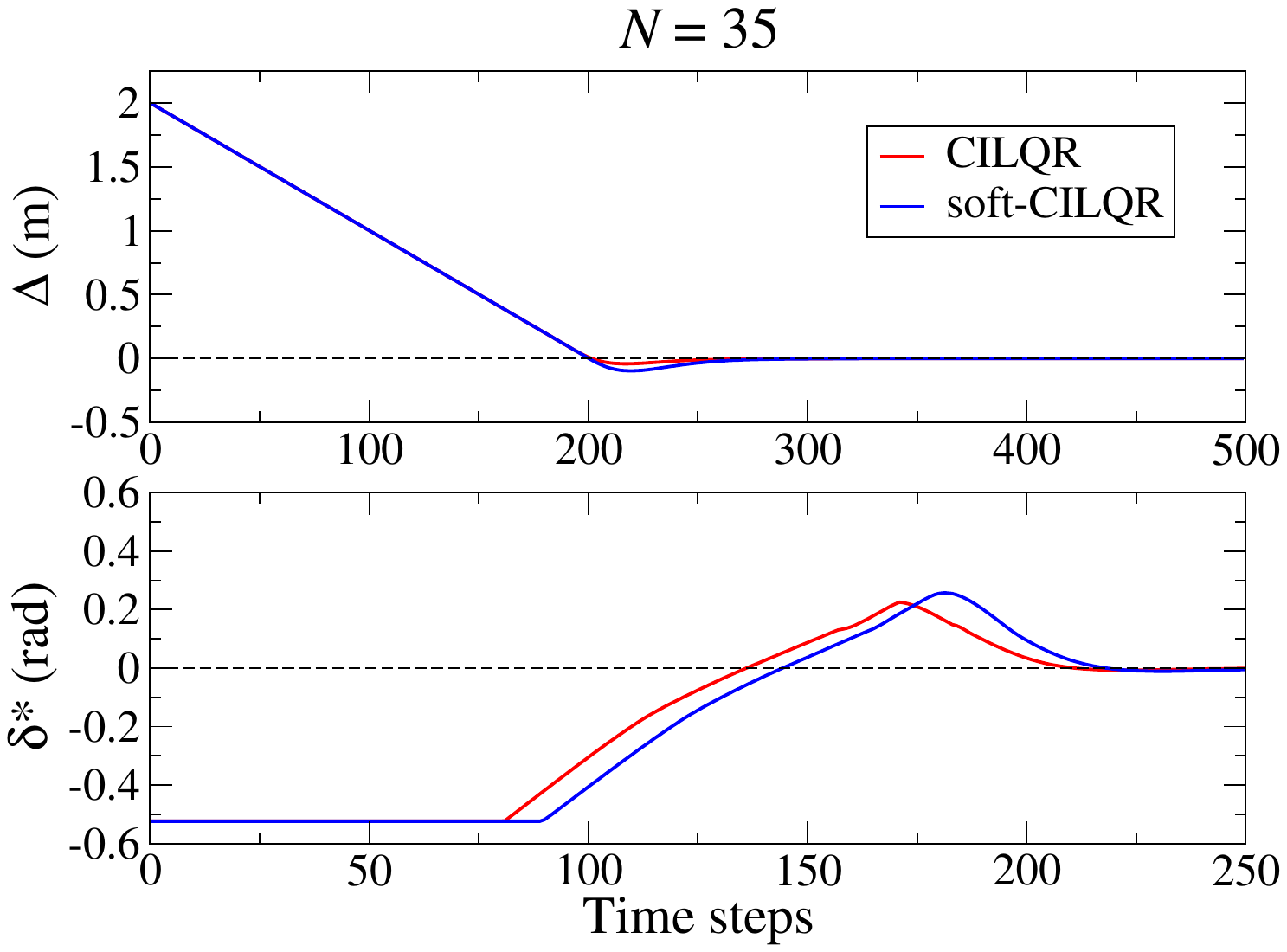}}%
{\includegraphics*[width=0.33\linewidth]{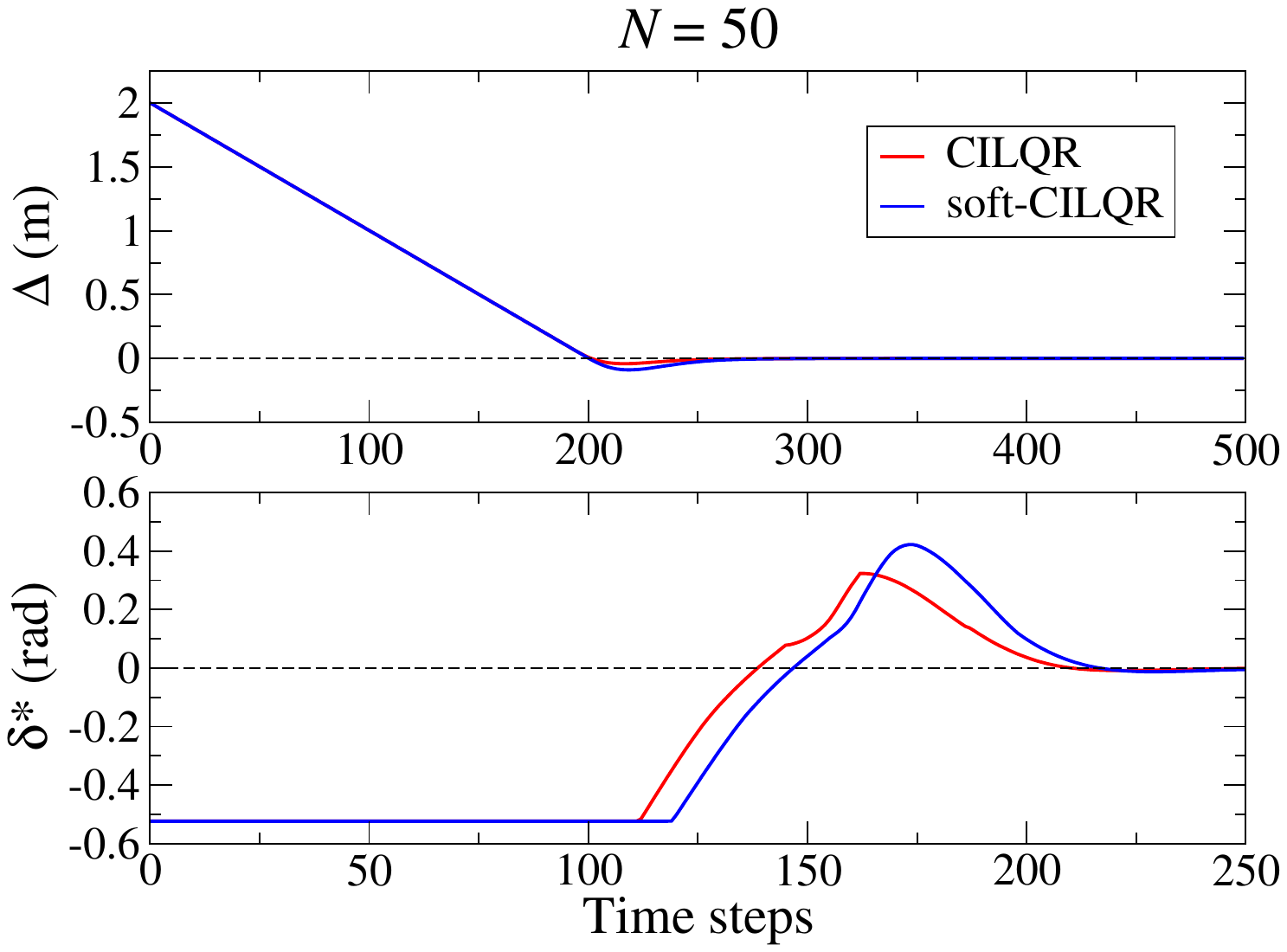}}%
\caption{Numerical simulation results obtained for the trajectories of $\Delta$ by using the soft-CILQR solver when $v_x$ = 20.0 m/s, $S$ = 0.01, $\varepsilon _{\max}$ = 49, $\sigma $ = 0.0, and $N$ = 30, 35, 50.}
\end{figure*}

\begin{figure}[!t]
{\includegraphics*[width=0.5\linewidth]{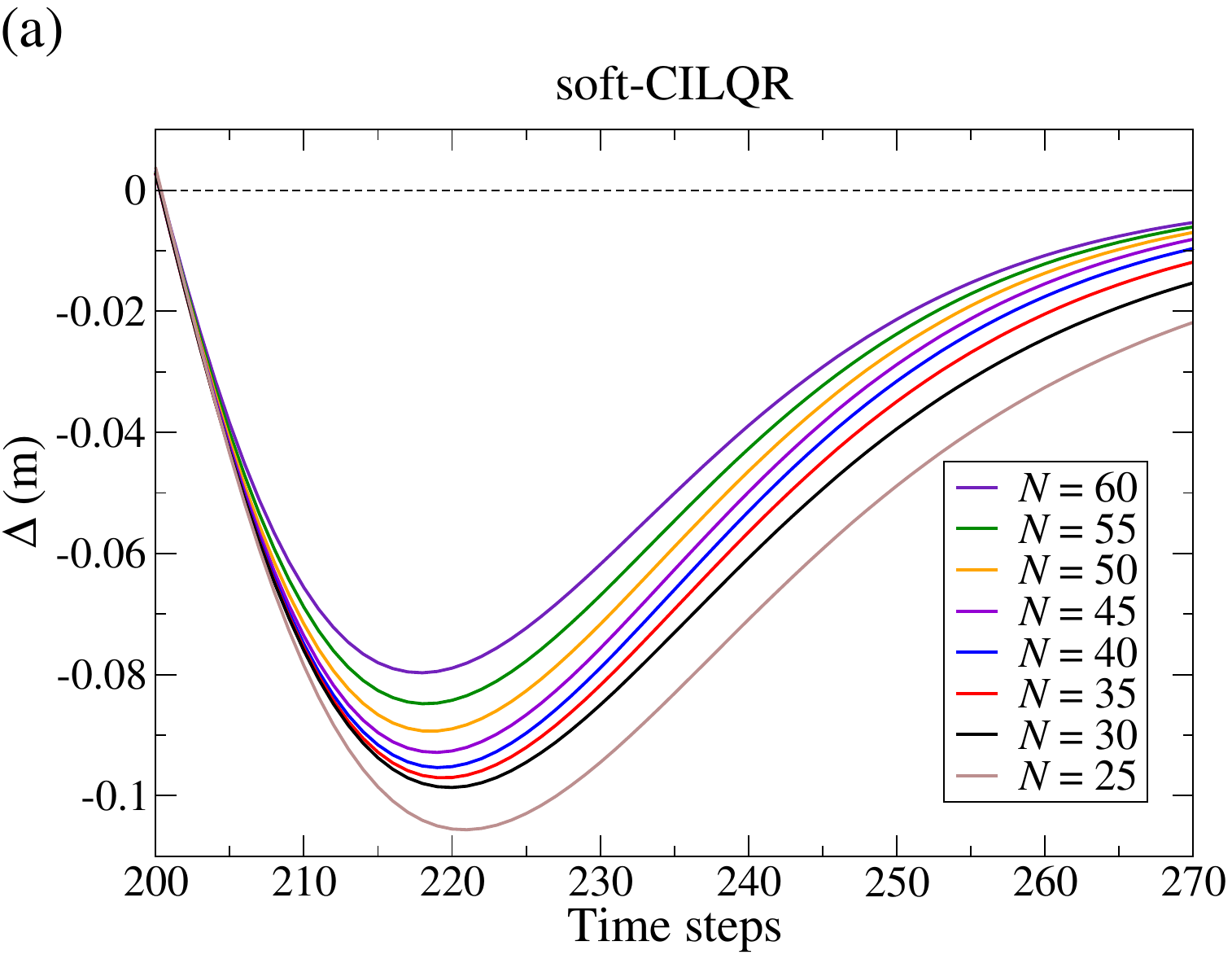}}%
{\includegraphics*[width=0.5\linewidth]{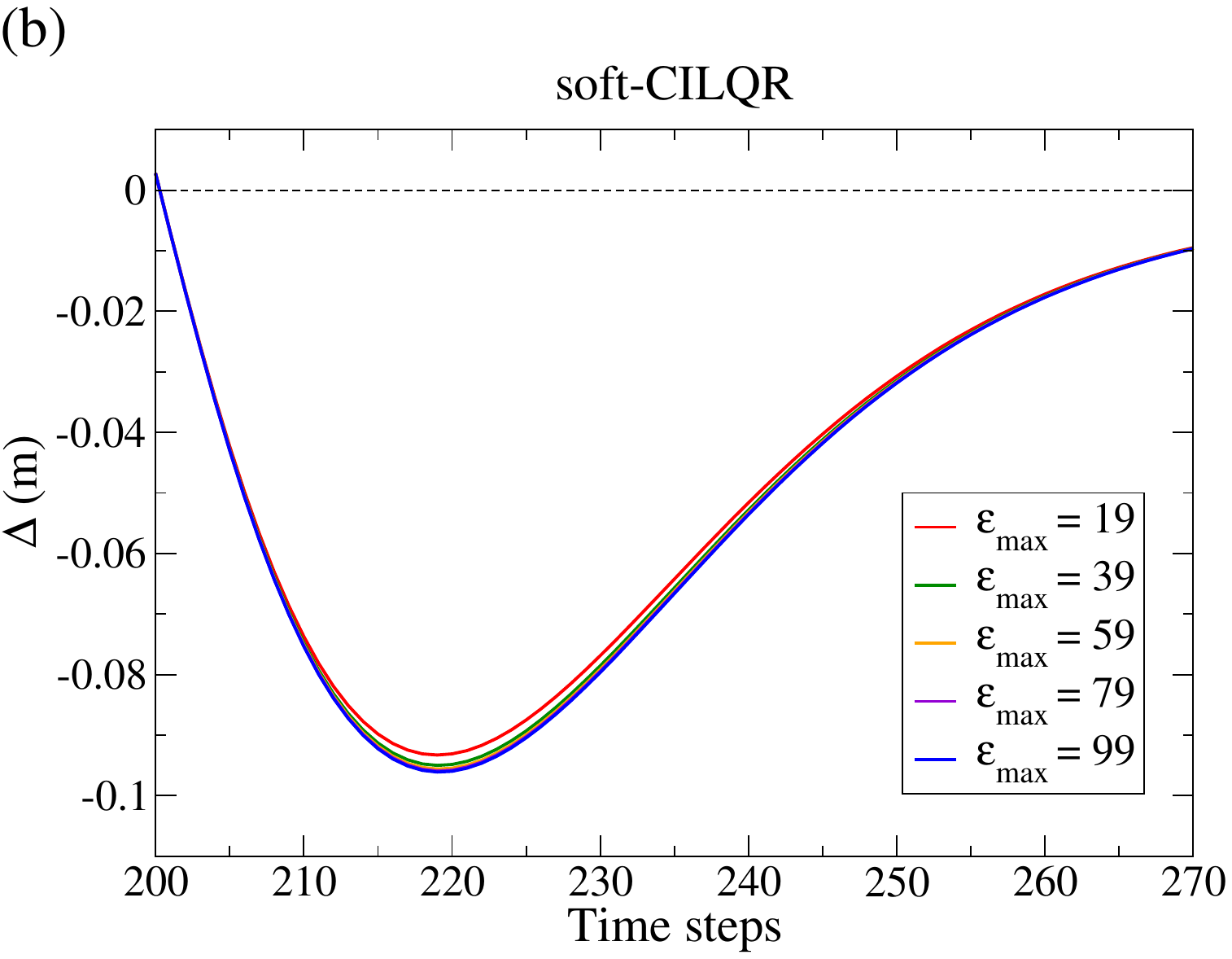}}%
\caption{Numerical simulation results obtained for the trajectories of $\Delta$ by using the soft-CILQR solver for (a) $v_x$ = 20.0 m/s, $S$ = 0.01, $\varepsilon _{\max}$ = 49, $\sigma $ = 0.0, and various values of $N$. Dashed lines represent reference data points (zeros). The minimum values of $\Delta $ were  $-$0.0797,  $-$0.0953, and $-$0.1056  m for $N$ = 60, 40, and 25, respectively. Trajectories for (b) $\sigma $ = 0.0, $N$ = 40, and various values of $\varepsilon _{\max}$. The minimum values of $\Delta $ were $-$0.0932, $-$0.0950, $-$0.0956, $-$0.0.0959, and $-$0.0961 m when $\varepsilon _{\max}$ = 19, 39, 59, 79, and 99, respectively.}
\end{figure}

Fig. 3(a) presents the $\Delta $ and $\theta $ trajectories generated by the soft-CILQR and CILQR solvers when $\sigma $ = 0.0. Both algorithms could control the vehicle system asymptotically to enable it to reach the reference point (${\bf 0}$). Moreover, the soft-CILQR  controller with relaxed constraints was less conservative than the CILQR controller; the trajectory obtained for $\Delta $ reached the target point earlier with the soft-CILQR solver than with the CILQR solver. When the noise level  increased to $\sigma $ = 1.0 and 2.0 [Fig. 4(a) and Fig. 5(a), respectively], the  control performance of both algorithms was poor. The system exhibited obvious instability when the state trajectory approached  points near $t = 200$ at which the magnitude of the state vector ${\bf x}$ was close to that of the disturbance. The trajectories obtained using the CILQR and soft-CILQR algorithms exhibited similar behaviors at long times, and the maximum $\Delta $ deviations between the two methods were approximately 0.18 and 0.36 m when $\sigma $ = 1.0 and 2.0, respectively.

Fig. 3(b), Fig. 4(b), and Fig. 5(b) present the trajectories obtained for the optimal steering angle ($\delta ^* $) when $\sigma$ = 0.0, 1.0, and 2.0, respectively. The $\delta ^* $ values were  clipped for short times ($ t <$ 100) to prevent the steering signal from exceeding the relevant physical limitation ($-\pi/6$). Compared with the solution curves obtained using the CILQR algorithm, $\delta ^* $ trajectories obtained using the soft-CILQR algorithm exhibited greater overshoot around $t$ = 170 when $\sigma$ = 0.0. Similar behavior was also observed in \cite{ Fel17a}, in which the overshoot of the control input curve increased with the extent of constraint relaxation. Fig. 3(c), Fig. 4(c), and Fig. 5(c) display the trajectories of  the first element of the optimal slack  sequences ${\bf e}^*  = \left[ {\begin{array}{*{20}c}
   {\varepsilon _l^* } & {\varepsilon _s^* }  \\
\end{array}} \right]^T $. Overall, the  ${\varepsilon _l^* }$ and ${\varepsilon _s^* }$ trajectories did not exceed their corresponding upper limits ($\varepsilon _{\max}$ = 49). The ${\varepsilon _s^* }$ trajectories exhibited overshoot at around $t$ = 170. This overshoot corresponded to the overshoots of the $\delta ^*$ curves displayed in Fig. 3(b), Fig. 4(b), and Fig. 5(b) at the same time step. In particular, the soft-CILQR solver yielded  smoother $\delta ^*$ curves than did the CILQR solver when 100 $< t <$ 200 [Fig. 3(b), Fig. 4(b), and  Fig. 5(b)].

\begin{figure*}[!t]
{\includegraphics*[width=0.33\linewidth]{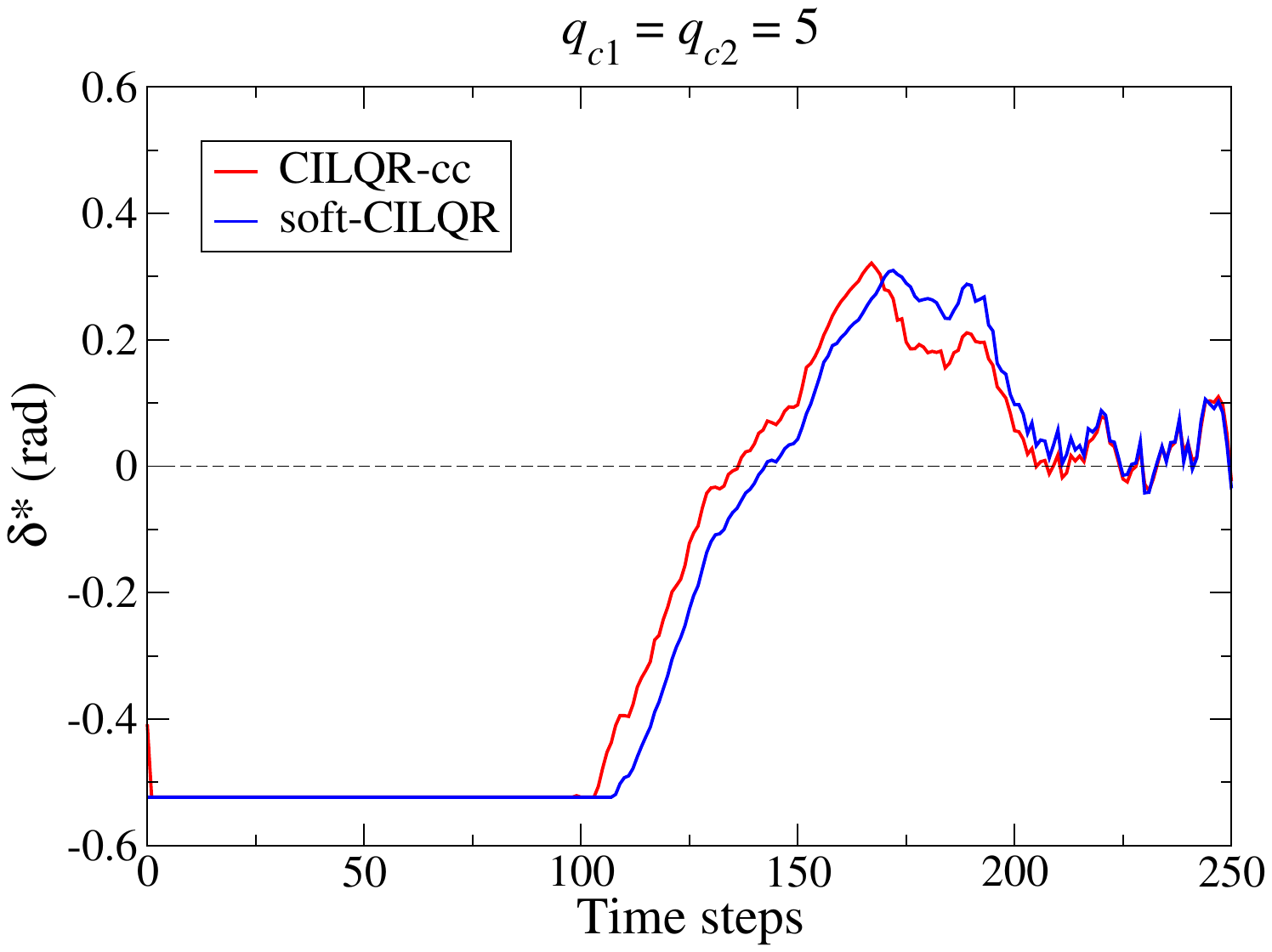}}%
{\includegraphics*[width=0.33\linewidth]{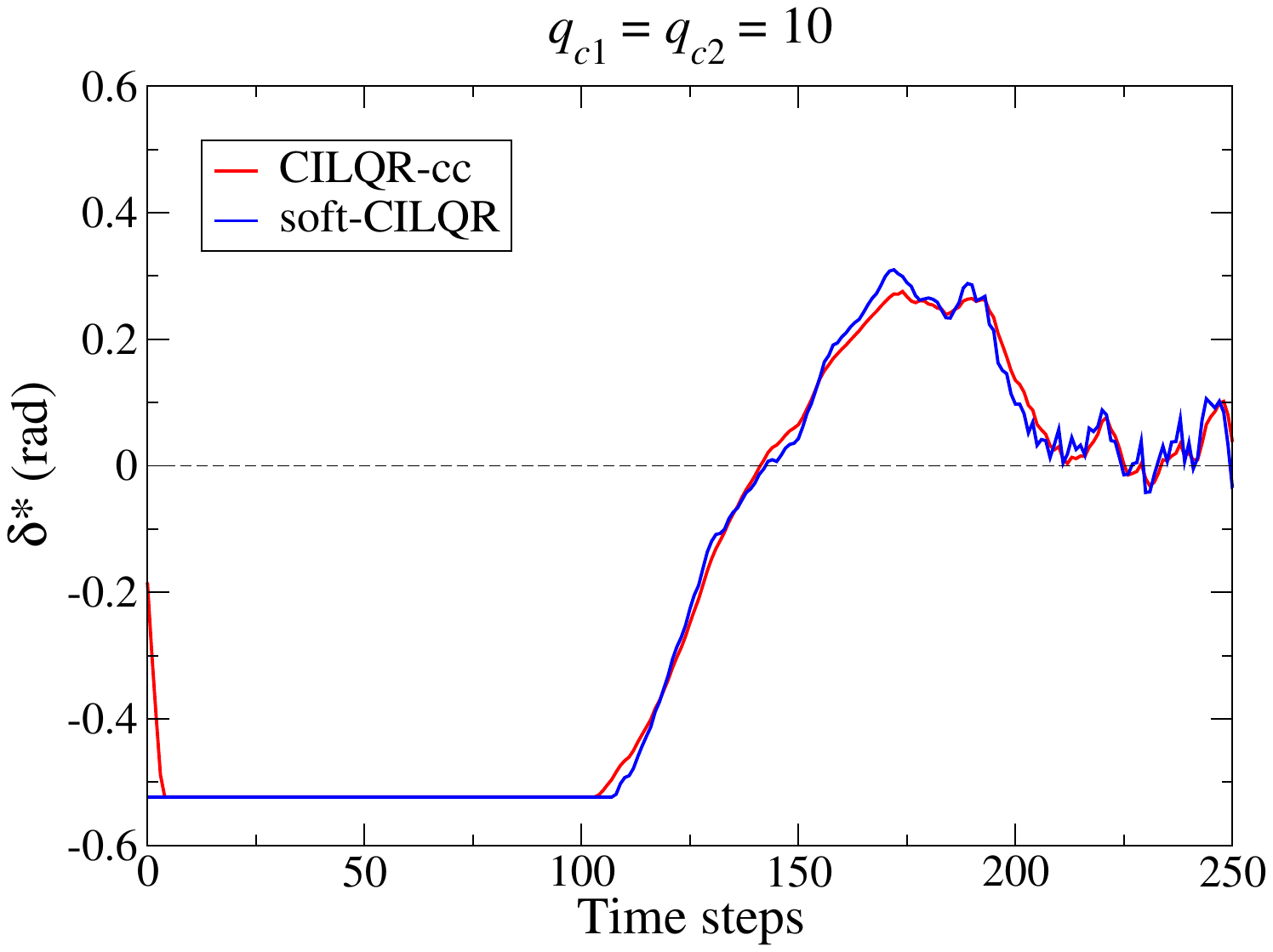}}%
{\includegraphics*[width=0.33\linewidth]{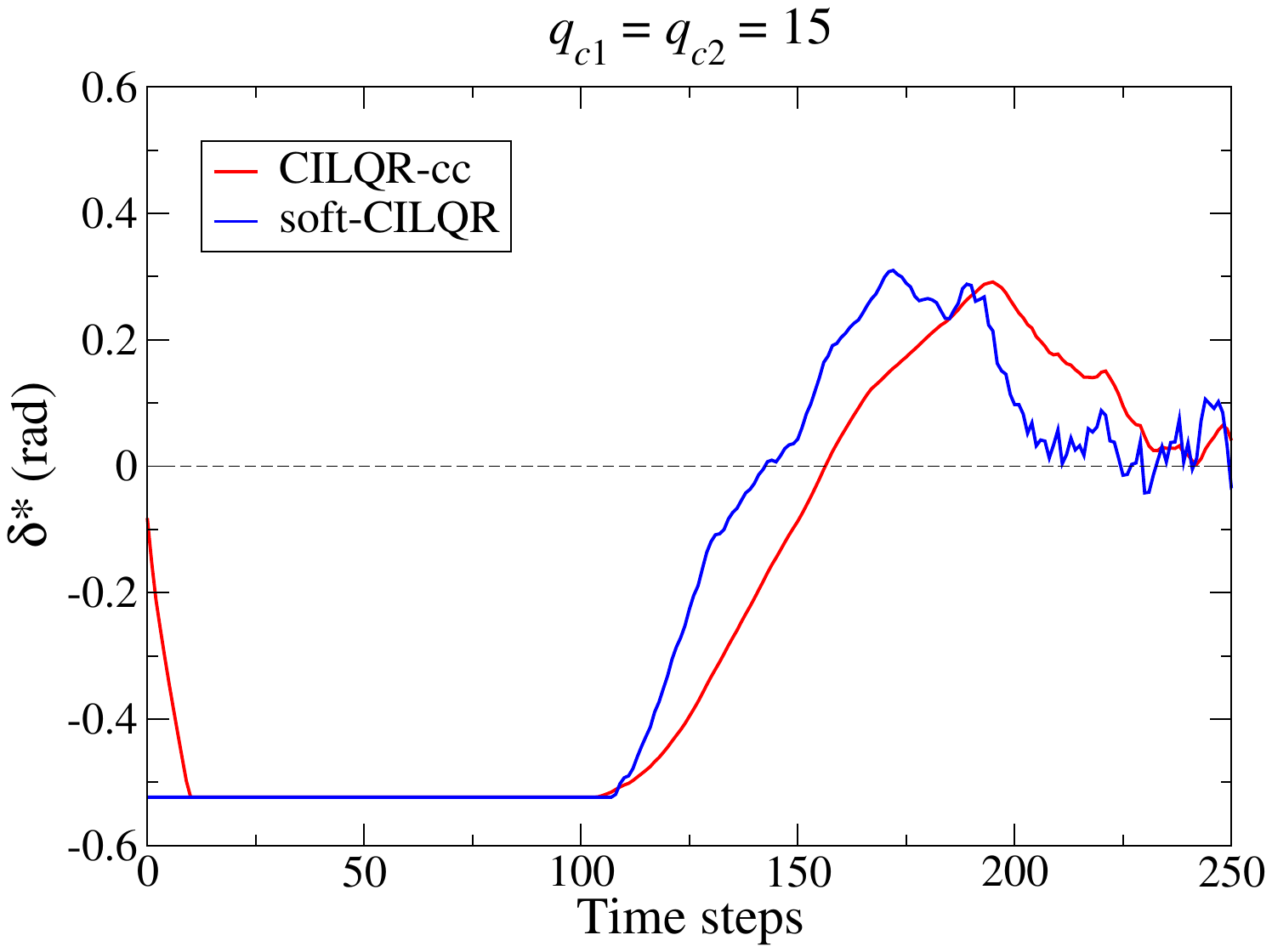}}%
\caption{Numerical simulation results obtained for the trajectories of the first component of the optimal steering angle by using the CILQR-cc and soft-CILQR solvers when $v_x$ = 20.0 m/s, $S$ = 0.01, $\varepsilon _{\max}$ = 49, $N$ = 40, and $\sigma $ = 2.0. For CILQR-cc, $q_{c1,2}$ = 5, 10, 15 were used.}
\end{figure*}

\begin{figure*}[!t]
{\includegraphics*[width=0.33\linewidth]{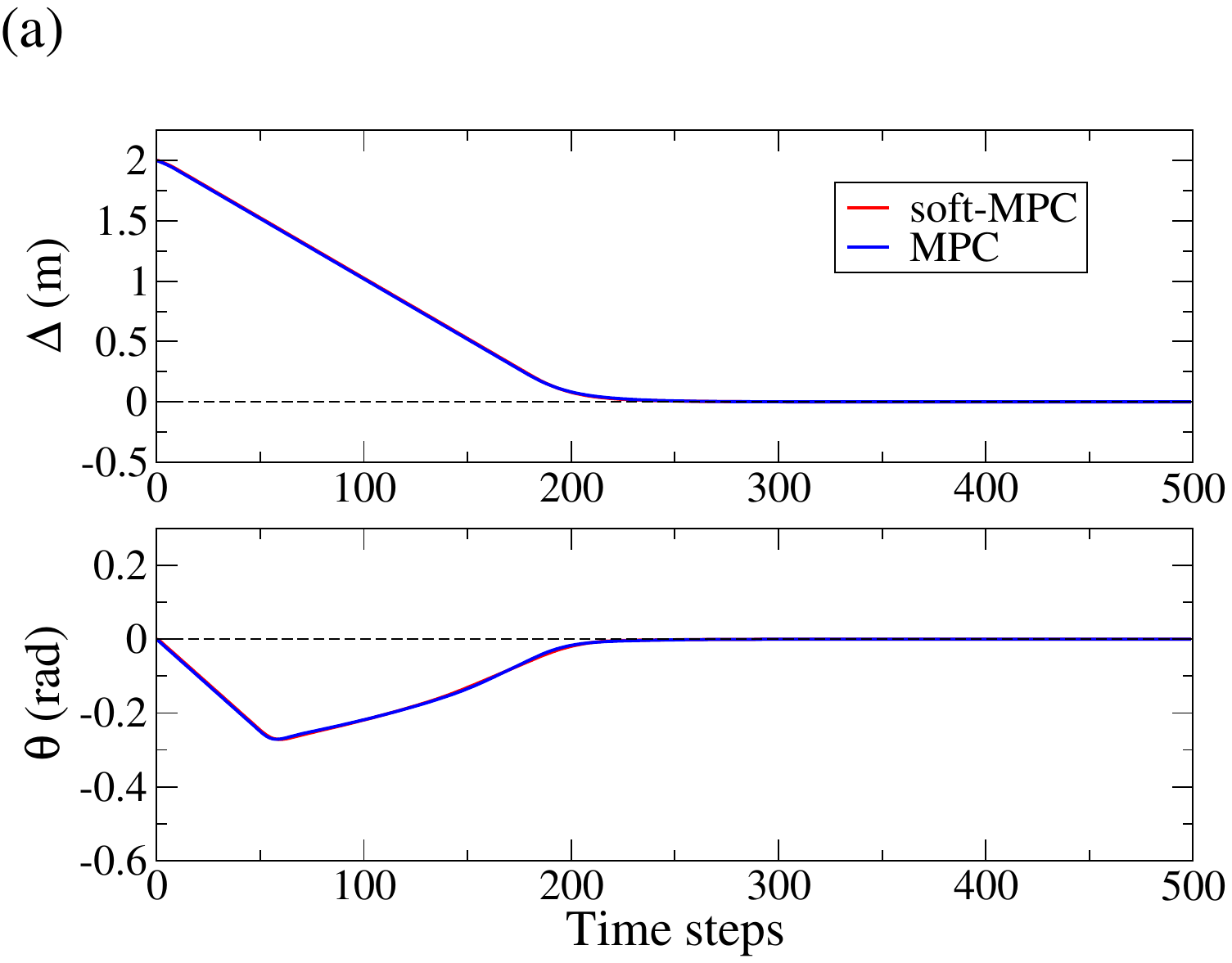}}%
{\includegraphics*[width=0.33\linewidth]{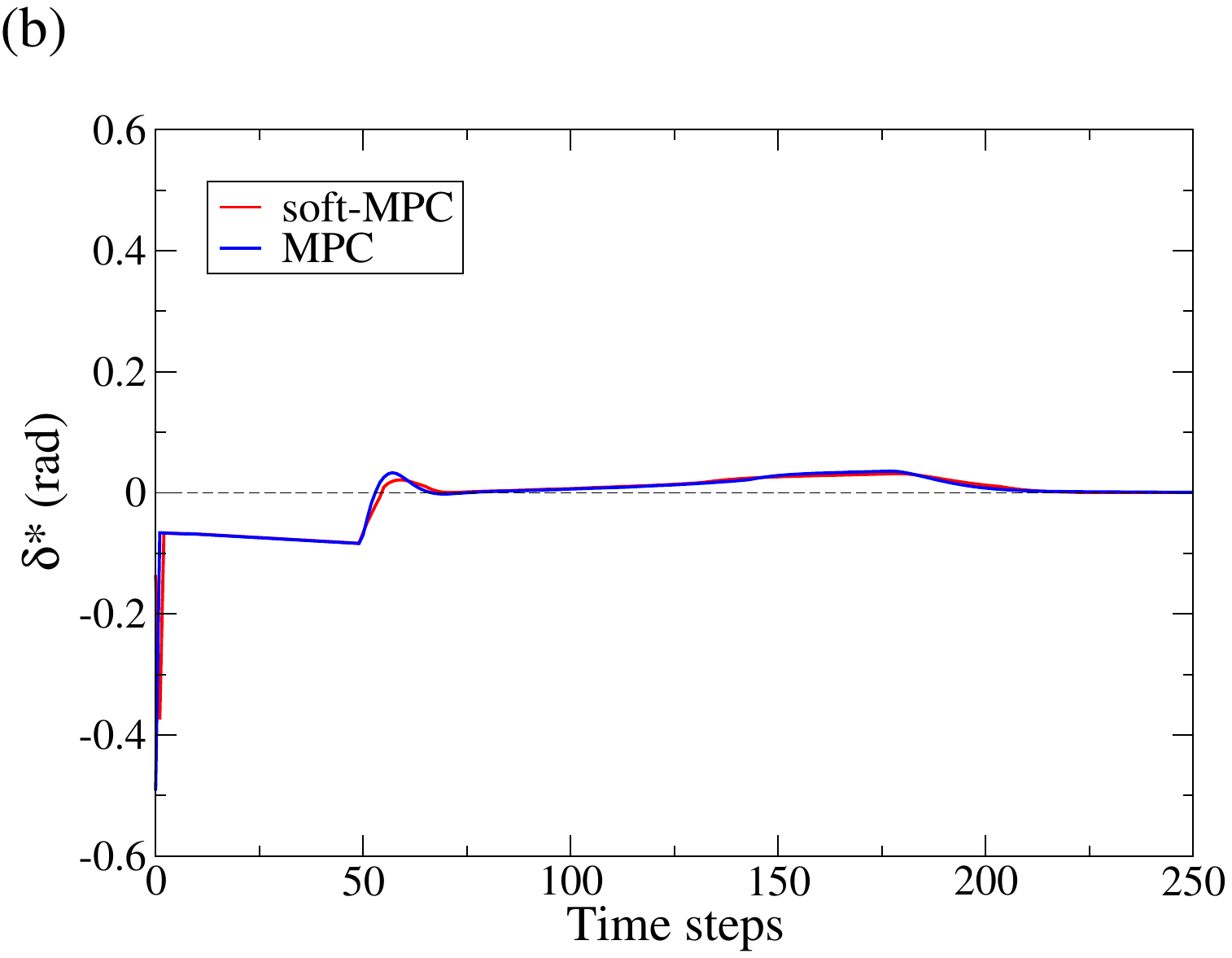}}%
{\includegraphics*[width=0.33\linewidth]{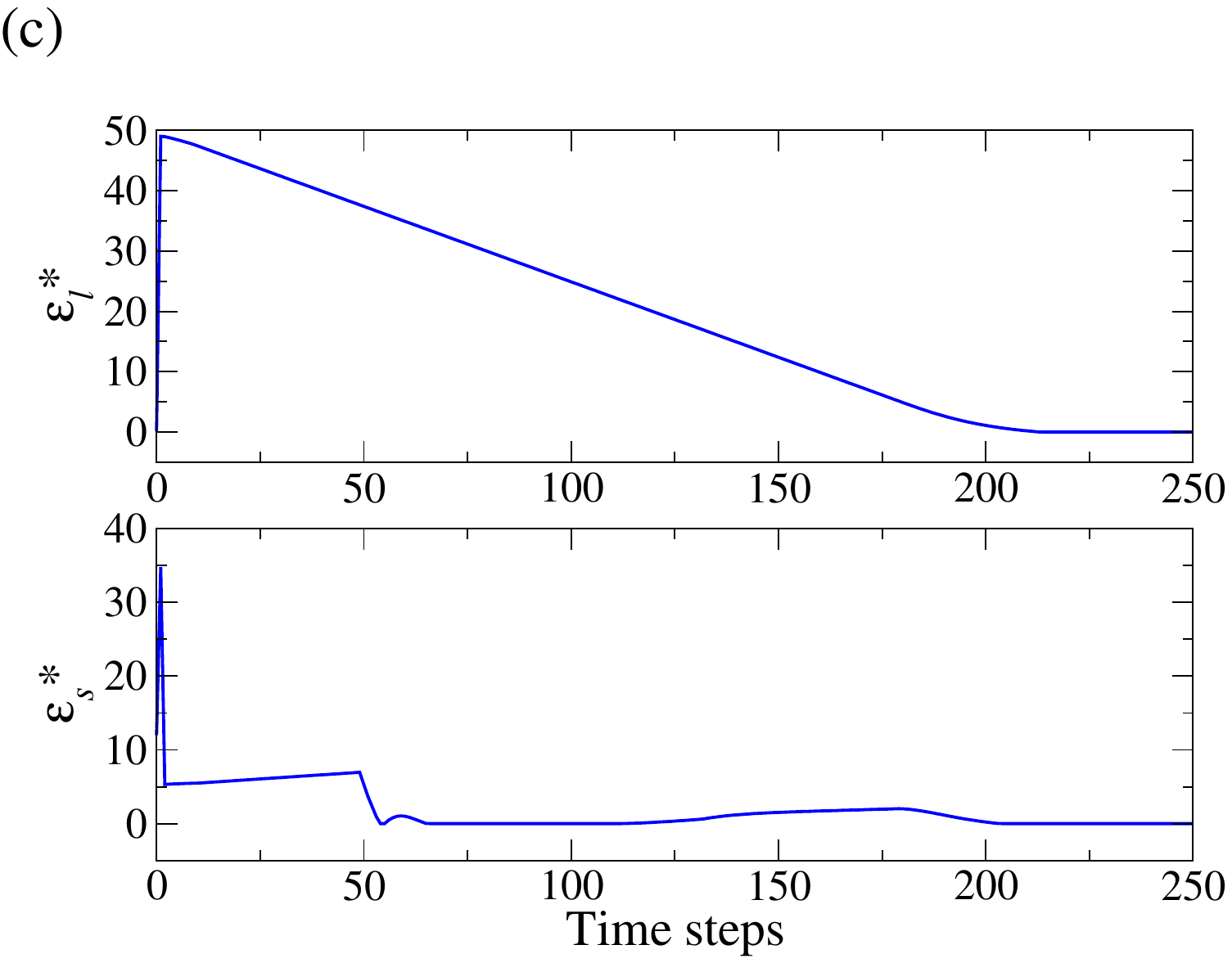}}%
\caption{Numerical simulation results obtained for the soft-MPC and MPC solvers when $v_x$ = 20.0 m/s, $N$ = 40, and $\sigma $ = 0.0. For soft-MPC, $\varepsilon _{\max}$ = 49 and $S$ = 0.01 were used. (a) Trajectories of $\Delta$ and $\theta $. Trajectories of the first component of the optimal (b) steering angle  and (c) slack sequences.}
\end{figure*}

\begin{figure*}[!t]
{\includegraphics*[width=0.33\linewidth]{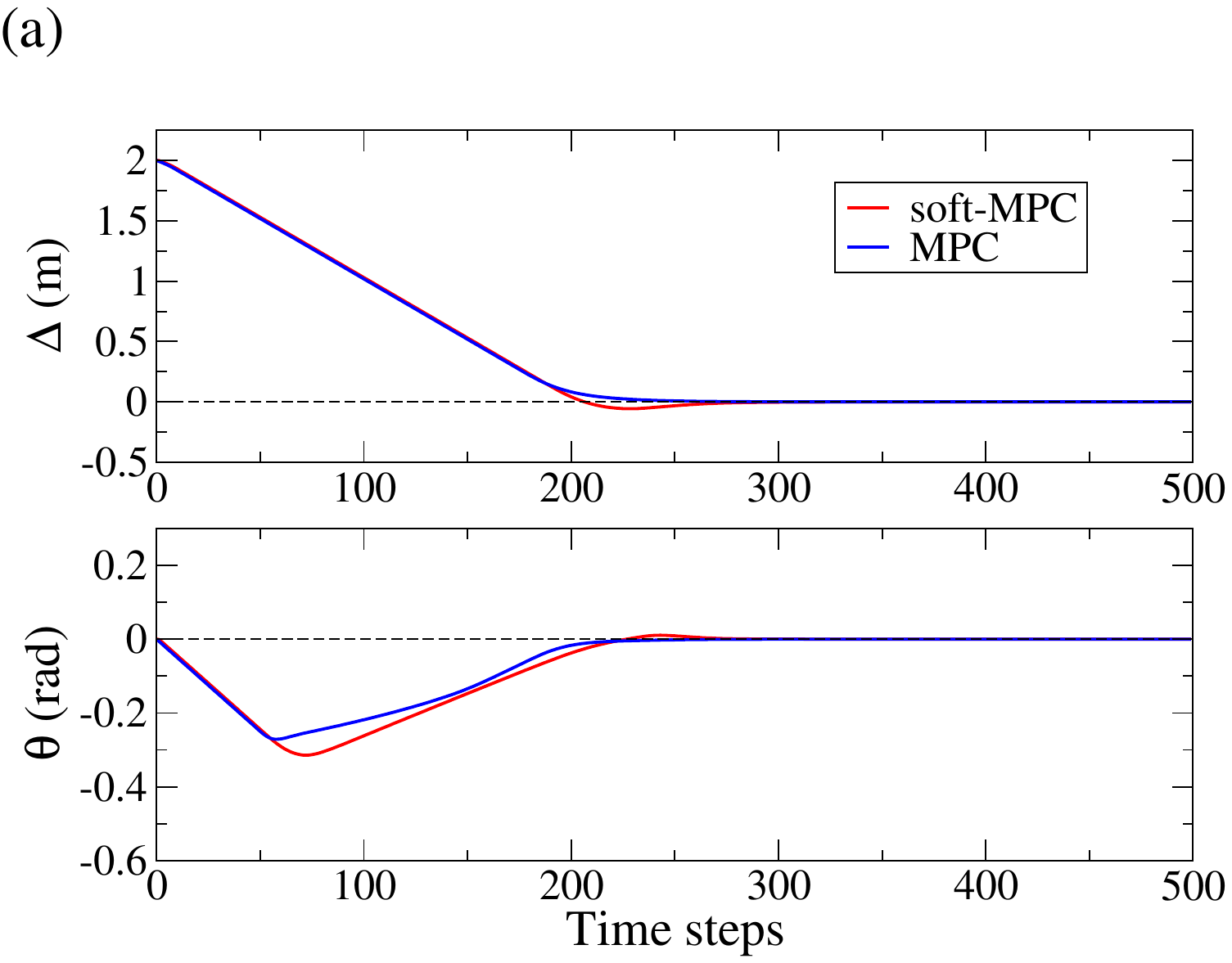}}%
{\includegraphics*[width=0.33\linewidth]{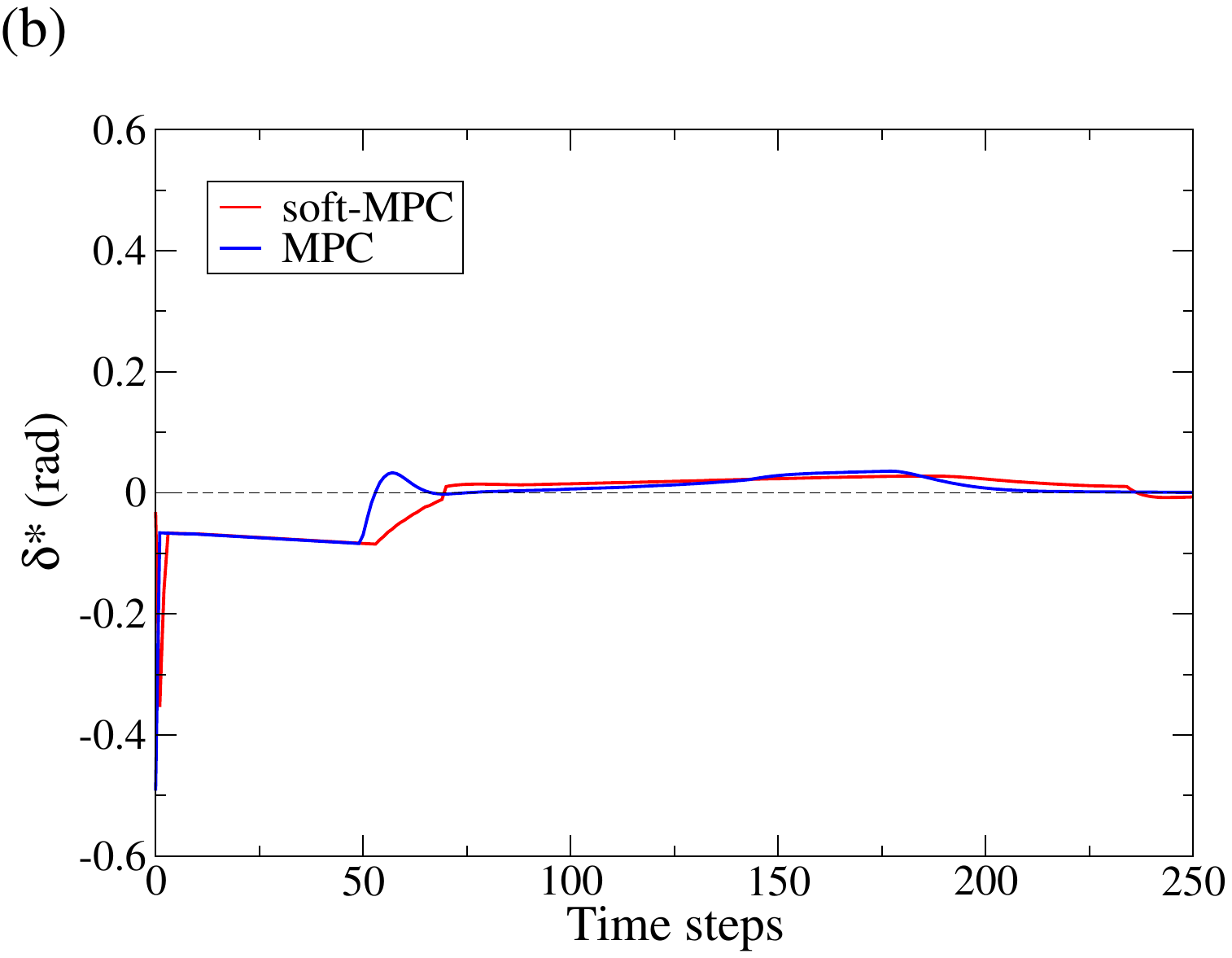}}%
{\includegraphics*[width=0.33\linewidth]{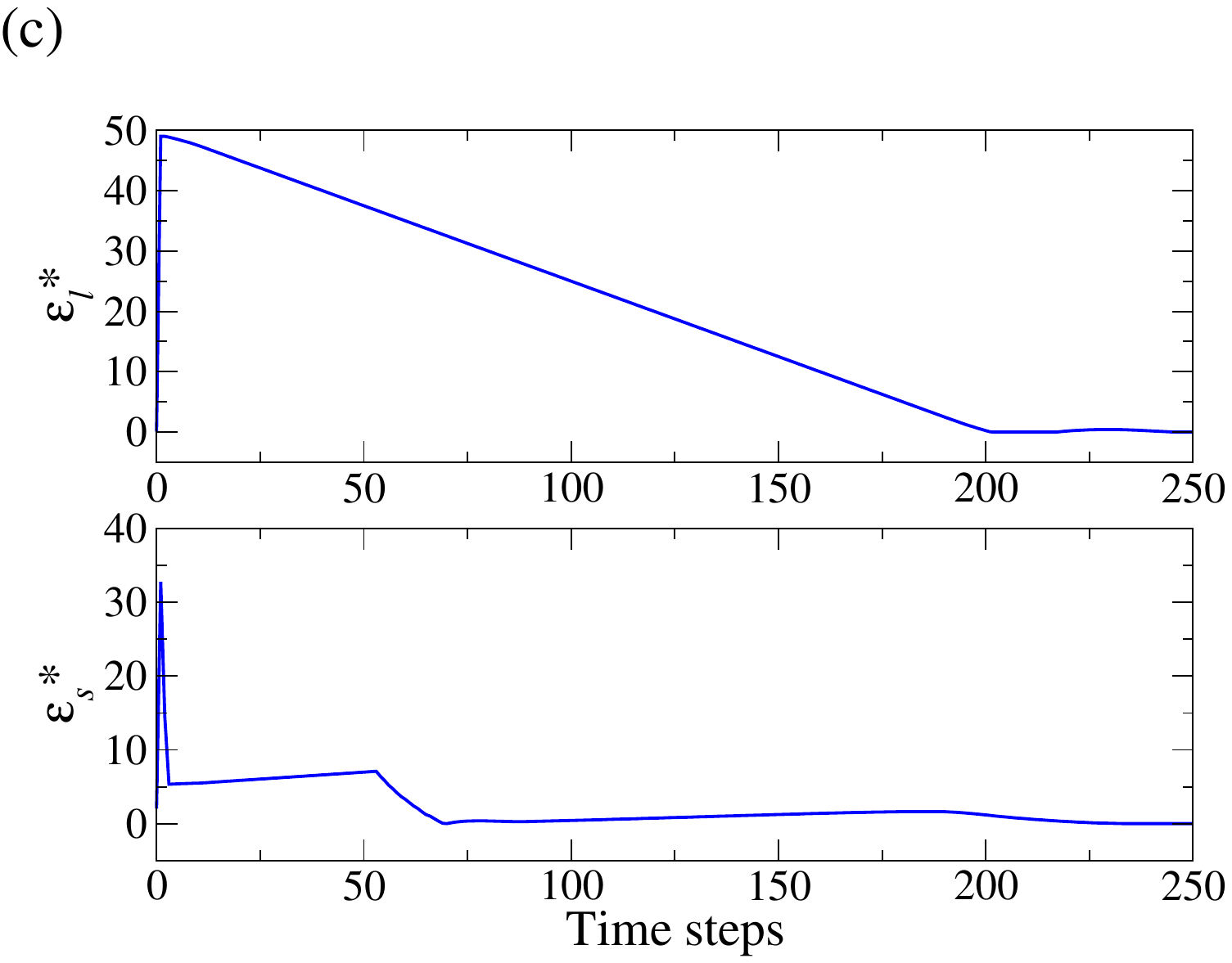}}%
\caption{Numerical simulation results obtained for the soft-MPC and MPC solvers when $v_x$ = 20.0 m/s, $N$ = 40, $\sigma $ = 0.0. For soft-MPC, $\varepsilon _{\max}$ = 49 and $S$ = 0.5 were used. (a) Trajectories of $\Delta$ and $\theta $. Trajectories of the first component of the optimal (b) steering angle and (c) slack sequences.}
\end{figure*}

\begin{figure*}[!t]
{\includegraphics*[width=0.33\linewidth]{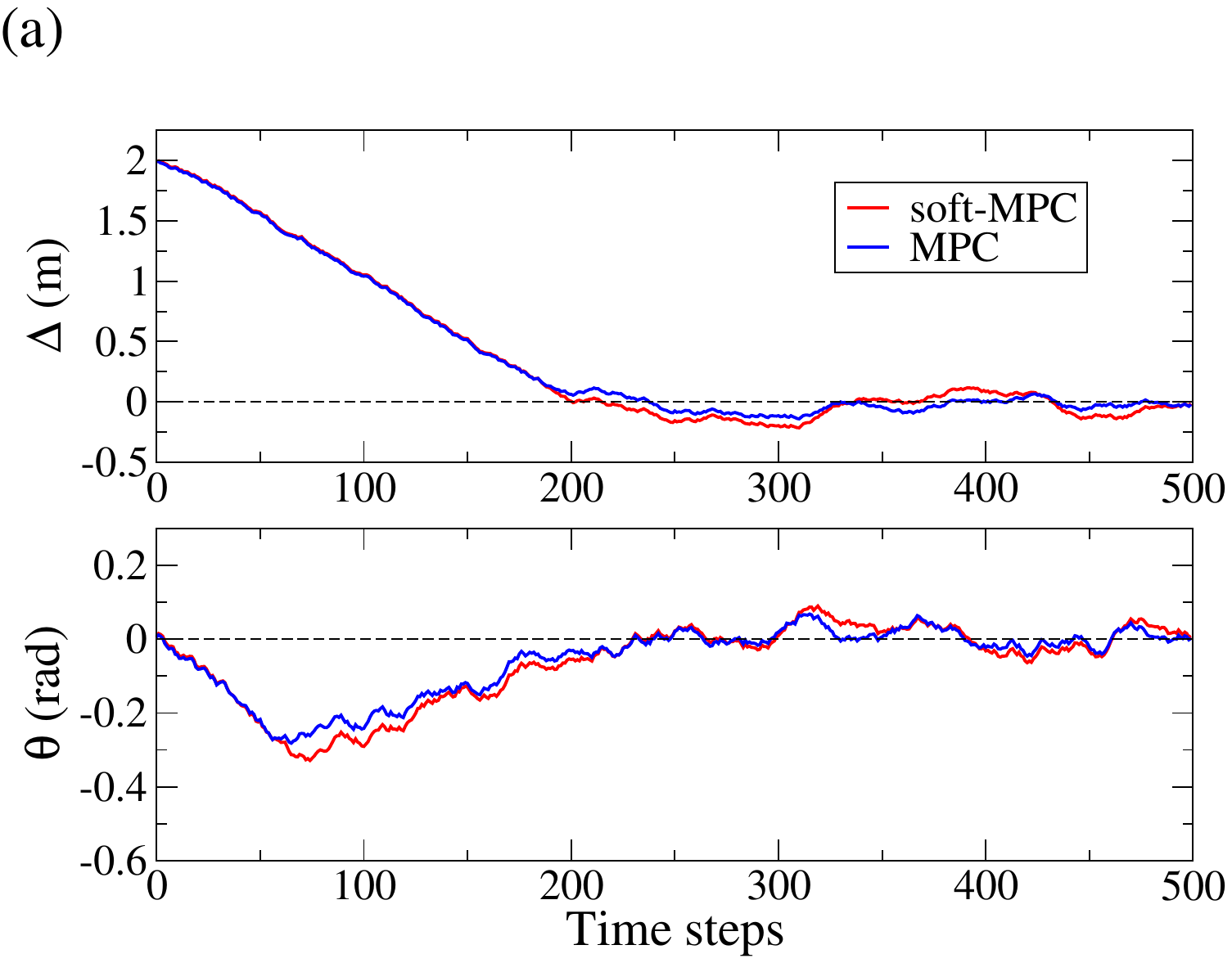}}%
{\includegraphics*[width=0.33\linewidth]{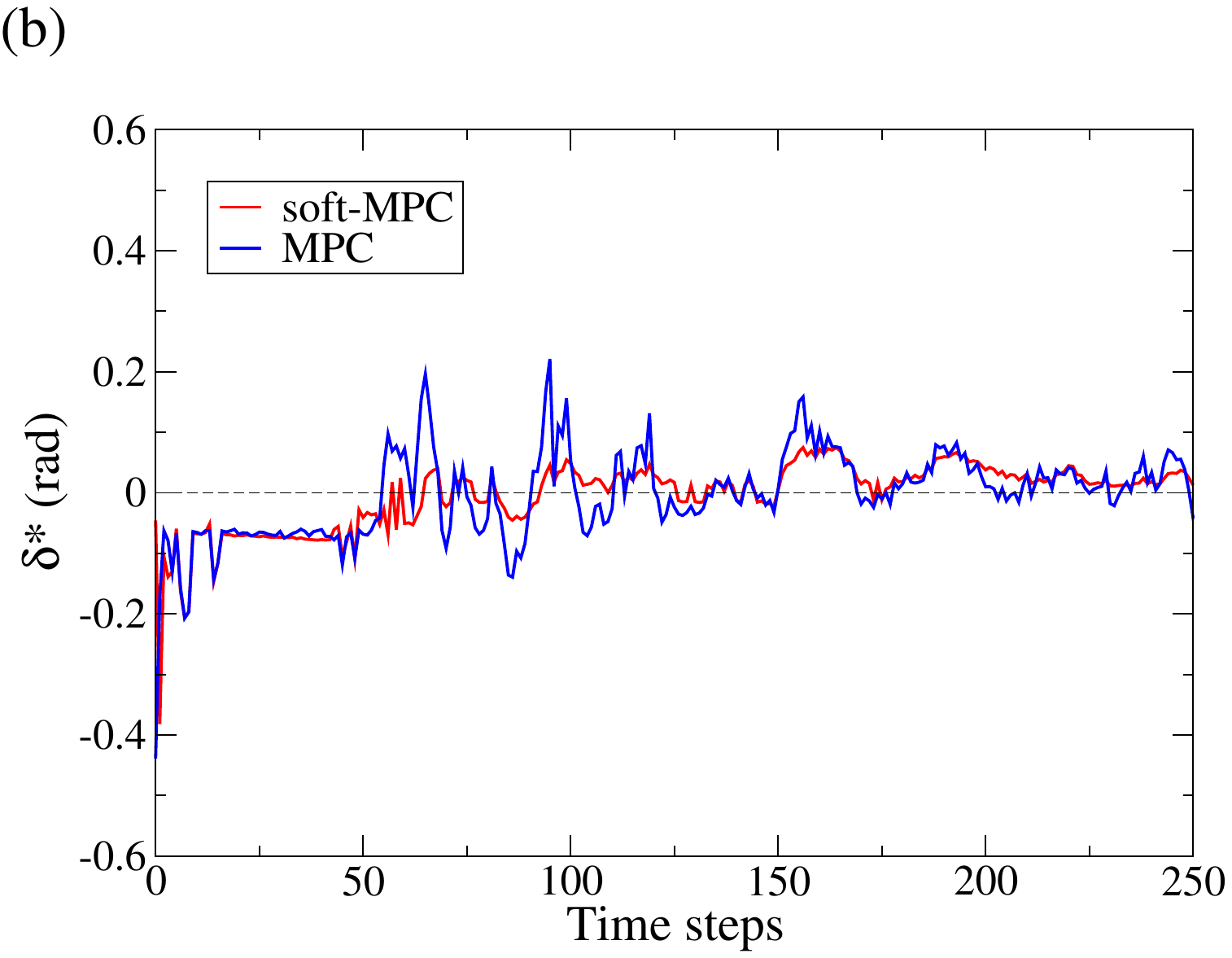}}%
{\includegraphics*[width=0.33\linewidth]{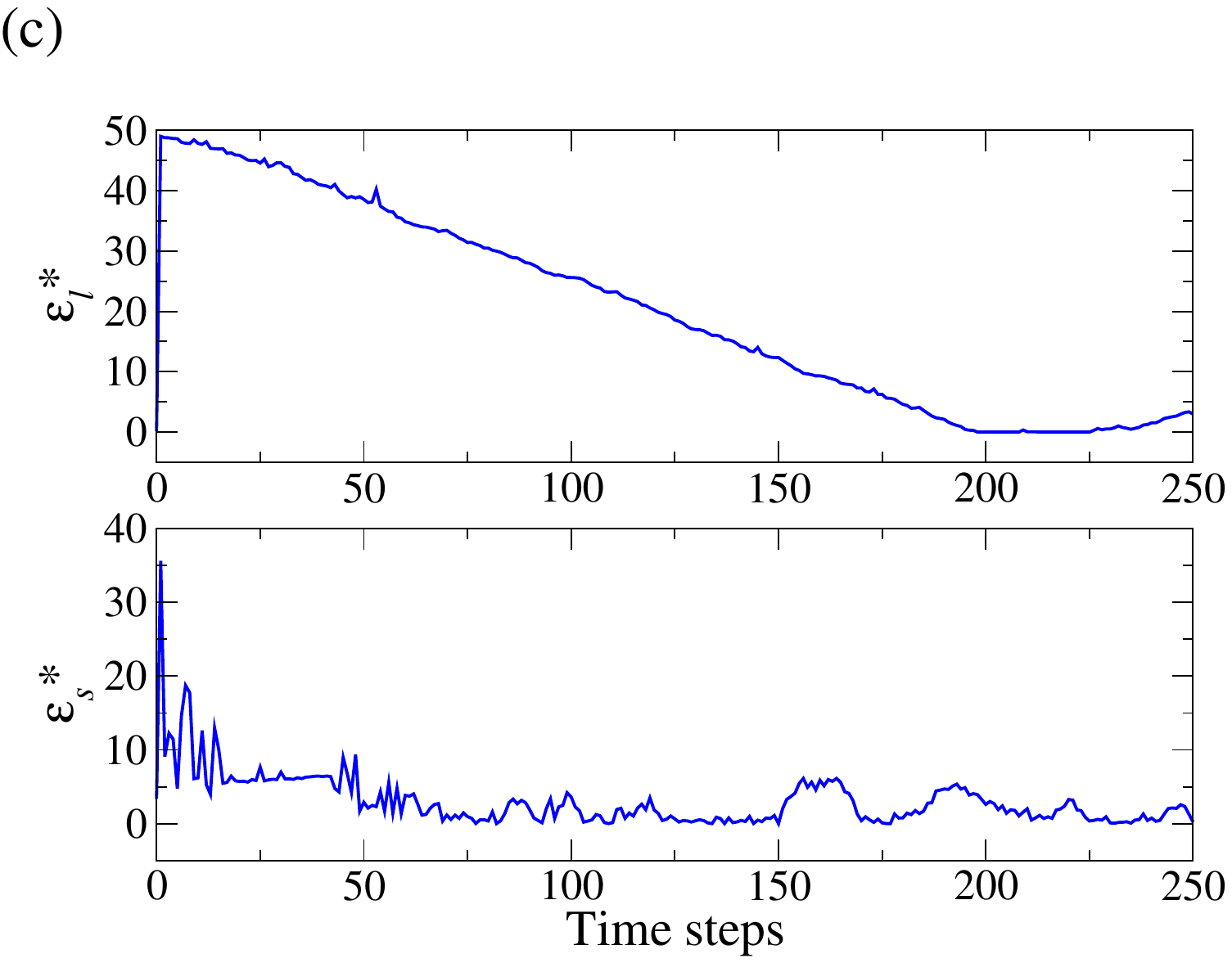}}%
\caption{Numerical simulation results obtained for the soft-MPC and MPC solvers when $v_x$ = 20.0 m/s, $N$ = 40, $\sigma $ = 1.0. For soft-MPC, $\varepsilon _{\max}$ = 49 and $S$ = 0.5 were used. (a) Trajectories of $\Delta$ and $\theta $. Trajectories of the first component of the optimal (b) steering angle and (c) slack sequences.}
\end{figure*}

The $\Delta$ and $\delta ^* $ trajectories obtained when $\sigma $ = 0 at different values of $N$ (30, 35, and 50) were compared (Fig. 6). The results indicated that the maximum $\delta ^* $ value increased with $N$. The maximum $\delta ^* $ value produced by the soft-CILQR solver at an $N$ value of 50 did not exceed the upper bound of the steering angle ($\pi /6$). Fig. 7(a) presents the evolution of the solution trajectories obtained for $\Delta $ by using the soft-CILQR solver when 200 $< t <$ 270 at $N$ values ranging from 25 to 60. An increase in $N$ from 25 to 60 resulted in the generation of conservative trajectories. Consequently, the maximum deviation  in the $\Delta$ values decreased as $N$ increased. Fig. 7(b) presents the evolution of the $\Delta$ trajectories obtained at $\varepsilon _{\max } $ values ranging from 19 to 99. The maximum deviation of the $\Delta$ values increased with $\varepsilon _{\max } $. Because $\varepsilon _{\max } $ and $N_\nu $ had an approximately linear relationship (Fig. 1), the soft-CILQR solver provided less conservative trajectories as $N_\nu $ increased.

Overall, the overshoot of $\Delta$ trajectories highlighted in Fig. 7 stemmed from the less conservative behavior of the soft-CILQR algorithm compared with the CILQR algorithm without soft constraints. This behavior also occurs for the state trajectories of a classical double integrator system when  the same constraint-softening technique is applied (Fig. 1 in \cite{Svr23}). However, this overshoot leads to inherently larger lateral derivations (with respect to zeros) of the $\Delta$ trajectories obtained by the soft-CILQR algorithm relative to those of the CILQR algorithm [Fig. 3(a)]. We noticed that the $\Delta$ solution trajectory of the soft-CILQR algorithm [Fig. 3(a)] is similar to a lateral error response curve for vehicle trajectory following generated by a nonlinear  model including uncertainty in tire stiffness parameters (Fig. 4 in \cite{Guo18}), demonstrating that soft-CILQR can generate realistic solution trajectories without solving computationally expensive nonlinear models.

Fig. 8 presents the  $\delta ^* $  trajectories generated by the soft-CILQR and CILQR-cc solvers for $\sigma $ = 2.0. The CILQR-cc solution trajectories were generated with three sets of cost coefficients: ($q_{c1}$, $q_{c2}$) = (5, 5), (10, 10), and (15, 15). The CILQR-cc controller tended to eliminate feature peaks of the $\delta ^* $  trajectories, and this behavior increased as the penalty for steering signal changes increased by increasing $q_{c1}$ and $q_{c2}$. This in turn decreased the vehicle yaw agility and was thus unfavorable for high-speed cornering. In subsequent vision-based experiments (Fig. 17), this lack of agility resulted in the ego vehicle controlled by CILQR-cc leaving the lane unintentionally at high speed; the ego vehicle guided by soft-CILQR completed the lane-keeping task in this scenario.

\begin{table}[!t]
\caption{Average computation time of the CILQR, soft-CILQR, MPC, and soft-MPC algorithms$^{a}$}
\begin{center}
\begin{tabular}{c|c|c|c|c}
\hline
Method &CILQR &soft-CILQR$^{b}$ & MPC &  soft-MPC$^{b}$ \\ \hline
Latency (ms)&0.96& 2.55 & 15.03 & 49.80 \\  \hline
\multicolumn{5}{l}{$^{a}$\scriptsize{All algorithms were implemented in C++.}} \\
\multicolumn{5}{l}{$^{b}$\scriptsize{The worst-case running time for soft-CILQR/soft-MPC is 
3.53/149.64 ms.}} 
\end{tabular}
\end{center}
\end{table}

Fig. 9--11 present the solution trajectories of Problem 3 obtained by the soft-MPC and MPC solvers using the IPOPT software package. Comparing Fig. 9 ($S$ = 0.01) and Fig. 10 ($S$ = 0.5), the soft-MPC solver with a larger weight parameter $S$ = 0.5 generates a similar $\Delta$ trajectory to that of soft-CILQR solver [Fig. 3(a)]. The  magnitude of $\theta$ and $\delta ^* $ solutions produced by MPC-based solvers are generally smaller than those produced by CILQR-based solvers [Fig. 3(b)] because the IPOPT solver was constructed on the basis of the primal--dual barrier method \cite{Wac06}, which strongly rejects solutions that approach the constraint boundaries during optimization, whereas the CILQR-based solvers only perform variable optimization in primal space. Consequently, solutions yielded by CILQR-based solvers may exceed the constraint boundaries and must be truncated [Fig. 3(a) and 3(b)]. However, in Fig. 11(b), the small control signals yield by MPC-based solvers had substantial fluctuations due to disturbance, and the resulting  $\delta ^* $  trajectories were therefore noisier than those of CILQR-based solvers at the same noise level ($\sigma $ = 1.0) [Fig. 4(b)]. In particular, the soft-MPC solver generated smoother $\delta ^* $  trajectories than did the MPC solver when 100 $< t <$ 200; this behavior was similar to that of the soft-CILQR solver. Table III presents the average computation time for obtaining a solution for the CILQR-based and MPC-based algorithms. Because CILQR and MPC have fewer variables than do soft-CILQR and soft-MPC, these soft algorithms have longer  latency. The soft-CILQR solver achieved faster computations than did the soft-MPC solver;  in particular, the computation time for the soft-MPC solver was 19.5 times longer than that of the soft-CILQR solver. Long-latency controllers, such as soft-MPC, are liable to not only increasing ego vehicle instability but also engaging in unsafe cornering at high speed in TORCS environments \cite{Lee22}.

\begin{figure}[!t]
{\includegraphics*[width=0.5\linewidth]{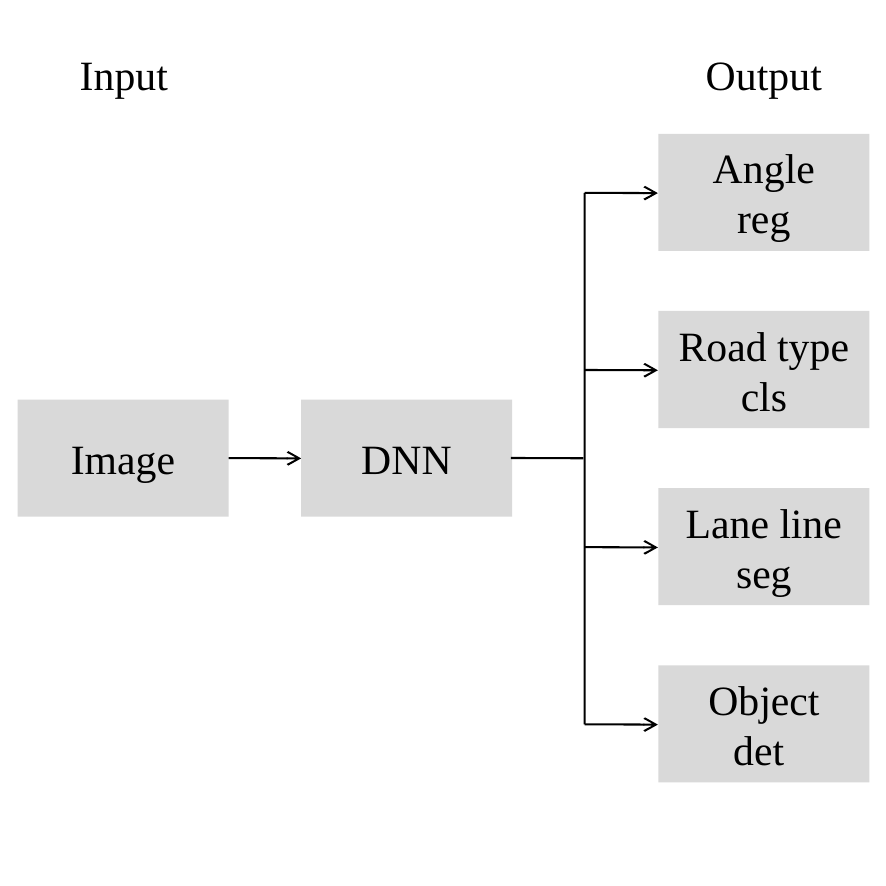}}%
{\includegraphics*[width=0.5\linewidth]{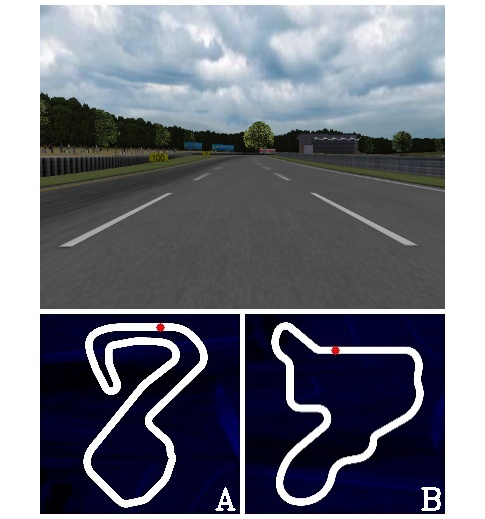}}%
\caption{Illustration of the vision-based driving simulations. (Left) Perception module. (Right) Example traffic scene in TORCS and testing tracks A and B. The red symbols on the track maps represent the starting locations of the ego vehicle. The driving direction was counterclockwise.}
\end{figure}

\begin{figure}[!t]
\centerline{\includegraphics[scale=0.33]{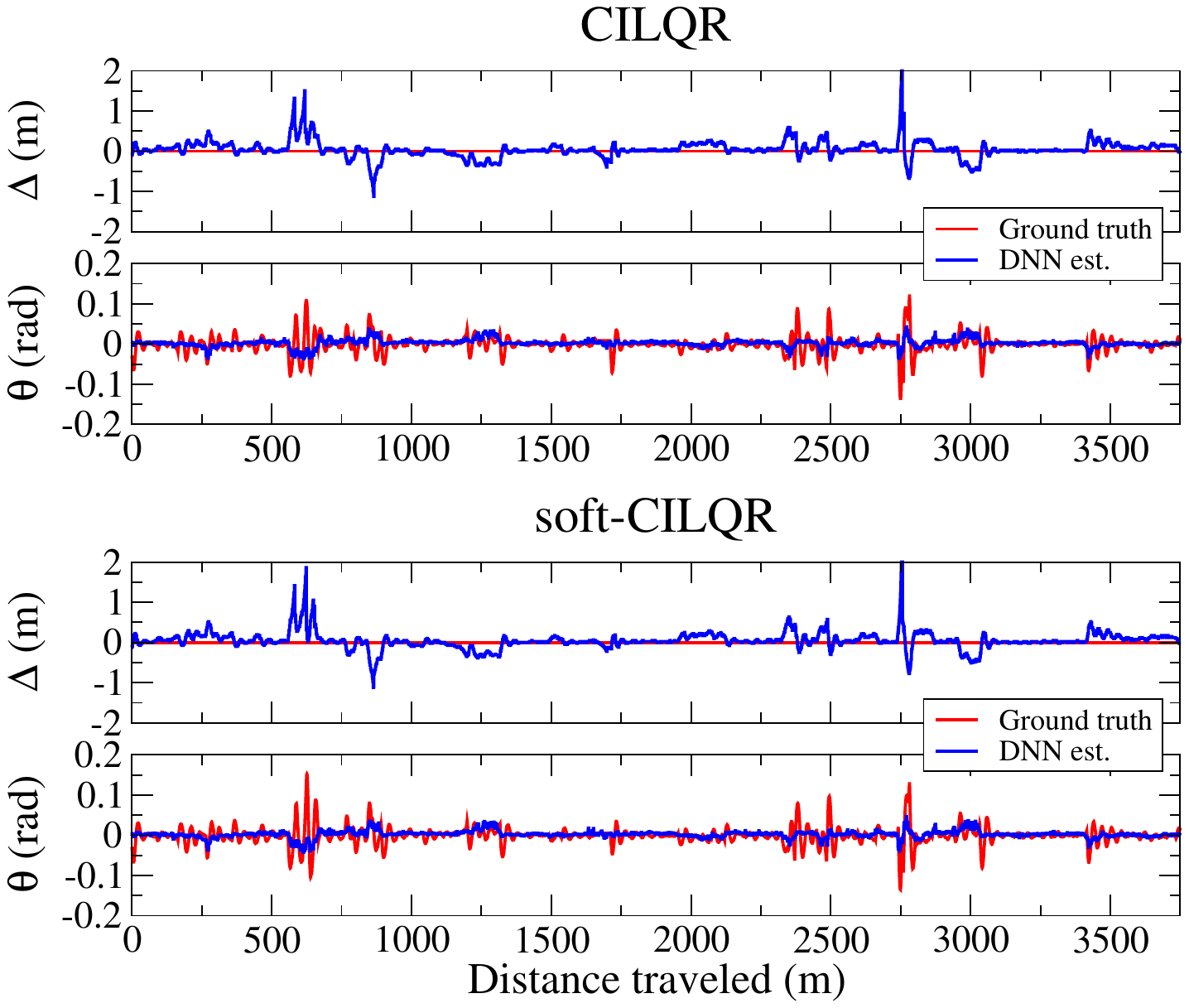}}
\caption{Trajectories obtained for $\Delta$ and $\theta $ by using the CILQR and soft-CILQR controllers under $\sigma$ = 0.0 and $fr$ = 1.10, when the ego vehicle traveled on track A at 72 km/h. This experiment corresponds to Exp. V01 in Table  IV.}
\end{figure}

\begin{table*}[!h]
\caption{Performance of the CILQR and soft-CILQR algorithms in the vision-based experiments }
\begin{center}
\begin{tabular}{c|c|c|c|ccc|ccc}\hline
\multirow{3}{*}{Exp.} &\multirow{3}{*}{Track}&  \multirow{3}{*}{$\sigma $}&  \multirow{3}{*}{$fr$ }& \multicolumn{3}{c|}{CILQR} & \multicolumn{3}{c}{soft-CILQR} \\ \cline{5-10}
&& & & $\Delta $ (m)  &  $\theta$  (rad) &  $\delta ^*$  (rad)  &  $\Delta $ (m)  &  $\theta$  (rad) &  $\delta ^*$  (rad) \\ 
&& &&  MAE  & MAE  & RMS  & MAE  & MAE   & RMS  \\ \cline{5-10} \cline{1-4}
V01 & A& \multirow{2}{*}{0.0} & \multirow{6}{*}{$1.10 $}&  0.134443   &  0.012826   & 0.087147  & 0.136039   &  0.013431  & 0.088188 \\ 
V02 &B&  &&   0.151334  &  0.009582   &  0.082993  &    0.154751  &   0.009818  & 0.083336  \\ \cline{5-10} \cline{1-3}
V03 & A& \multirow{2}{*}{1.0} &&  0.142675    &   0.014832 &   0.091773 &   0.139885  &    0.013952 &  0.090899  \\ 
V04 &B&  &&   0.155663  & 0.010493  &  0.084347 &  0.154971   &  0.010357  &  0.084236 \\ \cline{5-10} \cline{1-3}
V05&  A& \multirow{2}{*}{2.0} &&  0.153503   & 0.017434 &  0.100032 & 0.145407  &  0.016240	  &  0.094481 \\ 
V06 &B&  &&   0.158078  &   0.012392 & 0.088490  &    0.160305  & 	0.012106 & 0.088841 \\ \hline
V07 & A& \multirow{2}{*}{0.0} & \multirow{6}{*}{$1.00 $}&  0.136733  &  0.013782  & 0.090276  &    0.146863  & 0.014916   &  0.096403   \\ 
V08 &B&  &&  0.152588 &  0.009855   & 0.083584    &    0.156006  &  0.010198 &   0.084547  \\ \cline{5-10} \cline{1-3}
V09 & A& \multirow{2}{*}{1.0} &&  0.142147   & 0.015187   &  0.095604  &  0.145087   &  0.015163   & 0.093450   \\ 
V10 &B&  &&  0.155764   &   0.010807 &  0.085627 & 0.157716  & 0.010798  & 0.085890 \\ \cline{5-10} \cline{1-3}
V11&  A& \multirow{2}{*}{2.0} && 0.157751     &   0.018291  &0.103462   &  0.155017   & 0.017254   &   0.101396 \\ 
V12 &B&  &&   0.161967  &  0.013593  & 0.091674   &  0.163669   &  0.013295 & 0.089640 \\ \hline
V13&  A& \multirow{6}{*}{$1.0 $}&  \multirow{2}{*}{$0.90$}  &   0.158768    &  0.017165   & 0.103390   &   0.158103 &   0.017509  &   0.102171 \\ 
V14 &B&  &&   0.160325   &  0.011621  &  0.088449  &   0.161509   & 0.011940     &   0.087304 \\ \cline{4-10} \cline{1-2}
V15 & A&&  \multirow{2}{*}{$0.88$} & 0.157093   &  0.017256  &  0.102968  &  0.159489  &  0.017774  &  0.102968 \\ 
V16 &B&  &&  0.162341 &  0.012023  &  0.088251 &   0.159742 & 0.011485   &  0.086521 \\ \cline{4-10} \cline{1-2}
V17 & A&  & \multirow{2}{*}{$0.86$}&  0.160879   & 0.018133     &  0.104509   &   0.164370  &  0.018396  &   0.106209  \\ 
V18 &B&  &&  0.160974  &   0.011908   &  0.088465  & 0.166148    &  0.012416   & 0.089899   \\ \hline
\multicolumn{4}{c|}{Average}  &  0.153501   &  0.013732   &  0.092280  &   0.154726 &  0.013724
  &   0.092021
 \\ \hline
\end{tabular}
\end{center}
\end{table*}

\begin{figure}[!t]
{\includegraphics*[width=0.33\linewidth]{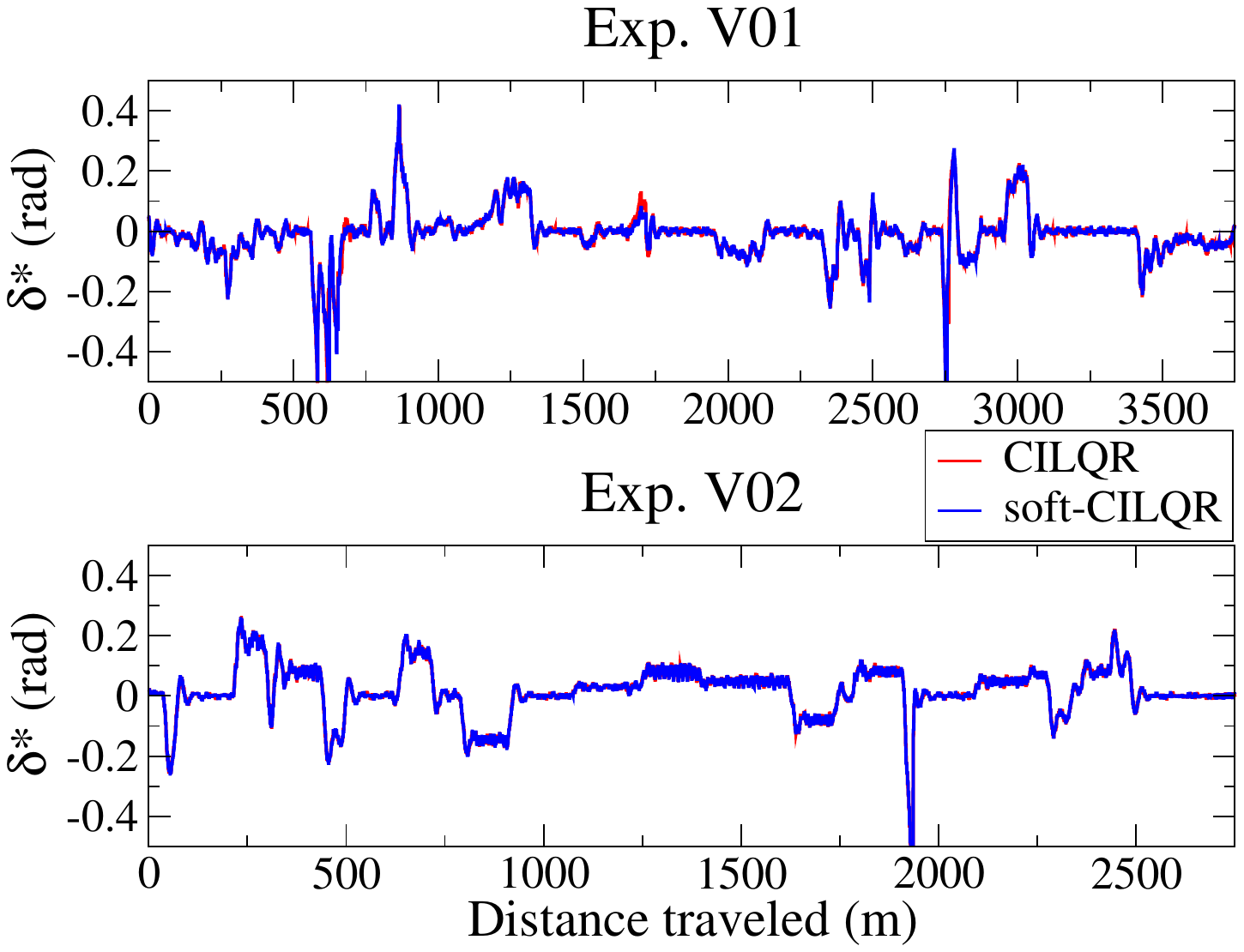}}%
{\includegraphics*[width=0.33\linewidth]{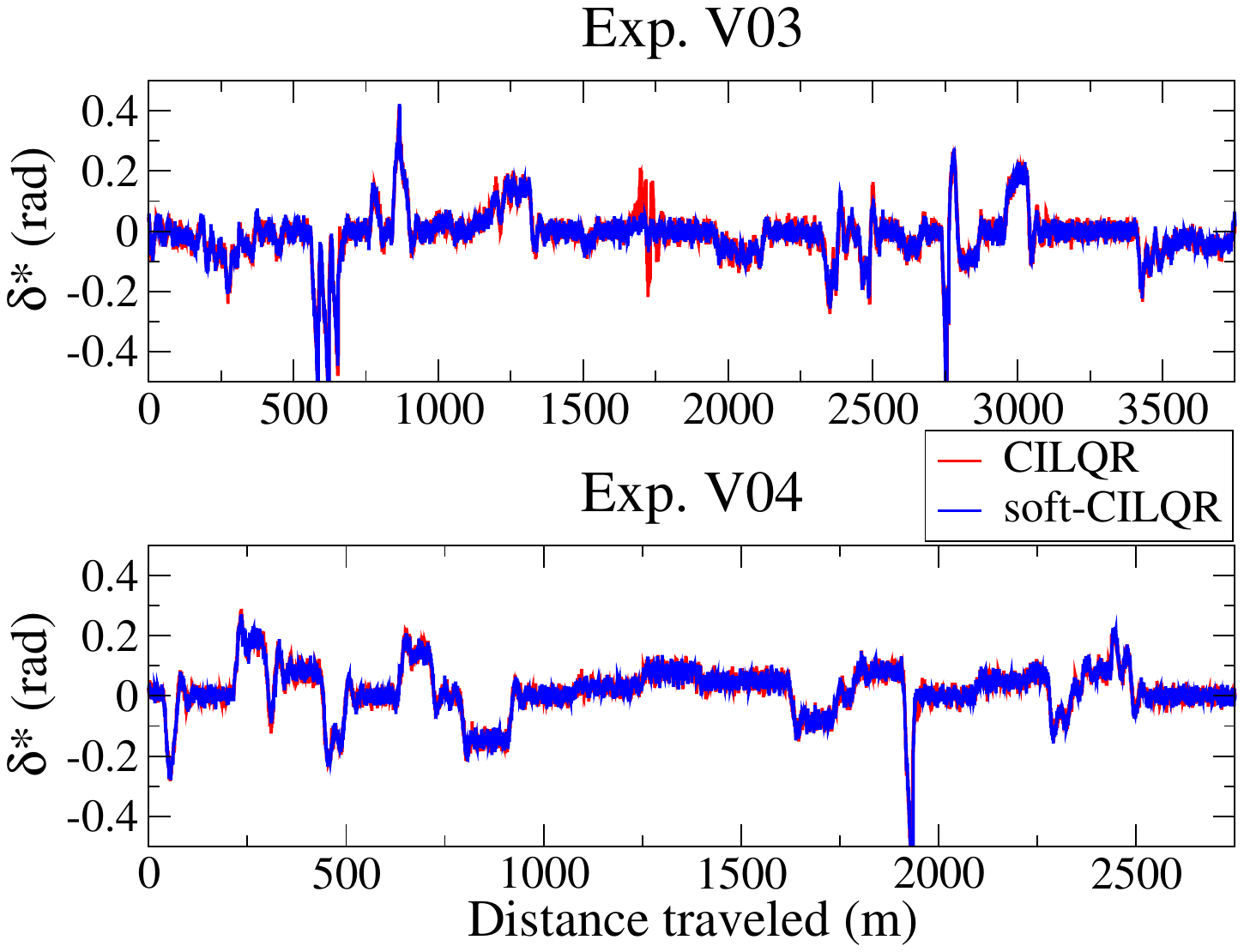}}%
{\includegraphics*[width=0.33\linewidth]{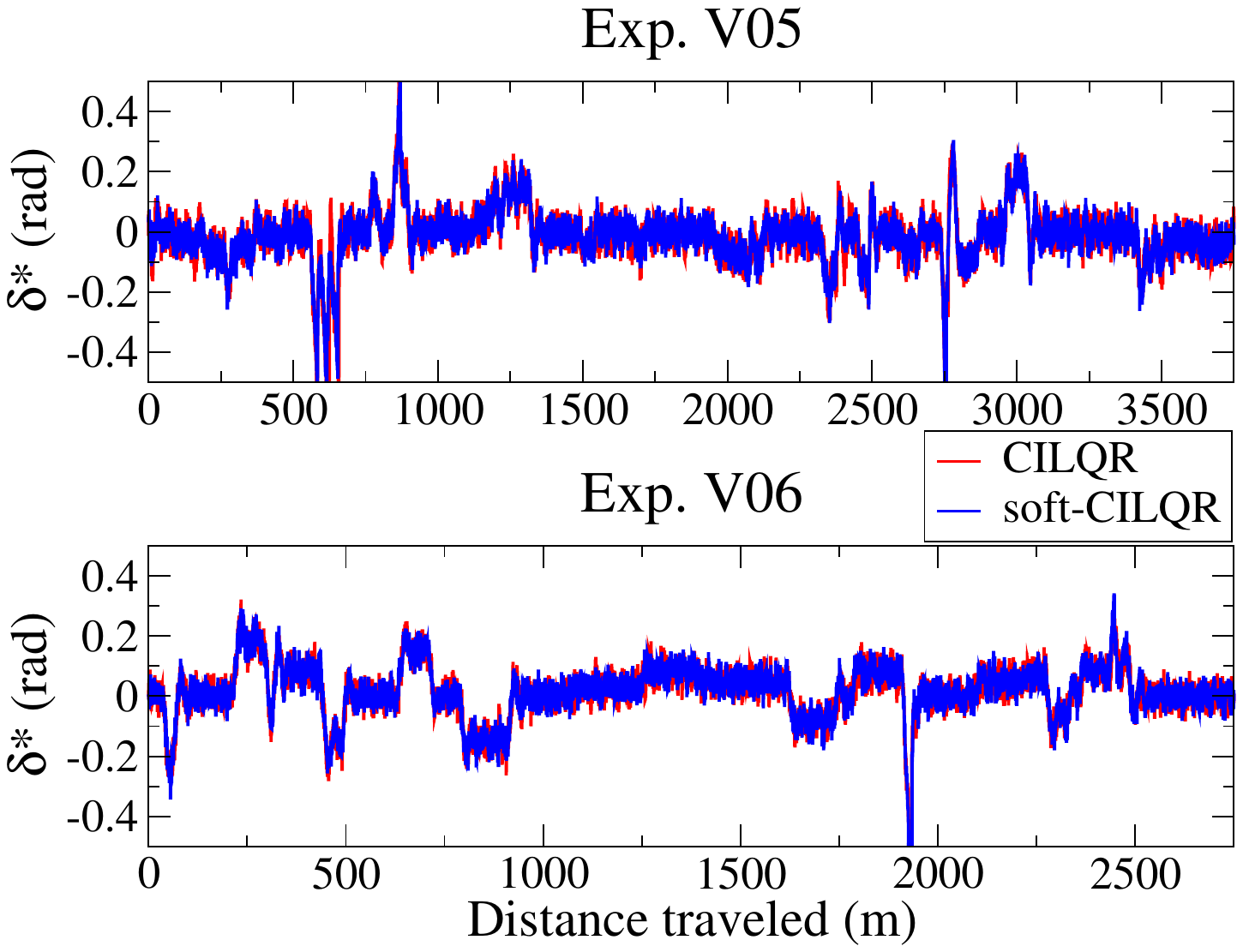}}%
\caption{Trajectories obtained for $\delta ^* $ in Exp. V01/02, Exp. V03/04, and Exp. V05/06  when $fr$ = 1.10 and $\sigma$ = 0.0, 1.0, and 2.0, respectively. The results of the associated quantitative analyses are presented in Table  IV.}
\end{figure}

On the basis of above numerical simulation results, we used a $N$ value of 40 and a $\varepsilon _{\max } $ value of 49 in the vision-based experiments described in next section to validate the performance of the proposed soft-CILQR algorithm and to compare its performance with those of the CILQR and CILQR-cc algorithms.

\begin{figure}[!t]
{\includegraphics*[width=0.33\linewidth]{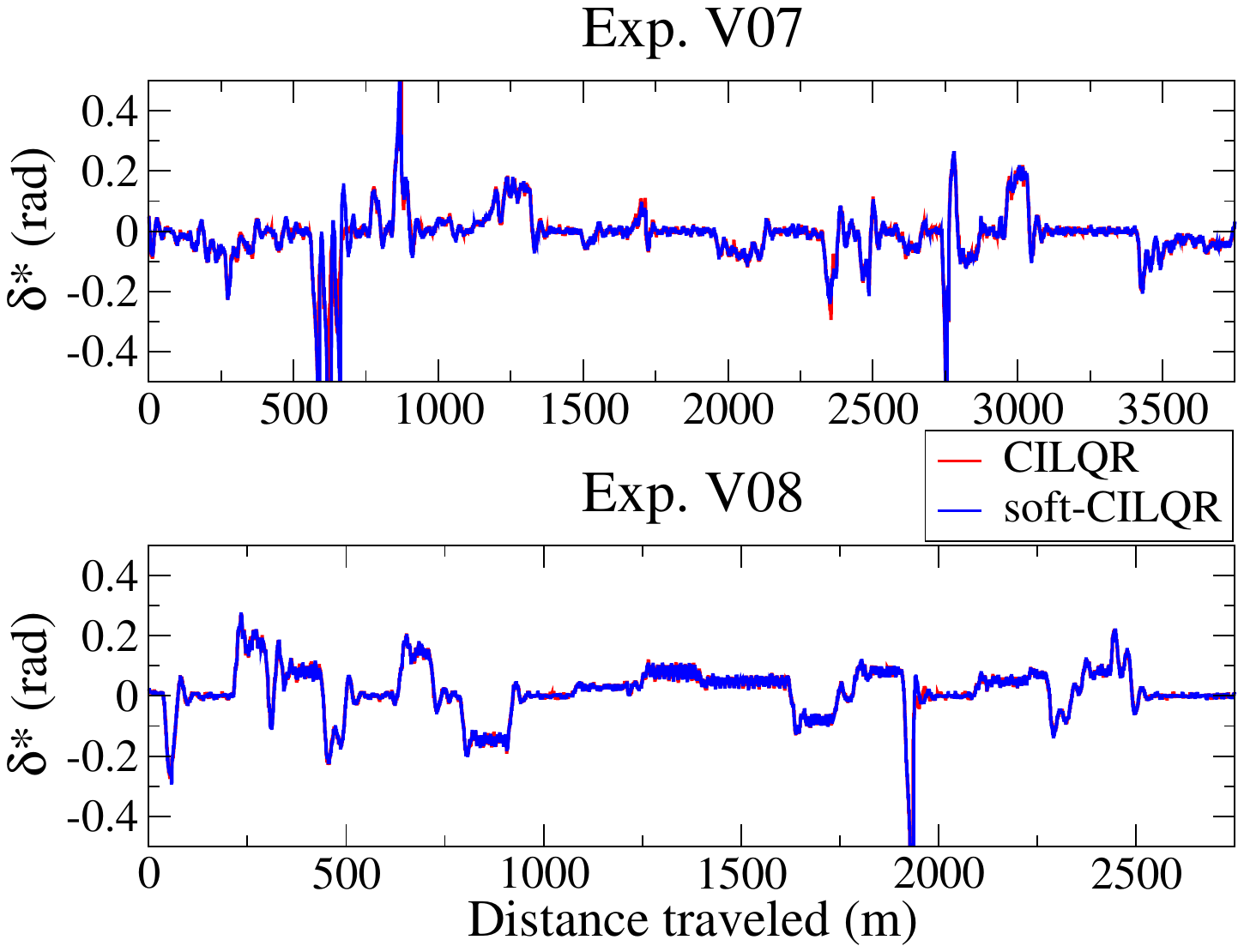}}%
{\includegraphics*[width=0.33\linewidth]{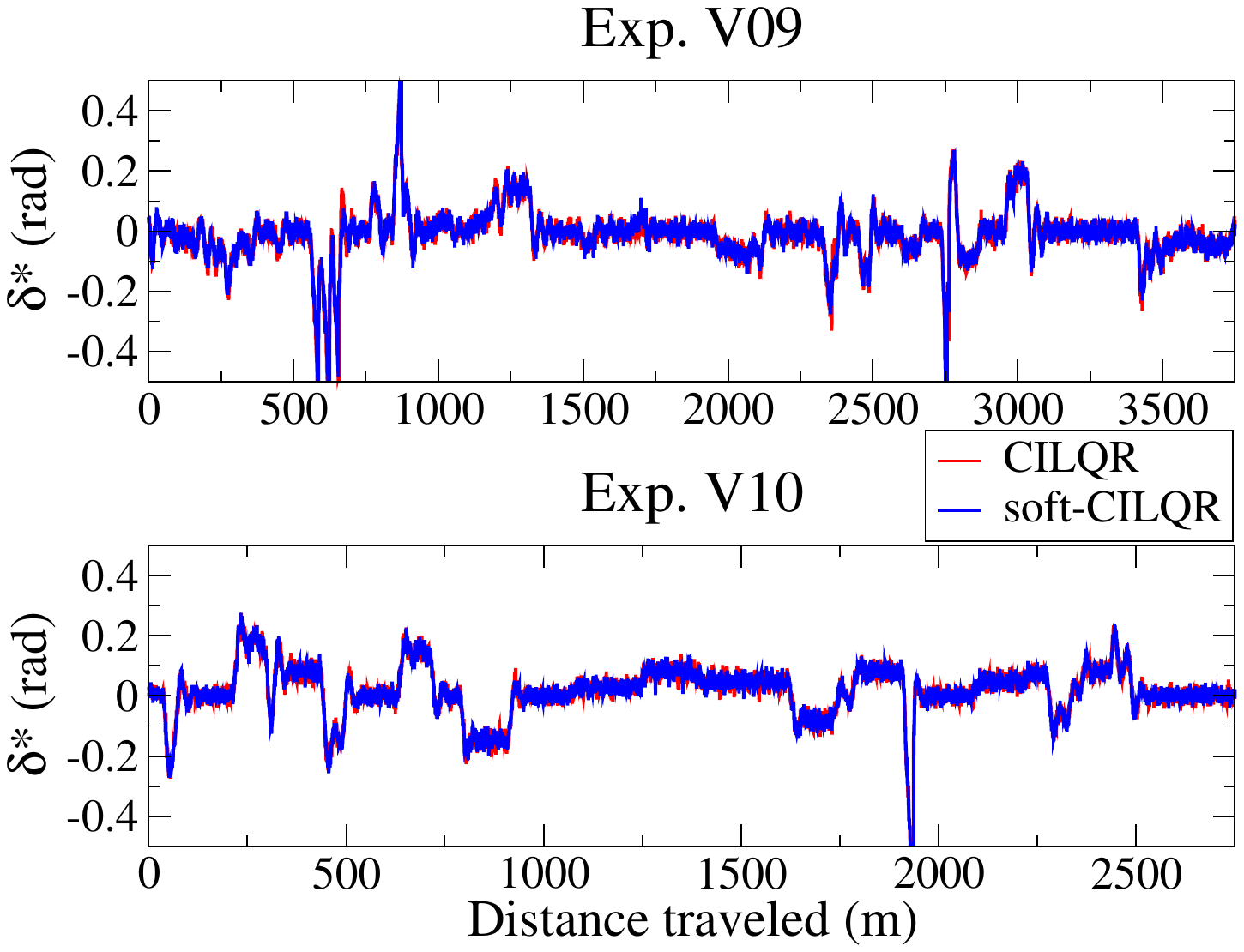}}%
{\includegraphics*[width=0.33\linewidth]{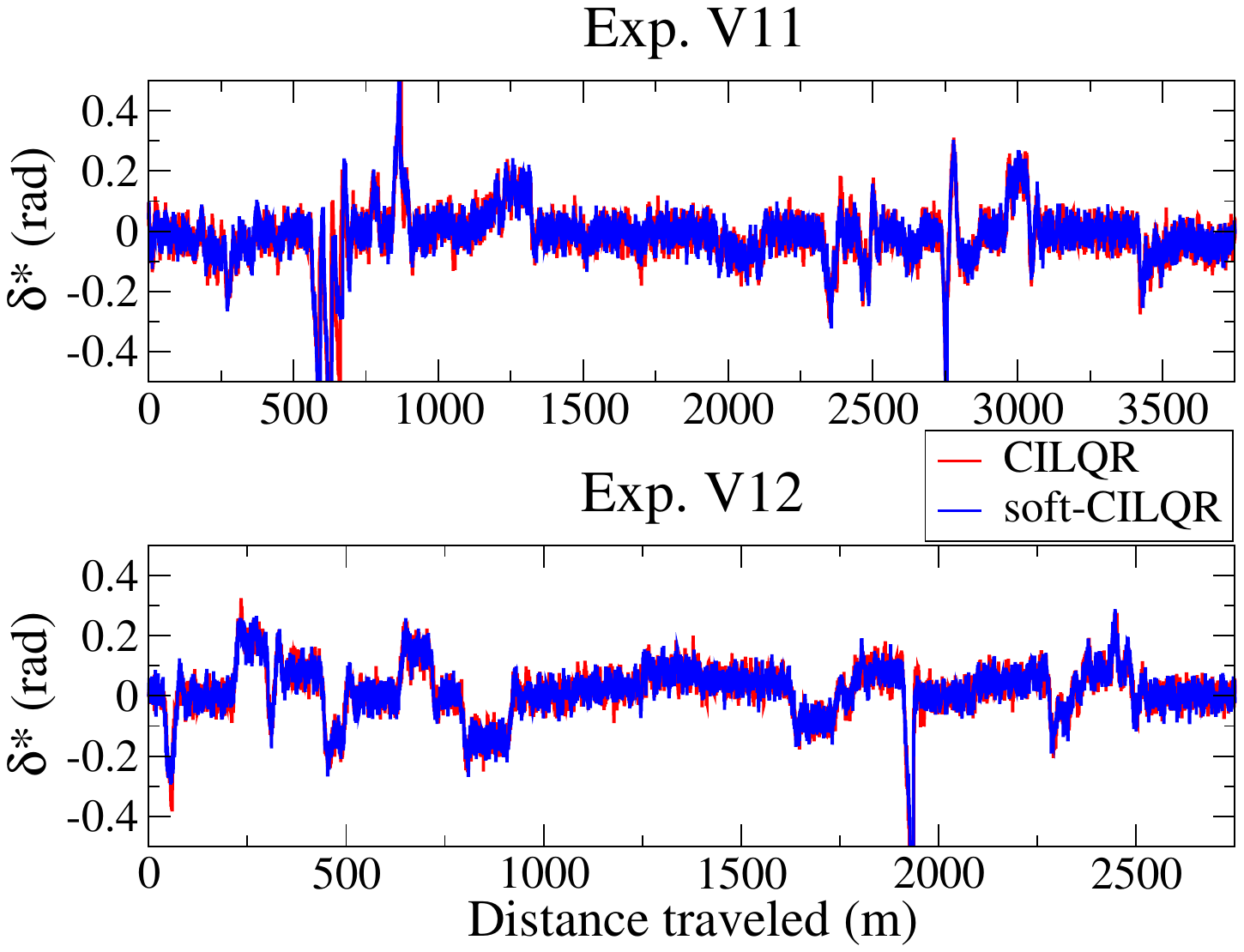}}%
\caption{Trajectories obtained for $\delta ^* $ in Exp. V07/08, Exp. V09/10, and Exp. V11/12  when $fr$ = 1.00 and  $\sigma$ = 0.0, 1.0, and 2.0, respectively. The results of the associated quantitative analyses are presented in Table  IV.}
\end{figure}

\begin{figure}[!t]
{\includegraphics*[width=0.33\linewidth]{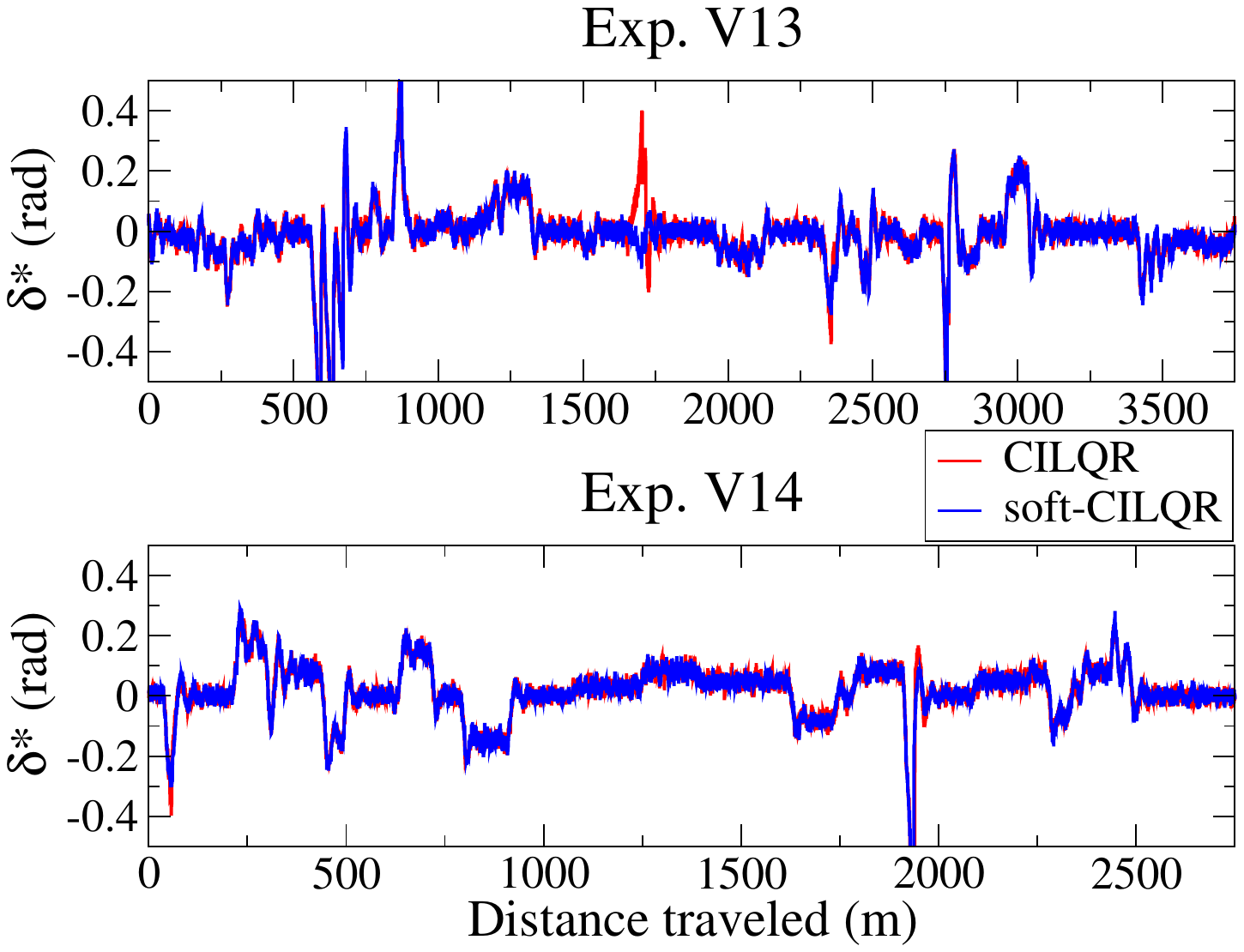}}%
{\includegraphics*[width=0.33\linewidth]{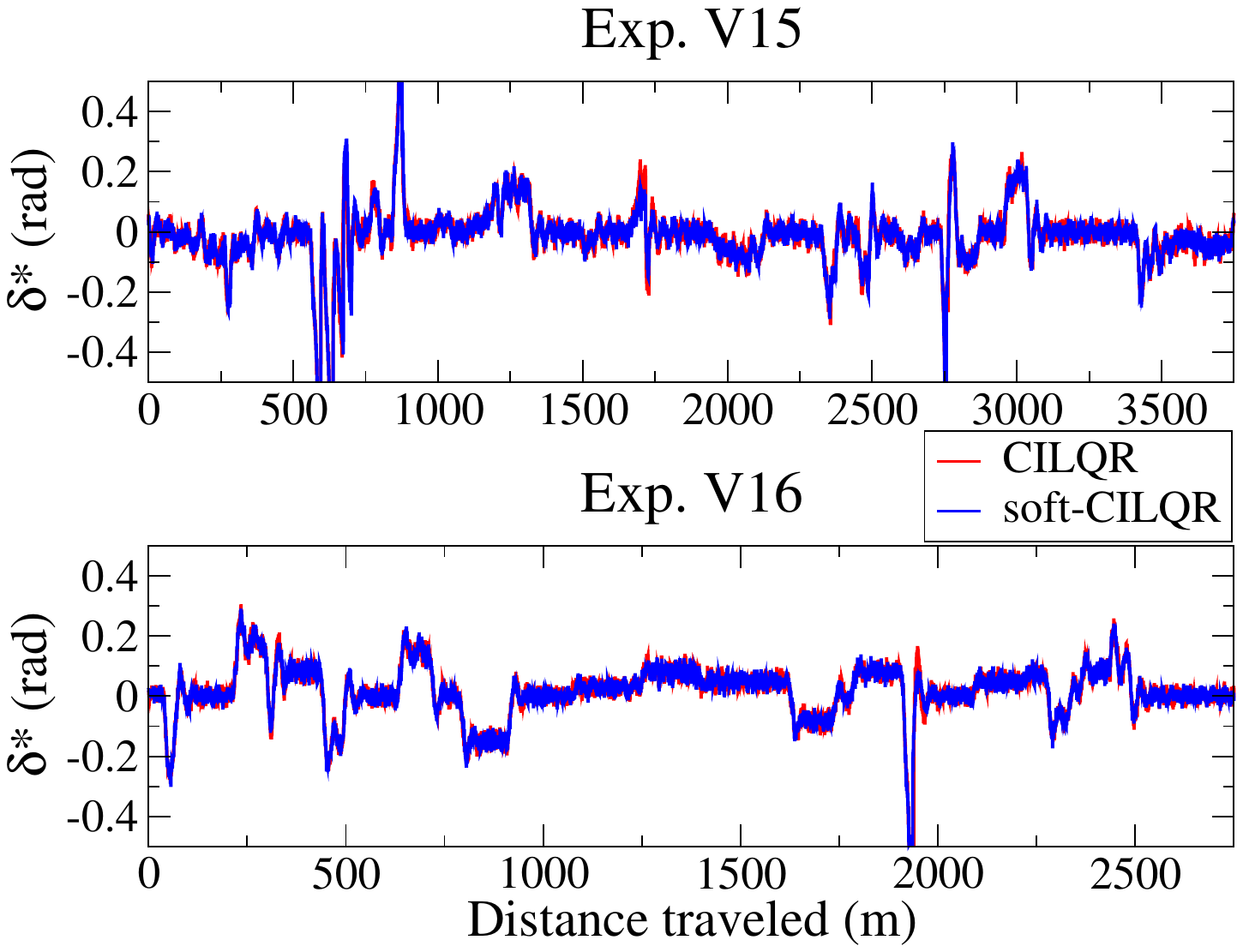}}%
{\includegraphics*[width=0.33\linewidth]{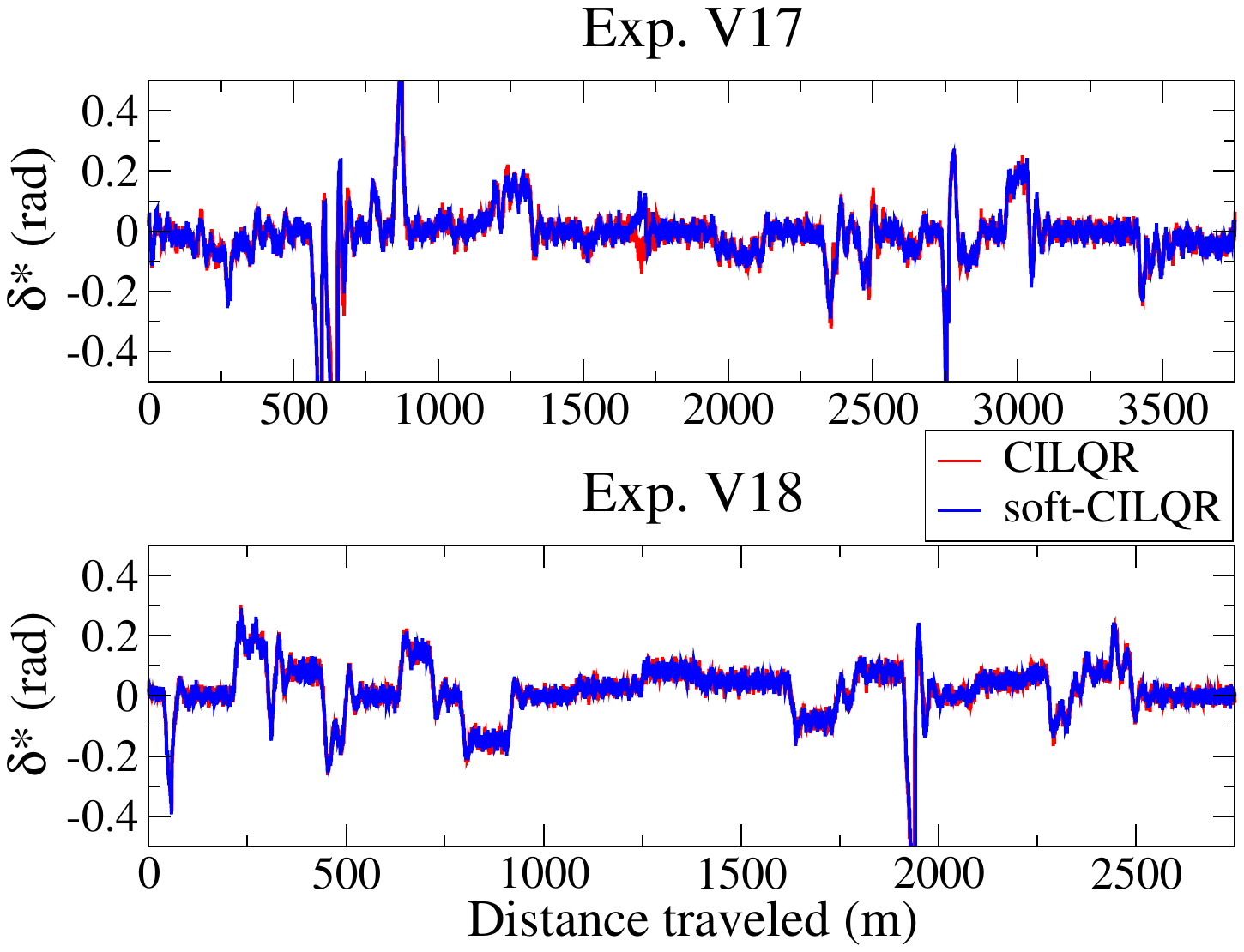}}%
\caption{Trajectories obtained for $\delta ^* $ in Exp. V13/14, Exp. V15/16, and Exp. V17/18  when  $\sigma$ = 1.0 and $fr$ = 0.90, 0.88, and 0.86, respectively. The results of the associated quantitative analyses are presented in Table  IV.}
\end{figure}

\begin{figure}[!t]
\centerline{\includegraphics[scale=0.33]{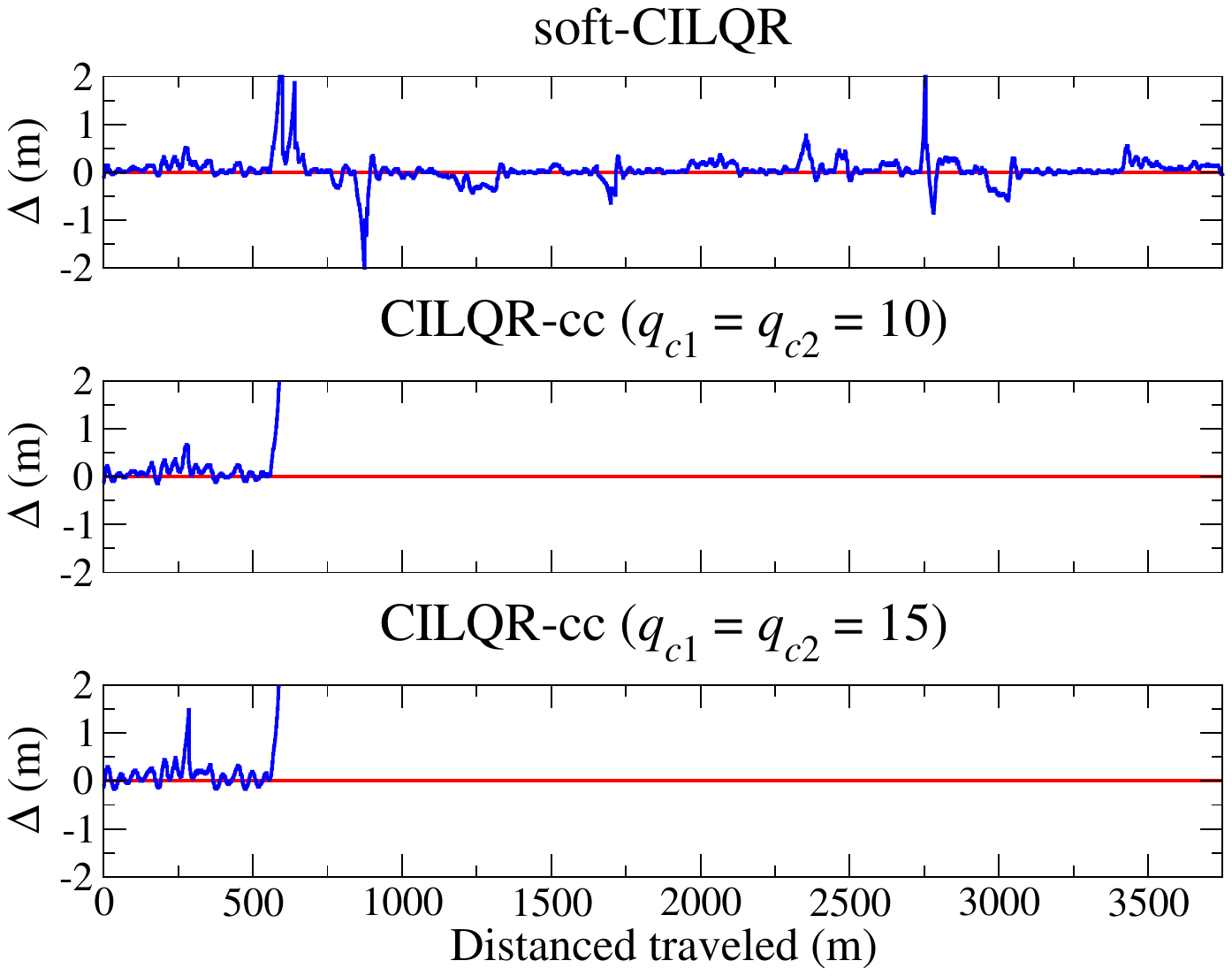}}
\caption{Trajectories obtained for $\Delta$ by using the soft-CILQR and CILQR-cc controllers under  $\sigma$ = 1.0 and $fr$ = 1.00 for the ego vehicle traveling on track A at 80 km/h. }
\end{figure}

\subsection{Vision-Based Lane-Keeping Experiments}
The experimental setup used in the vision-based experiments (Fig. 12) is briefly described as follows. Additional details are provided in our previous study \cite{Lee22}. First, a DNN  (a multi-task UNet \cite{Ron15}) with 25.50 million parameters and an input image size of 228 $\times$ 228 pixels was used for environment perception. The adopted DNN can extract features from RGB  images and then perform heading angle regression, road type classification, lane line segmentation, and traffic object detection simultaneously for the ego vehicle at a speed of approximately 40 FPS (frames per second). Second, the example driving scene considered in the vision-based experiments on TORCS was a highway with a lane width of 4 m. Third, two tracks were used in the autonomous driving simulation, namely tracks A and B, which had total lengths of 3919 and 2843 m, respectively. The maximum curvatures of tracks A and B were approximately 0.05 and 0.03 1/m, respectively. These two tracks were not used for DNN training; they were instead used to test the DNN model's generalization performance \cite{Wu23}. The CILQR-based algorithms were used to control the ego vehicle in the central lane of tracks A and B through lane-keeping maneuvers. A proportional\textendash integral (PI) longitudinal speed controller \cite{Sam21} was  adopted to control the ego vehicle on tracks A and B at a constant speed of 72/80 and 60 km/h (20.0/22.2 and 16.6 m/s), respectively. The other parameter and constraint values are listed in Tables I and II. A time delay problem occurred in this vision-based vehicle control framework because the latency of the vehicle actuators (6.66 ms) was shorter than the end-to-end latency between the perception and control modules \cite{Lee22}. This problem might result in self-driving vehicles performing poorly on curved roads. Because the average computation times of the soft-CILQR (2.55 ms) and CILQR (0.96 ms) algorithms were shorter than the actuator latency in this study, the influence of this time delay problem was ignored in this study.

Vision-based experiments were performed at various noise levels ($\sigma $ = 0.0, 1.0, and 2.0). Moreover, road adhesion was also varied to test the performance of the controllers; the road adhesion can be specified in the TORCS track files by setting the friction parameter (denoted as $fr$) \cite{Bon17}. The test values in this study for $fr$ were in [0.86, 1.10]; for a typical road, $fr$ is approximately 1.0.  The  solvers were initialized by setting the control and slack variables to zeros. The vector ${\bf y}_t$ for computing the optimal steering angle sequence is defined as:
\begin{equation}
{\bf y}_t  = {\bf Cx}_t  + \sigma {\bf w}_t, 
\end{equation}
where
\[
{\bf x} = \left[ {\begin{array}{*{20}c}
   {\Delta _D }  \\
   0  \\
   {\theta _D }  \\
   0  \\
\end{array}} \right],\quad{\bf C} = \left[ {\begin{array}{*{20}c}
   1 & 0 & 0 & 0  \\
   0 & 1 & 0 & 0  \\
   0 & 0 & 1 & 0  \\
   0 & 0 & 0 & 1  \\
\end{array}} \right].
\]
The current states $\Delta _{D}$ and $\theta _{D}$ were obtained using the DNN and relevant postprocessing methods \cite{Lee22}. After the optimization process was completed, the first element of the optimal steering angle sequence ($\delta ^*$) was  entered into the TORCS  engine to control the ego vehicle.

The validation results obtained for the CILQR and soft-CILQR solvers on track A when $\sigma$ = 0.0 are  displayed in Fig. 13. In the figure, the DNN-estimated and  ground-truth $\Delta $ and $\theta $ data are also shown. When the CILQR and soft-CILQR solvers were used at different $\sigma$ and $fr$ values on tracks A and B (a total of 36 maneuvers), the ego vehicle completed the entire range of lane-keeping maneuvers  without leaving the road unintentionally. The mean absolute errors (MAEs) of $\Delta $ and $\theta $ \cite{Lee22} as well as the root mean square errors (RMSs) of the optimal steering angle $\delta ^* $ \cite{Chi21} for both solvers are listed in Table IV. The MAEs and RMSs can be considered to represent tracking errors and  steering smoothness scores, respectively, with lower RMS values for  $\delta ^* $ indicating smoother steering. The standard deviation (SD) can also reflect control performance  \cite{Wu23}. As presented in Table IV, as $\sigma$ increased, the output was more susceptible to disturbance, and the ego vehicle controlled using the soft-CILQR controller exhibited smoother steering that was generally similar to that in the numerical simulations [Fig. 4(b) and Fig. 5(b)]. Consequently, the average $\delta ^* $-RMS value derived for the soft-CILQR controller was lower than that derived for the CILQR controller by 0.000259 rad. However, due to the inherently less conservative behavior of the soft-CILQR controller as described in numerical simulations [Fig. 3], the ego car had higher tracking errors with the soft-CILQR controller ($\Delta $-MAE) than with the CILQR controller. Therefore, the results of these vision-based experiments were consistent with those of the numerical simulations in the previous section. Fig. 14--16 presents the $\delta ^* $ trajectories obtained using the CILQR (red curves) and soft-CILQR (blue curves) controllers. The soft-CILQR controller generally attenuated the jitter in the steering inputs to a greater extent than did the CILQR controller.

Lane-keeping maneuvers as illustrated in Fig. 17 were implemented on track A at a higher cruise speed (80 km/h) to compare the performance of the CILQR-cc and soft-CILQR controllers. The ego vehicle guided  by the soft-CILQR controller successfully completed one lap on track A; however, the CILQR-cc controllers caused the car to leave the track unintentionally at approximately 600 m  when $q_{c1,2}$ = 10 or 15.  As shown in Fig. 8, the soft-CILQR controller could adaptively suppress fluctuations of $\delta ^* $,  maintaining the agility of the self-driving car in the yaw direction; the CILQR-cc controller could not do so, resulting in unsafe driving behavior when the ego car entered tight turns.

\section{Conclusions}
In this study, we integrated a constraint softening method \cite{Svr23} and the  CILQR algorithm \cite{Che17,Che19} to develop a novel soft-CILQR algorithm  for solving the lane-keeping control problem for automated vehicles under linear system dynamics. The proposed soft-CILQR algorithm was evaluated in numerical simulations and vision-based experiments using a path-tracking model \cite{Li19, Mat19, Lee19} under conditions involving additive disturbances. In the numerical simulations, state trajectories generated by the soft-CILQR algorithm  were less conservative than those of the CILQR algortihm; the soft-CILQR algorithm also guaranteed the asymptotic convergence of the state trajectories to the target point. The CILQR algorithm achieved this convergence in the absence of noise ($\sigma $ = 0.0). However, if noise was present ($\sigma $ = 1.0 or 2.0), the soft-CILQR solver generated smoother steering inputs than did the CILQR solver without the use of additional smoothing filters. In the vision-based experiments performed using the experimental setup from our previous work \cite{Lee22}, the soft-CILQR algorithm exhibited better performance in terms of steering smoothness than did the CILQR algorithm when controlling an autonomous vehicle on TORCS during lane-keeping tasks. The fluctuations of the steering solution trajectories yielded by the soft-CILQR solver can also be adaptively suppressed, preserving the yaw agility of ego vehicle and ensuring driving safety during high-speed cornering (unlike the ego car guided by the CILQR-cc solver). The soft-CILQR solver had a fast computation time of 2.55 ms in these experiments, whereas the soft-MPC solver based on IPOPT software library  with the same soft constraints requires a  computation time 19.5 times longer for the same task. In conclusion, the consistency of the results of the numerical and vision-based tests demonstrate that the proposed soft-CILQR method has excellent potential for reducing steering instability to real-time autonomous driving systems under adverse  conditions.



\end{document}